\documentclass[superscriptaddress,
 amsmath,amssymb, aps, prb,citeautoscript, twocolumn
]{revtex4-2}

\usepackage{siunitx}
\usepackage{booktabs}
\usepackage{pifont}
\usepackage{bm}

\usepackage{tcolorbox}
\tcbuselibrary{skins, breakable}

\newcommand{\cmark}{\ding{51}}
\newcommand{\xmark}{\ding{55}}

\usepackage{hyperref}
\usepackage[capitalize]{cleveref}
\hypersetup{
    colorlinks,
    citecolor=blue,    filecolor=blue,
    linkcolor=black,    urlcolor=blue
}

\definecolor{jblueleft}{RGB}{29, 50, 138}
\definecolor{black}{RGB}{0, 0, 0}
\definecolor{jblueinner}{RGB}{251, 252, 252}

\tcbset{
  textmarker/.style={
    enhanced,
    sharp corners,
    boxrule=0pt,  
  }}

\makeatletter
\newtcolorbox{exampleBox}{textmarker,
  borderline west={3pt}{0pt}{jblueleft},
    colback=jblueinner, breakable, enhanced
  }
 
\newtcolorbox{Note}{textmarker,
  borderline west={3pt}{0pt}{black},title=Note,
    colback=jblueinner, breakable, enhanced
  }

 \makeatother

\newcommand{\example}[1]{%
  \begin{exampleBox}   
    #1%
  \end{exampleBox}%
}

\begin{document}
\setcitestyle{super}

\title{Data-driven Science and Machine Learning Methods in Laser-Plasma Physics}

\author{Andreas Döpp}
\email{a.doepp@lmu.de}
\affiliation{Fakultät für Physik, Ludwig--Maximilians--Universit{\"a}t M{\"u}nchen, Am Coulombwall 1, 85748 Garching, Germany}%
\affiliation{Department of Physics, Clarendon Laboratory, University of Oxford, Parks Road, Oxford OX1 3PU, United Kingdom}%

\author{Christoph Eberle}
\affiliation{Fakultät für Physik, Ludwig--Maximilians--Universit{\"a}t M{\"u}nchen, Am Coulombwall 1, 85748 Garching, Germany}%

\author{Sunny Howard}
\affiliation{Fakultät für Physik, Ludwig--Maximilians--Universit{\"a}t M{\"u}nchen, Am Coulombwall 1, 85748 Garching, Germany}%
\affiliation{Department of Physics, Clarendon Laboratory, University of Oxford, Parks Road, Oxford OX1 3PU, United Kingdom}%

\author{Faran Irshad}
\affiliation{Fakultät für Physik, Ludwig--Maximilians--Universit{\"a}t M{\"u}nchen, Am Coulombwall 1, 85748 Garching, Germany}%

\author{Jinpu Lin}
\affiliation{Fakultät für Physik, Ludwig--Maximilians--Universit{\"a}t M{\"u}nchen, Am Coulombwall 1, 85748 Garching, Germany}%

\author{Matthew Streeter}
\affiliation{Centre for Plasma Physics, Queens University Belfast, Belfast BT7 1NN, United Kingdom}%

\begin{abstract}

Laser-plasma physics has developed rapidly over the past few decades as high-power lasers have become both increasingly powerful and more widely available. Early experimental and numerical research in this field was restricted to single-shot experiments with limited parameter exploration. However, recent technological improvements make it possible to gather an increasing amount of data, both in experiments and simulations. This has sparked interest in using advanced techniques from mathematics, statistics and computer science to deal with, and benefit from, big data. At the same time, sophisticated modeling techniques also provide new ways for researchers to effectively deal with situations in which still only sparse amounts of data are available. This paper aims to present an overview of relevant machine learning methods with focus on applicability to laser-plasma physics, including its important sub-fields of laser-plasma acceleration and inertial confinement fusion.

\end{abstract}

\maketitle

\newpage

\tableofcontents

\newpage

\section{Introduction}

\subsection{Laser-Plasma Physics}
Over the past decades, the development of increasingly powerful laser systems\cite{Danson.2015,Danson.2019} has enabled the study of light-matter interaction across many regimes. 
Of particular interest is the interaction of intense laser pulses with plasma,  which is characterized by strong nonlinearities that occur across many scales in space and time\cite{Pukhov.2003,Marklund.2006}. These laser-plasma interactions are both of interest for fundamental physics research and as emerging technologies for potentially disruptive applications. 

Regarding fundamental research, high-power lasers have for instance been used to study transitions from classical electrodynamics to quantum electrodynamics via radiation reaction, where a particle's backreaction to its radiation field manifests itself in an additional force\cite{Cole.2018wy7,Poder.2018,blackburn2020radiation}. Recent proposals to extend intensities to the Schwinger limit\cite{Quere.2021}, where the electric field strength of the light is comparable to the Coulomb field, could allow the study of novel phenomena  expected to occur due to a breakdown of perturbation theory. In an only slightly less extreme case, high-energy density physics (HEDP)\cite{Drake.2009} research uses lasers for the production and study of states of matter that cannot be reached otherwise in terrestrial laboratories. This includes creating and investigating material under extreme pressures and temperatures, leading to exotic states like warm-dense matter\cite{Mahieu.2018, Kettle.2019vmn,Hussein2022PRL}.

Apart from the fundamental interest, there is also considerable interest in developing novel applications that are enabled by these laser-plasma interactions. Two particularly promising application areas have emerged over the past decades, namely the production of high-energy radiation beams (electrons, positrons, ions, X-rays, gamma-rays) and laser-driven fusion.

Laser-plasma acceleration (LPA) aims to accelerate charged particles to high energies over short distances by inducing charge separation in the plasma, e.g. in the form of plasma waves to accelerate electrons or by stripping electrons from thin-foil targets to accelerate ions. The former scenario is called laser wakefield acceleration (LWFA)\cite{Esarey.2009, Malka.2012, Hooker.2013}. Here a high-power laser propagates through a tenuous plasma and drives a plasma wave which can take the shape of a spherical ions cavity directly behind the laser. The fields within this so-called bubble typically reach around 100 GV/m, allowing LWFA to accelerate electrons from rest to GeV energies within centimeters \cite{Leemans.2006, Kneip2009PRL, Clayton2010PRL, Wang2013NC, Leemans2014PRL, Gonsalves.2019}. While initial experiments were single-shot in nature and with significant shot-to-shot variations, the performance of LWFA has drastically increased in recent years. 
Particularly worth mentioning in this regard are the pioneering works on LWFA at kHz-level repetition rate, starting with an early demonstration in 2013\cite{he2013high} and in the following mostly developed by J. Faure et al.\cite{Faure.2018,rovige2020demonstration} and H. Milchberg's group\cite{salehi2021laser}, as well as the day-long stable operation achieved by A. Maier et al.\cite{Maier.2020}.
As a result of these efforts, typical experimental data sets in publications have significantly increased in size. To give an example, first studies on the so-called beamloading effect in LWFA were done on sets of tens of shots\cite{Rechatin.2010}, whereas newer studies include hundreds of shots\cite{Gotzfried.2020} or most recently thousands of shots\cite{Kirchen2021PRL}. 

Laser-driven accelerators can also function as bright radiation sources via the processes of bremsstrahlung emission\cite{Glinec.2005, Dopp.2016qp5}, betatron radiation\cite{Rousse.2004, Kneip.2010ysi} and Compton scattering\cite{Phuoc.2012, Powers.2014, Khrennikov.2015}. These sources have been used for a variety of proof-of-concept applications\cite{Albert.2016}, ranging from spectroscopy studies of warm-dense matter\cite{mahieu2018probing}, over imaging\cite{kneip2011x,fourmaux2011single} to X-ray computed tomography\cite{wenz2015quantitative,Cole.2015qcp,Cole.2018,guenot2022distribution}.
It has also recently been demonstrated that LWFA can produce electron beams with sufficiently high beam quality to drive free-electron lasers (FELs) \cite{Wang2021N,Labat.2022}, offering an potential alternative driver for next generation light sources \cite{Emma2022arxiv}.

Laser-ion acceleration\cite{Daido.2012, Macchi.2013} uses similar laser systems as LWFA, but typically operates with more tightly-focused beams to reach even higher intensities. Here the goal is to separate a population of electrons from the ions and then use this electron cloud to strip ions from the target. The ions are accelerated to high energies by the fields that are generated by the charge separation process. This method has been used to accelerate ions to energies of a few tens of MeV/u in recent years. In an alternative scheme, radiation pressure acceleration\cite{Robinson.2008, Henig.2009}, the laser field is used to directly accelerate a target. Even though it uses the same or similar lasers, ion acceleration typically operates at much lower repetition rate because of its thin targets which are not as easily replenished as gas based plasma sources used in LWFA. Recent target design focuses on the mitigation of this issue, for instance using cryogenic jets\cite{Fraga2012RSI,gauthier2016high, Kraft2018FirstTarget, rehwald2022towards} or liquid crystals \cite{Schumacher2017LiquidAcceleration}.

\begin{figure*}
    \centering
    \includegraphics[width=.90\linewidth]{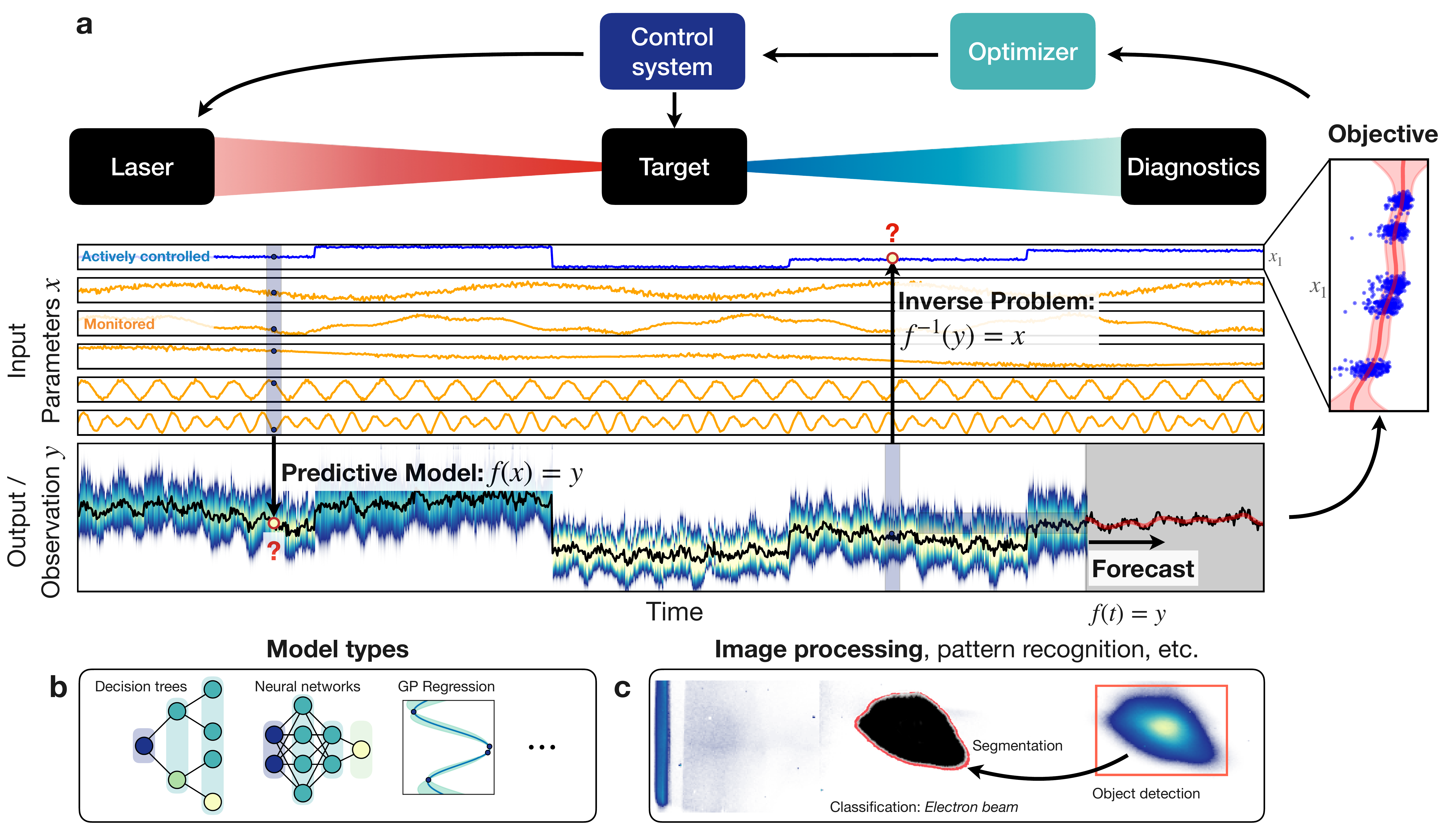}
    \caption{\textbf{Overview of some of the machine learning applications discussed in the manuscript}. (a) General configuration of laser-plasma interaction setups, applicable to both experiments and simulations. The system will have a number of input parameters of the laser and target. Some of these are known and actively controlled (e.g. laser energy, plasma density, etc.), some are monitored, and others are unknown and essentially contribute as noise to the observations. Predictive models take the known input parameters $x$ and use some model to predict the output $y$. These models are discussed in \cref{sec:pred_models} and some of them are sketched in (b). Inversely, in some case one will want to derive the initial conditions from the output. These inverse problems are discussed in \cref{scn:inverse_problems}. In other cases one might be interested in a temporal evolution, discussed in \cref{forecasting}. The output from observations or models can be used to optimize certain objectives, which can then be fed back to the control system to adjust the input parameters, see \cref{optimization}. Observations may also require further processing, e.g. image processing in (c) to detect patterns or objects. Note that the sub-figure (a) is for illustrative purposes only and based on synthetic data.}
    \label{fig:overview}
\end{figure*}

Another application of potentially high societal relevance is laser-driven inertial confinement fusion (ICF)\cite{Moses.2009, Betti.2016}, where the aim is to induce fusion by heating matter to extremely high temperatures through laser-plasma interaction. As the name suggests, confinement is reached via the inertia of the plasma, which is orders of magnitude larger than the thermal energy. To achieve spatially homogenous heating that can penetrate deep into the fusion target, researchers commonly resort to driving the ignition process indirectly. In this method, light is focused into an empty cylindrical cavity, called a hohlraum, which is used to radiate a nearly isotropic blackbody spectrum that extents into the X-ray regime and is subsequently absorbed by the imploding capsule\cite{Murakami.1991}. Opposed to this, direct-drive methods aim to directly drive the thermonuclear fusion process\cite{Craxton.2015}. In this case, the laser is focused directly onto the fuel capsule. Direct-drive poses some challenges that are not present in indirect-drive schemes. For instance, the light has to penetrate through the high-density plasma shell surrounding the capsule and the illumination is less homogeneous, because of which the compressed target can be subject to large hydrodynamic instabilities. Advanced ignition schemes aim to separate compression of the thermonuclear fuel from triggering the ignition process. Examples are fast ignition\cite{Kodama.2001}, which uses the high-intensity laser pulse to directly heat the compressed and dense fusion target, or shock ignition\cite{Betti.2007}, which uses a shock wave to compress the target. Recently, a first ICF experiment at the National Ignition Facility has reported reaching the burning-plasma state via combination of indirect-drive with advanced target design\cite{Zylstra.2022}. This breakthrough has re-enforced scientific and commercial interest in ICF, which is now also pursued by a number of start-up companies.\\

\subsection{Why data-driven techniques?}

In recent years, data-driven methods and machine learning have been applied to a wide range of physics problems\cite{carleo2019machine}, including for instance quantum physics\cite{Dunjko.2018, schutt2020machine}, particle physics\cite{radovic2018machine}, condensed matter physics\cite{bedolla2020machine}, electron microscopy\cite{ede2021deep} and fluid dynamics\cite{kutz2017deep}. In comparison, its use in laser-plasma physics is still in its infancy and is curiously driven by both data \emph{abundance} and data \emph{scarcity}. Regarding the former, fast developments in both laser technology\cite{Leemans.2017} and plasma targetry\cite{prencipe2017targets, george2019high, chagovets2021automation, condamine2021high} nowadays permit operation of laser-plasma experiments - in particular laser-plasma accelerators - at Joule-level energies and multi-Hz to kHz repetition rates \cite{nakamura2017diagnostics, sistrunk2017all, roso2018high, pilar2018characterization, jourdain2021l4n}. The vast amounts of data generated by these experiments can be used to develop data-driven models, which are then employed in lieu of conventional theoretical or empirical approaches, or to augment them. In contrast, laser systems used for inertial fusion research produce MJ-level laser pulses and operate at repetition rates as low as one shot per day. With such a sparse amount of independent experimental runs, data-driven models are used to extract as much information as possible from the existing data or combine them with other information sources such as simulations.

The success of all of the above applications depend on the precise control of a complex nonlinear system. 
In order to optimize and control the process of laser-plasma interaction, it is essential to understand the underlying physics and to be able to \emph{model} the complex plasma response to an applied laser pulse. 
However, this is complicated by the fact that it is a strongly nonlinear, multi-scale, multi-physics phenomenon. 
While analytical and numerical models have been essential tools for understanding laser-plasma interactions, they have several limitations. 
Firstly, analytical models are often limited to low-order approximations and therefore cannot accurately predict the behavior of complex laser-plasma systems. 
Secondly, accurate numerical simulations require immense computational resources, often millions of core hours, which limits their use for optimizing and controlling real-world laser-plasma experiments. 
In addition, the huge range of temporal and spatial scales mean that in practice many physical processes (ionisation, particle collisions, etc.) can only be  treated approximately in large scale numerical simulations.
Because of this, one active area of research is to automatically extract knowledge from data in order to build faster computational models that can be used for \emph{prediction}, \emph{optimization}, \emph{classification} and other tasks. 

Another important problem is that the diagnostics employed in laser-plasma physics experiments typically only provide incomplete and insufficient information about the interaction and key properties must be inferred from the limited set of available observables. Such \emph{inverse problems} can be hard to solve, especially when some information is lost in the measurement process and the problem becomes ill-posed. Modern methods, such as compressed sensing and deep learning are strong candidates to facilitate the solution of such problems and thus retrieve so-far elusive information from experiments.

The goal of this review is to summarize the rapid, recent developments of data-driven science and machine learning in laser-plasma physics, with particular emphasis on laser-plasma acceleration, and to provide guidance for novices regarding the tools available for specific applications. 
We would like to start with a disclaimer that the lines between machine learning and other methods are often blurred\footnote{For instance, deep learning is a subfield of machine learning where deep neural networks are used, and this term is nowadays often used interchangeably with neural networks. Similarly, data-driven science is sometimes used instead of machine learning, but also includes a variety of methods from computer science, applied mathematics and statistics, as well as data science, which is a field in itself. Even within itself, machine learning is a heavily segmented research field, whose community can famously be divided into five ``tribes'', namely symbolists, connectionists, evolutionists, Bayesians and analogizers. Each of these groups has pioneered different tools, all of which have in recent years experienced resurgence in popularity. The arguably most popular branch is connectionism, which focuses on the use of artificial neural networks. However, some challenges seen in laser-plasma physics require different approaches and, for instance, Bayesian optimization has recently drawn considerable research interest. Another line of division is often drawn between supervised and unsupervised methods. While supervised methods are usually trained from known data sets to build a model that can classify new data, unsupervised methods attempt to find some structure in the data without such pre-existing knowledge. Alternatively, one can distinguish between online and batch methods, where the former learn from data as it becomes available, and can therefore be used in an experimental setting, while the latter require access to the full dataset before learning can begin. Yet another important distinction can be made between parametric and non-parametric methods, where the former rely on a set of parameters that is fixed and known in advance, while the latter do not make this assumption, but learn the model parameters from the data.}. 
Given the multitude of often competing subdivisions, we have chosen to organize this review based on a few broad classes of problems, which we believe to have highest relevance for laser-plasma physics and its applications. These problems are modeling \& prediction (\cref{prediction}), inverse problems (\cref{scn:inverse_problems}), optimization (\cref{optimization}), unsupervised learning for data analysis (\cref{scn:correlation}) and last, image analysis with supervised learning techniques (\cref{classification}). Partial overlap between these applications and the tools used is unavoidable and is where possible indicated via cross-references. This is particularly true in the case of neural networks, which have found a broad usage across applications. Each section includes introductions to the most common techniques that address the problems outlined above. We explicitly include what can be considered as ``classical'' data-driven techniques in order to provide a better context for more recent methods. Furthermore, examples for implementations of specific techniques in laser-plasma physics and related fields are highlighted in separate text boxes. We hope that these will help the reader to get a better idea which methods might be most adequate for their own research. An overview of the basic application areas is shown in \cref{fig:overview}.

To maintain brevity and readability, some generalizations and simplifications are made. For detailed descriptions and strict definitions of methods, the reader is kindly referred to the references given throughout the text. Furthermore, we would like to draw the reader's attention to some recent reviews on the application of machine learning techniques in the related fields of plasma physics\cite{Spears2018PoP, anirudh20222022, kambara2022science}, ultrafast optics\cite{Genty.2021}, and high-energy-density physics\cite{Hatfield2021N}.

\section{Modeling \& prediction}\label{prediction}

Many real-life and simulated systems are expensive to evaluate in terms of time, money or other limited resources. 
This is particularly true for laser-plasma physics, which either hinges on the limited access to high-power laser facilities or requires high-performance computing to accurately model the ultra-fast laser-plasma interaction. It is therefore desirable to find models of the system, sometimes-called \textit{digital twins}\cite{Rasheed.2020}, which are comparatively cheap to evaluate and whose predictions can be used for extensive analyses. In engineering and especially in the context of model-based optimization (see \cref{optimization}) such lightweight models are often referred to as \emph{surrogate models}. \emph{Reduced order models} feature less degrees of freedom than the original system, which is often achieved using methods of dimensionality reduction (see \cref{scn:dimensionality_redcution}).

The general challenge in modeling is to find a good approximate $f^*(x)$ for the real system $f(x)$ based on only a limited number of examples $f(x_n) = y_n$, the training data. Here $x_n$ is an $n$-sized set of vector-valued input parameters and $y_n$ are the corresponding outputs. To complicate things further, any real measurement will be subject to noise, so $y_n$ has to be interpreted as a combination of the true value and some random noise contribution. Another complication arises from having imperfect or unrepeatable controllers for the input parameters $x_n$. This can result in having different output values for the same set of input parameters.

The predictive models discussed in \cref{sec:pred_models} below are mostly used to interpolate between training data, whereas the related problem of forecasting (\cref{forecasting}) explicitly deals with the issue of extrapolating from existing data to new, so-far unseen data. In \cref{sec_feedback} we briefly discuss how models can be used to provide direct feedback for laser-plasma experiments.   

\subsection{Predictive models}\label{sec:pred_models}

In this section, we describe some of the most common ways to create predictive models, starting with the ``classic'' approaches of spline interpolation and polynomial regression, before discussing some modern machine learning techniques. 

\begin{figure}
   \centering
   \includegraphics[width=8.5cm]{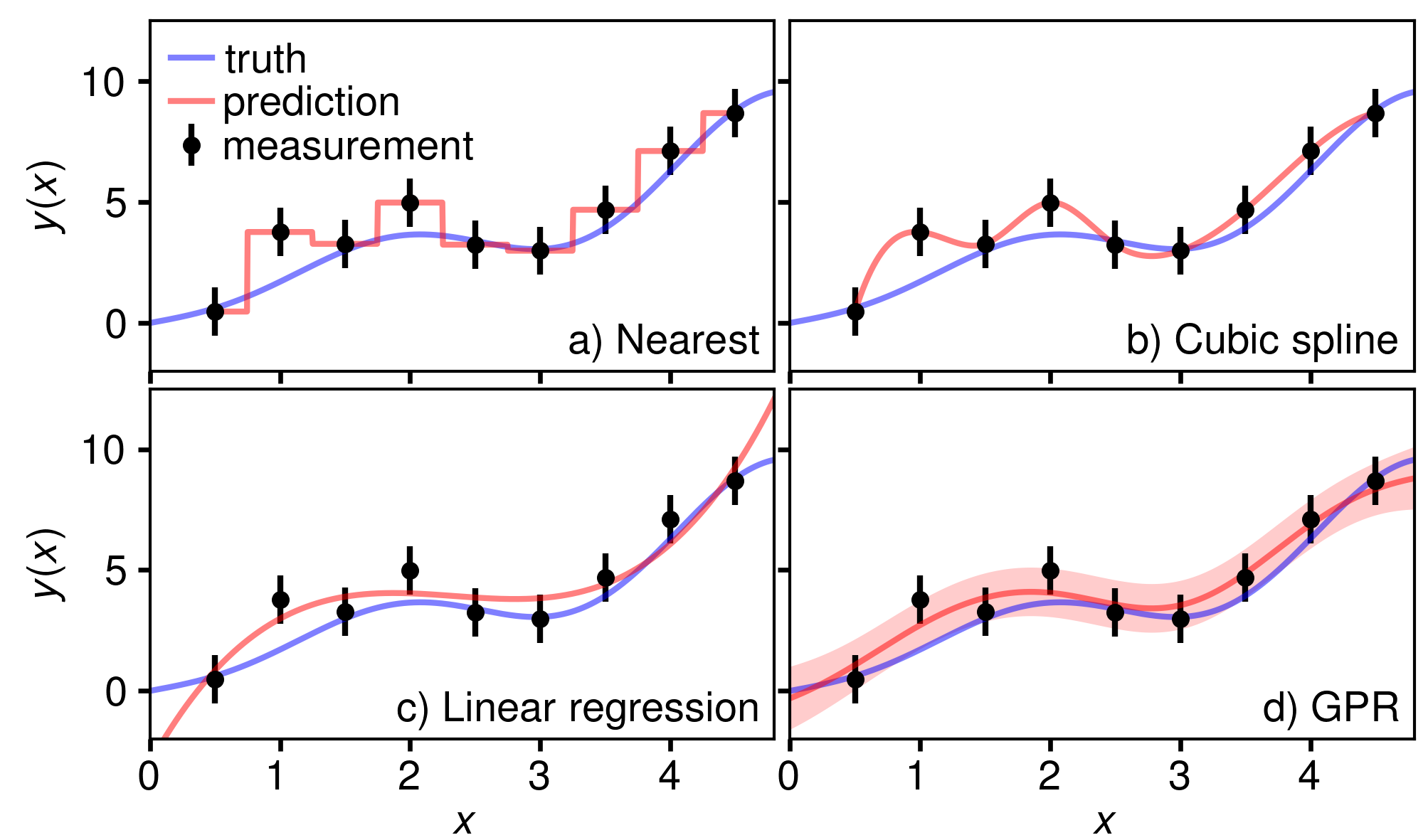}
   \caption{
   \textbf{Illustration of standard approaches to making predictive models in machine learning.}
   The data was sampled from the function $y = x(1+\sin{x}^2) + \epsilon$ with random Gaussian noise, $\epsilon$, for which $\langle \epsilon^2\rangle=1$.
   The data has been fitted by a) nearest neighbour interpolation, b) cubic spline interpolation, c) linear regression of a 3rd order polynomial and d) Gaussian process regression.
   }
   \label{fig:model_comp_examples}
\end{figure}

\subsubsection{Spline Interpolation}

The simplest way of constructing a model for a system, be it a real-world system or some complex simulation, is to use a set of $n$ training points and to predict that every unknown position will have the same value as the nearest neighboring training point (see \cref{fig:model_comp_examples}a). A straightforward extension with slightly higher computational requirements is to connect the training points with straight lines, resulting in a piecewise linear function. Both methods however, are not continuously differentiable, and for instance less suited for the integration of the model into an optimization process (see \cref{optimization}). A more advanced approach to interpolating the training points is to use splines (see \cref{fig:model_comp_examples}b), which require the piecewise-interpolated functions to agree up to a certain derivative order. For instance, cubic splines are continuously differentiable up to the second derivative. While higher-order spline interpolation works well in one-dimensional cases, it becomes increasingly difficult in multi-dimensional settings. Furthermore, the interpolation approach does not allow for incorporating uncertainty or stochasticity, which is present in real-world measurements. Therefore, it will treat noise components as a part of the system, including for instance outliers.

\subsubsection{Regression}

In some specific cases the shortcomings of interpolation approaches can be addressed by using \emph{regression} models. For instance, simple systems can often be described using a polynomial model, where the coefficients of the polynomial are determined via a least-squares fit (see \cref{fig:model_comp_examples}c), i.e. minimising the sum of the squares of the difference between the predicted and observed data. The results are generally improved by including more terms in the polynomial but this can lead to overfitting – a situation where the model describes the noise in the data better than the actual trends, and consequently will not generalise well to unseen data. Regression is not restricted to polynomials, but may use all kinds of mathematical models, often motivated by a known, underlying physics model. 
Crucially, any regression model requires the prior definition of a function to be fitted, thus posing constraints on the relationships that can be modeled. 
In practice this is one of the main problems with this approach, as complex systems can scarcely be described using simple analytical models. Before using these models, it is thus important to verify their validity, e.g. using a measure for the goodness of fit such as the correlation coefficient $R^2$, the $\chi^2$ test or the mean-square error (MSE).

\subsubsection{Probabilistic models}\label{probabilistic}

The field of probabilistic modeling relies on the assumption that the relation between the observed data and the underlying system is stochastic in nature. This means that the observed data are assumed to be drawn from a probability distribution for each set of input parameters to a generative model. Inversely, one can use statistical methods to infer the parameters that best explain the observed data. We will discuss such \emph{inference} problems in more detail in the context of solving (ill-posed) inverse problems in \cref{scn:inverse_probabilistic}.

Probabilistic models can generally be divided into two methodologies, frequentist and Bayesian.
At the heart of the frequentist approach lies the concept of likelihood, $p(y|\theta)$, which is the probability of some outcomes or observations $y$ given the model parameters $\theta$ (see e.g. Ref.\cite{anderson2004model} for an introduction). A model is fitted by \emph{maximum likelihood estimation} (MLE), which means finding the model parameters $\hat{\theta}$ that maximize the likelihood, $p(y|\hat{\theta})$. When observations $y = (y_1,y_2 \dots)$ are independent, the probability of observing $y$ is the product of the probabilities of each observation, $p(y)=\prod_{i=1}^{n} p(y_i)$. As sums are generally easier to handle than products, this is often expressed in terms of the log-likelihood function that is given by the sum of the logarithms of each observation's probability, i.e.
\begin{equation}
\label{eq:log-likelihood-factorization}
\log p(y| \theta)=\sum_{i=1}^{n} \log p(y_i | \theta).
\end{equation}
Probabilities have values between 0 and 1, so the log-likelihood is always negative and, the logarithm being a strictly monotonic function, minimizing the log-likelihood maximizes the likelihood
\begin{equation}
\hat{\theta}_{\text{MLE}}=\underset{\theta}{\arg\max}\{ p(y| \theta)\}=\underset{\theta}{\arg\min} \{\log p(y\mid \theta)\}.
\end{equation}

The optimum can be found using a variety of optimization methods, e.g. gradient descent, which are described in more detail in \cref{optimization}. A simple example is the use of MLE for parameter estimation in regression problems. In case the error $(Ax-y)$ is normally distributed this turns out to be equivalent to the least squares method we discussed in the previous section.

The MLE is often seen as the simplest and most practical approach to probabilistic modeling. One advantage is that it does not require any \emph{a priori} assumptions about the model parameters, but only about the probability distributions of the data. But this can also be a drawback if useful prior knowledge of the model parameters is available. In this case one would turn to Bayesian statistics. Here one assesses information about the probability that a hypothesis about the parameter values $\theta$ is correct given a set of data $x$. In this context the probabilistic model is viewed as a collection of conditional probability densities  $p(\theta| y)$ for each set of observed data $y$, with the aim of finding the \emph{posterior distribution} $p(\theta| y)$, i.e. the probability of some parameters given the data observations. This can be done by applying Bayes' rule,
\begin{equation}
    p(\theta\mid y)=\frac{p(y| \theta)p(\theta)}{p(y)},
    \label{bayes_ rule}
\end{equation}
where $p(y| \theta)$ is known from above as likelihood function and $p(\theta)$ denotes the \emph{prior distribution}, i.e. our \emph{a priori} knowledge about the parameters before we observe any data $y$. 
The denominator, $p(y) = \int p(y| \theta^\prime)\cdot p(\theta^\prime)d\theta^\prime$, is called the evidence and ensures that both sides of \cref{bayes_ rule} are properly normalized by integrating over all possible input parameters $\theta$. Once the posterior distribution is known, we can maximize it and get the \emph{maximum  a posteriori} (MAP) estimate 
    
\begin{equation}
\hat{\theta}_{\text{MAP}}=\underset{\theta}{\arg\max} \{p(\theta| y)\}.
\end{equation}

As mentioned above, a particular strength of the Bayesian approach is that we can encode \emph{a priori} information in the prior distribution. Taking the example of polynomial regression, we could for instance set a prior distribution $p(\theta)$ for the regression coefficients $\theta$ that favors small coefficients, thus penalizing high-order polynomials. Another advantage of using the Bayesian framework is that one can quantify the uncertainty in the model result. This is particularly simple to compute in the special case of Gaussian distributions and their generalization, Gaussian \emph{processes}, which we are going to discuss in the next section.

\begin{figure*}[tb]
   \centering
   \includegraphics[width=.97\linewidth]{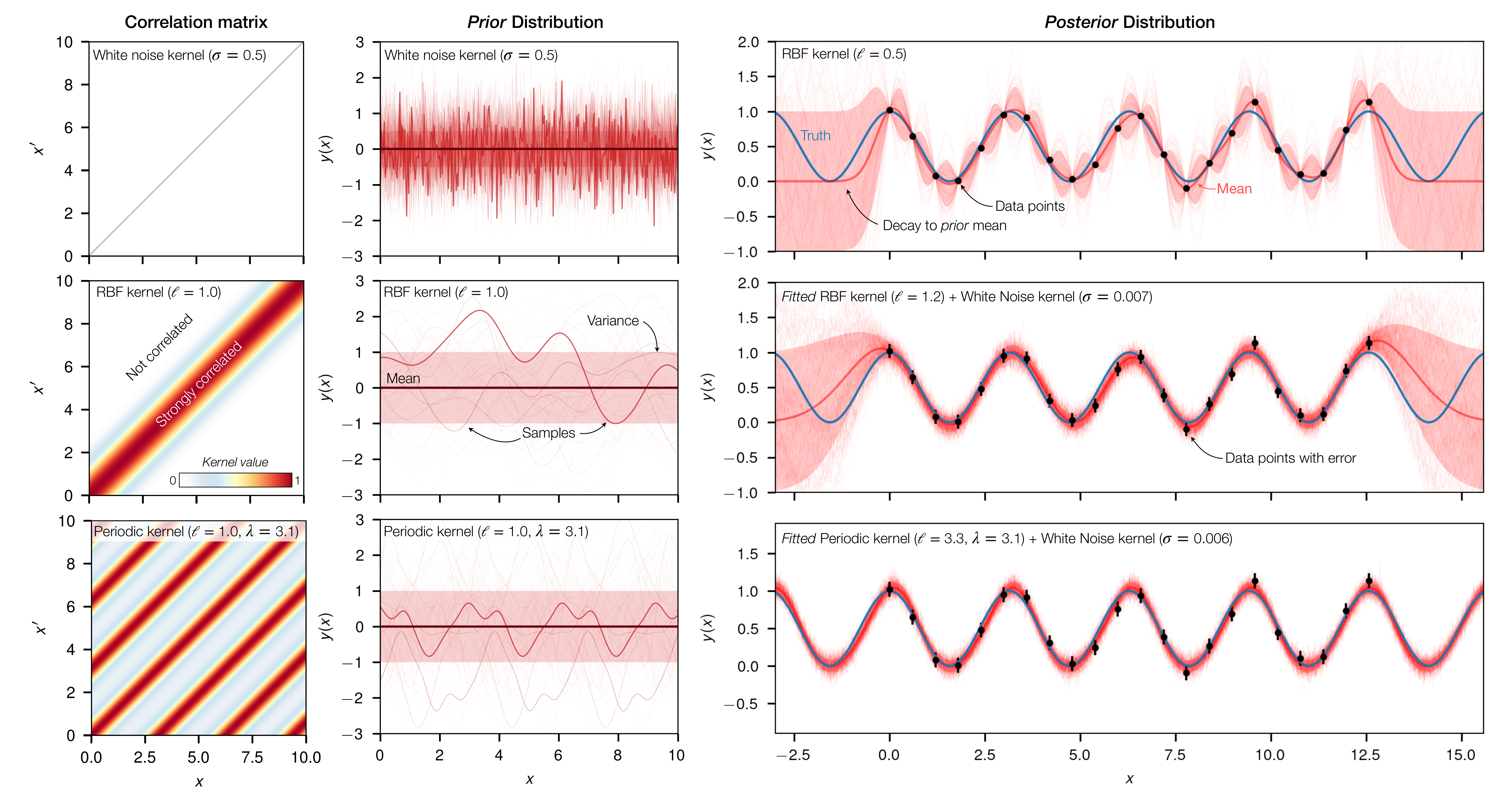}
   \caption{
   \textbf{Gaussian process regression: Illustration of different covariance functions, prior distributions and (fitted) posterior distributions.} \textit{Left:} Correlation matrices between two values $x$ and $x^{\prime}$ using different covariance functions (white noise, radial basis function and periodic). \textit{Center:} Samples of the prior distribution defined by the prior mean $\mu(x)=0$ and the indicated covariance functions. Note that the sampled functions are depicted with increasing transparency for visual clarity. \textit{Right:} Posterior distribution given observation points sampled from $y = \cos^2{x} + \epsilon$, where $\epsilon$ is random Gaussian noise with $\sigma_{\epsilon}=0.1$. Note how the variance between observations increases when no noise term is included in the kernel (top row). Within the observation window the fitted kernels show little difference, but outside of it the RBF kernel decays to the mean $\mu=0$ dependent on the length scale $\ell$. This can be avoided if there exists prior knowledge about the data that can be encoded in the covariance function, in this case periodicity, as can be seen in the regression using a periodic kernel. 
   }
   \label{fig:GP_figure}
\end{figure*}

\subsubsection{Gaussian process regression}\label{GPP}

A popular version of Bayesian probabilistic modeling is Gaussian process (GP) regression \cite{mackay1998introduction,rasmussen2003gaussian}. 
This kind of modeling was pioneered in geostatistics in the context of mining exploration and is historically also referred to as \textit{kriging}, after the South African engineer Danie G. Krige, who invented the method in the 1950s. Conceptually, one can think of it as loosely related to the spline interpolation method, as it is also locally ``interpolates" the training points. Compared to splines and conventional regression methods, kriging has a number of advantages. Being a regression method, kriging can deal with noisy training points as seen in experimental data. At the same time, the use of Gaussian processes involves minimal assumptions and can in principle model any kind of function. Last, the ``interpolation" is done in a probabilistic way, i.e. a probability distribution is assigned to the function values at unknown positions (see \cref{fig:model_comp_examples}d). This allows for quantifying the uncertainty of the prediction. These features make kriging an attractive method for the construction of surrogate models for complex systems, for which only a limited number of the function evaluations is possible, e.g. due to the long runtime of the system or the high costs of the function evaluations. GP regression forms the backbone of most implementations of Bayesian optimization, which we will discuss in \cref{BO}, including examples for potential use cases.

Mathematically speaking a Gaussian process is an infinite collection of normal random variables, where each finite subset is jointly normally distributed. The mean vector and covariance matrix of the multivariate normal distribution are thereby generalised to the mean function \(\mu(x)\) and the covariance function \(\sigma(x,x')\) respectively, where we use the short-hand notation \(x=(x_1,x_2,\dots)\) to denote the function inputs as a vector of real numbers.

A Gaussian process can be written in the form:
\begin{equation}
    \label{GP}
    f(x) \sim \mathcal{GP}(\mu(x),\sigma(x,x')),
\end{equation}
denoting that the random function $f(x)$ follows a Gaussian process with mean function $\mu(x)$ and covariance function $\sigma(x,x')$. 

The mean function $\mu(x)$ is defined as the expectation of the GP, i.e. $\mu(x) = \langle f(x)\rangle$, whereas the covariance function is defined as $\sigma(x,x') = \langle f(x)-\mu(x)(f(x')-\mu(x'))\rangle$. Note that in the special case of constant mean function $\mu(x) = 0$ and a constant diagonal covariance function $\sigma(x,x') = \sigma^2\delta(x-x')$, the Gaussian process simply reduces to a set of a normal random variable with zero mean and variance $\sigma^2$ commonly referred to as \textit{white noise}. The covariance function is also referred to as the \textit{kernel}. Its value at locations  $x$ and $x'$ is proportional to the correlation between the function values \(f(x)\) and \(f(x')\). A common choice for the covariance function is the squared exponential function, also known as the radial basis function (RBF) kernel
\begin{equation}
    \label{SE}
    \sigma(x,x^{\prime}) = \exp\left(-\frac{(x-x')^2}{2\ell^2}\right),
\end{equation}
where $\ell$ is the length scale parameter, which controls the rate at which the correlation between $f(x)$ and $f(x')$ decays. This kernel hyperparameter can usually be optimized by using the training data to minimize the log marginal likelihood. 

It is possible to encode prior knowledge by choosing a specialized kernel that imposes certain restrictions on the model. A variety of such kernels exist. For instance, the periodic kernel (also known as exp-sine-square kernel) given by
\begin{equation}
        \sigma(x,x^{\prime}) = \exp\left(-
        \frac{ 2\sin^2(\pi d(x,x^{\prime})/\lambda) }{ \ell^ 2} \right),
        \label{eq:periodic-kernel}
\end{equation}
with the Euclidean distance $d(x,x^{\prime})$, length scale $\ell$ and periodicity $\lambda$ as free hyperparameters, is particularly suitable to model systems that show an oscillatory behavior. 

A useful property of kernels is the generation of new kernels through addition or multiplication of existing kernels \cite{genton2001classes}. This property provides another way to leverage prior information about the form of the function to increase the predictive accuracy of the model. For instance, measurement errors incorporated into the model by adding a Gaussian white noise kernel, as for instance done in \cref{fig:model_comp_examples}.

\cref{fig:GP_figure} visualizes the covariance functions and their influence on the result of Gaussian process regression. This includes correlation matrices for the three covariance functions discussed above (white noise, RBF and periodic kernels). Once the mean and the covariance functions are fully defined, we can use training points for regression, i.e. fit the Gaussian process and kernel parameters to the data and obtain the \emph{posterior distribution}, see right panel of \cref{fig:GP_figure}.

\begin{figure}
    \centering
    \includegraphics[width=\linewidth]{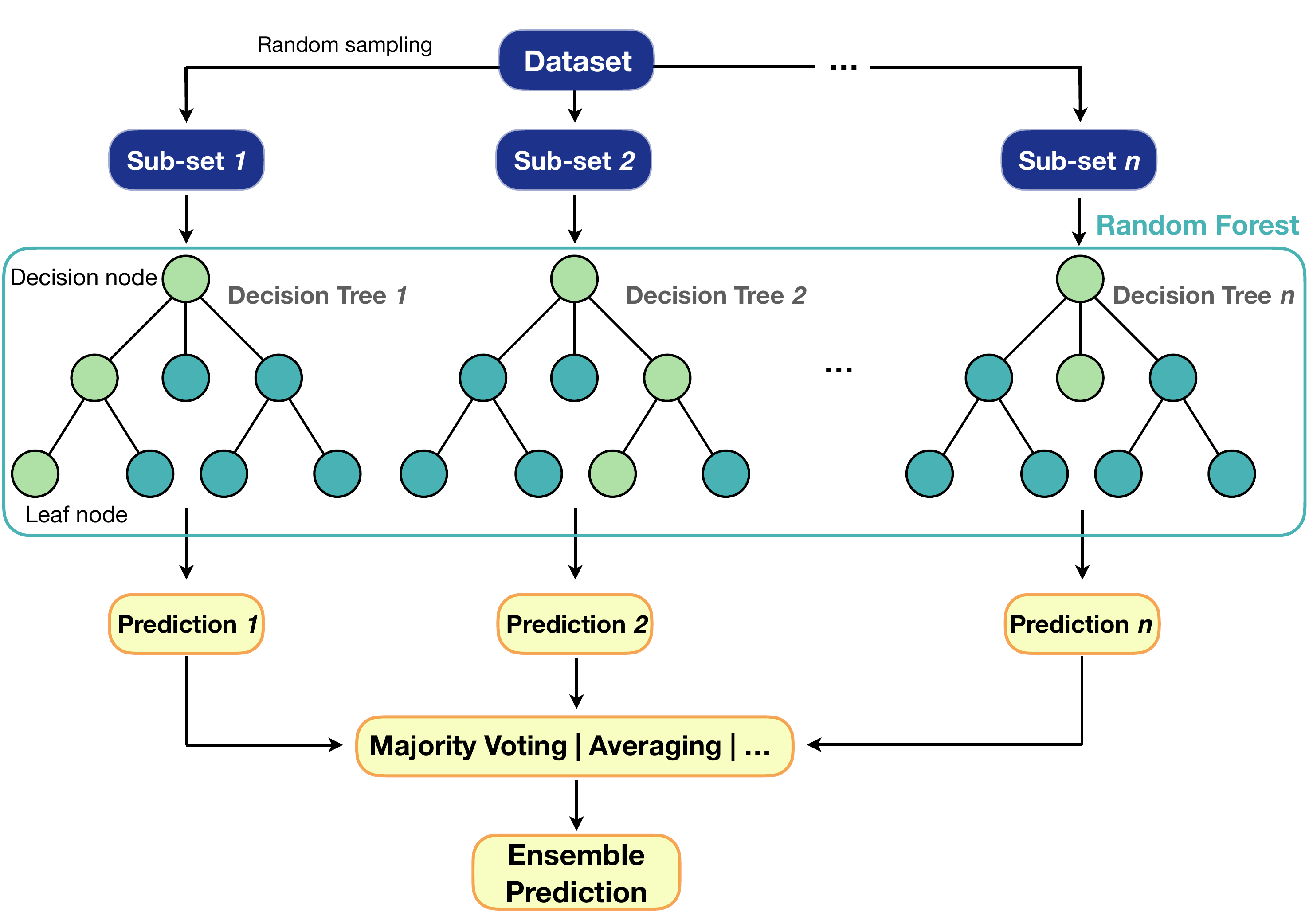}
    \caption{\textbf{Sketch of a random forest}, an architecture for regression or classification consisting of multiple decision trees, whose individual predictions are combined using into an ensemble prediction e.g. via majority voting or averaging.}
    \label{fig:randomForest}
\end{figure}

\begin{figure}
    \centering
    \includegraphics[width=\linewidth]{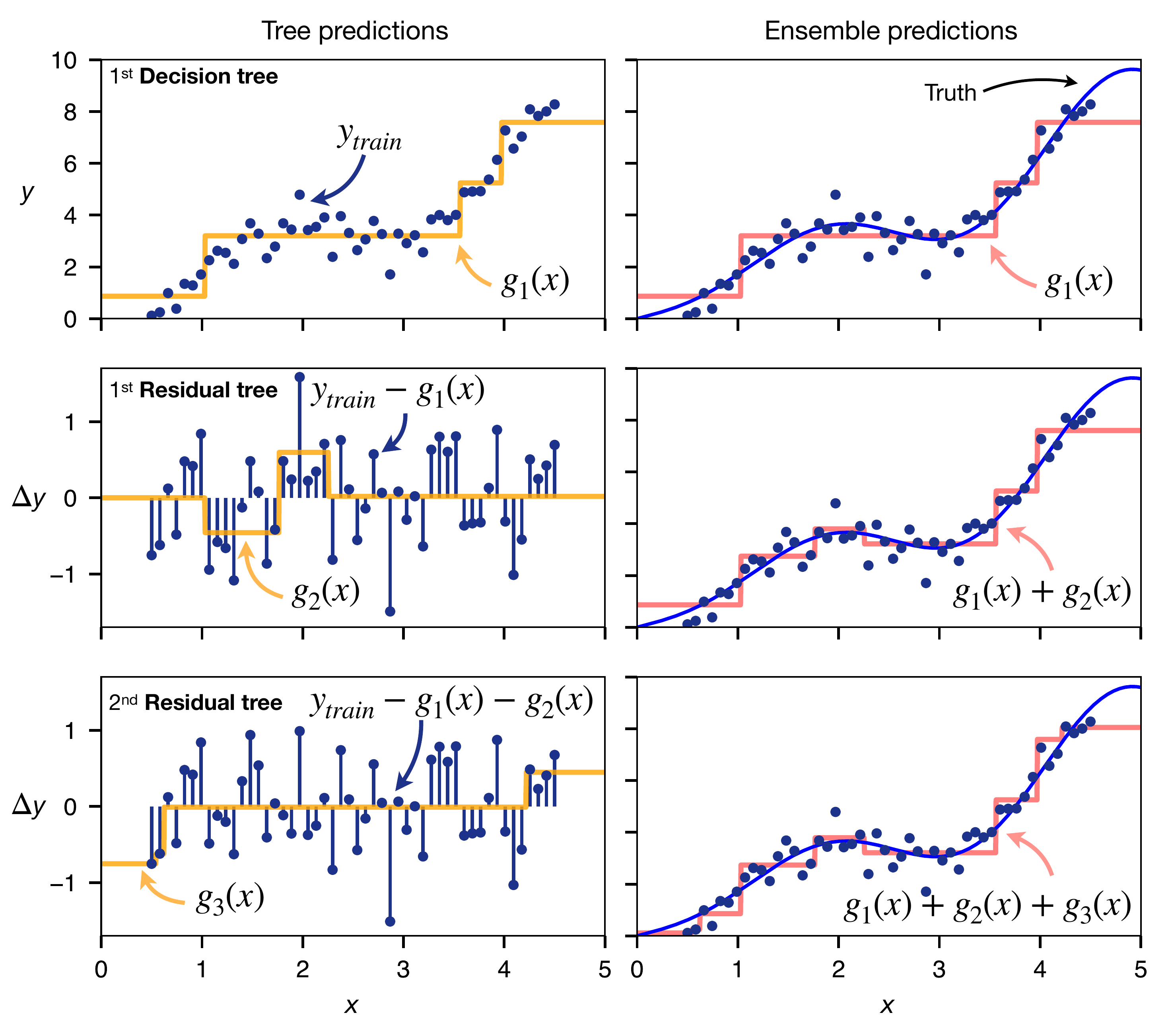}
    \caption{\textbf{Example of gradient boosting with decision trees}. First, a decision tree $g_1$ is fitted to the data. In the next step, the residual difference between training data and the prediction of this tree is calculated and used to fit a second decision tree $g_2$. This process is repeated $n$ times, with each new tree $g_n$ learning to correct only the remaining difference to the training data. Data in this example sampled from same function as in \cref{fig:model_comp_examples} and each tree has a maximum depth of two decision layers.}
    \label{fig:gradientBoost}
\end{figure}

\subsubsection{Decision trees and forests}\label{trees}

A decision tree is a general-purpose machine learning method that learns a tree-like model of decision rules from the data to make predictions\cite{breiman2017classification}. It works by splitting the dataset recursively into smaller groups, called \emph{nodes}. Each node represents a decision point, and the tree branches out from the node according to the decisions that are made. The \emph{leaves} of the tree represent the final prediction. This can be either a categorical value in classification tasks (see \cref{classification}) or a numerical value prediction in regression tasks. An advantage of this approach is that it can learn non-linear relationships in data without having to specify them.

To generate a decision tree one starts at the root of the tree and determines a decision node such that it optimizes an underlying decision metric like the mean squared error in regression settings or entropy and information gain in a classification setting. At each decision point the data set is split and subsequently the metric is re-evaluated for the resulting groups, generating the next layer of decision nodes. This process is repeated until the leaves are reached. The more layers decision layers are used, called the depth of the tree, the more complex relationships can be modeled.

Decision trees are easy to implement and can provide accurate predictions even for large data sets. However, with increasing number of decision layers they may become computationally expensive and may overfit the data, the latter being in particular a problem with noisy data. One method to address overfitting is called \emph{pruning}, where branches from the tree that do not improve the performance are removed. Another effective method is to use decision-tree-based \emph{ensemble} algorithms instead of a single decision tree. 

One example of such algorithm is the random forest, an ensemble algorithm that uses bootstrap aggregating or \emph{bagging} to fit trees to random subsets of the data and the predictions of individually fitted decision trees are combined by majority vote or average to obtain a more accurate prediction. Another type of ensemble algorithms is \emph{boosting}, where the trees are trained sequentially and each tries to correct its predecessor. A popular implementation is AdaBoost\cite{Freund.1999} where the weights of the samples are changed according to the success of the predictions of the previous trees. Gradient boosting methods\cite{Friedman.2001,Friedman.2002} also use the concept of sequentially adding predictors, but while AdaBoost adjust weights according to the \emph{residuals} of each prediction, gradient boosting methods fit new residual trees to the remaining differences at each step (see \cref{fig:gradientBoost} for an example).

Compared to other machine learning algorithms, a great advantage of a decision tree is its explicitness in data analysis, especially in nonlinear high-dimensional problems. By splitting the dataset into branches, the decision tree naturally reveals the importance of each variable regarding the decision metric. This is in contrast to ``black-box'' models, such as those created by neural networks, which must be interpreted by post hoc analysis.

\example{Decision trees can also be used to seed neural networks by giving the initial weight parameters to train a deep neural network. An example of using decision tree as an initializer are Deep Jointly-Informed Neural Networks (DJINN) developed by Humbird \textit{et al.}\cite{humbird2018deep}, which have been widely applied in the high power laser community, especially in analyzing inertial confinement fusion datasets. The algorithm first constructs a tree or a random forest with tree depth set as a tunable hyperparameter. It then maps the tree to a neural network, or maps the forest to an ensemble of networks. The structure of the network (number of neurons and hidden layer, initial weights, etc.) reflects the structure of the tree. The neural network is then trained using back-propagation. The use of decision trees for initialization largely reduces the computational cost while maintaining comparable performance to optimized neural network architectures. The DJINN algorithm has been applied to several classification and regression tasks in high-power laser experiments such as ICF\cite{humbird2019transfer, hsu2020analysis, kluth2020deep} and LWFA\cite{Lin2021PoP}.
}

\begin{figure*}
    \centering
    \includegraphics[width=.95\linewidth]{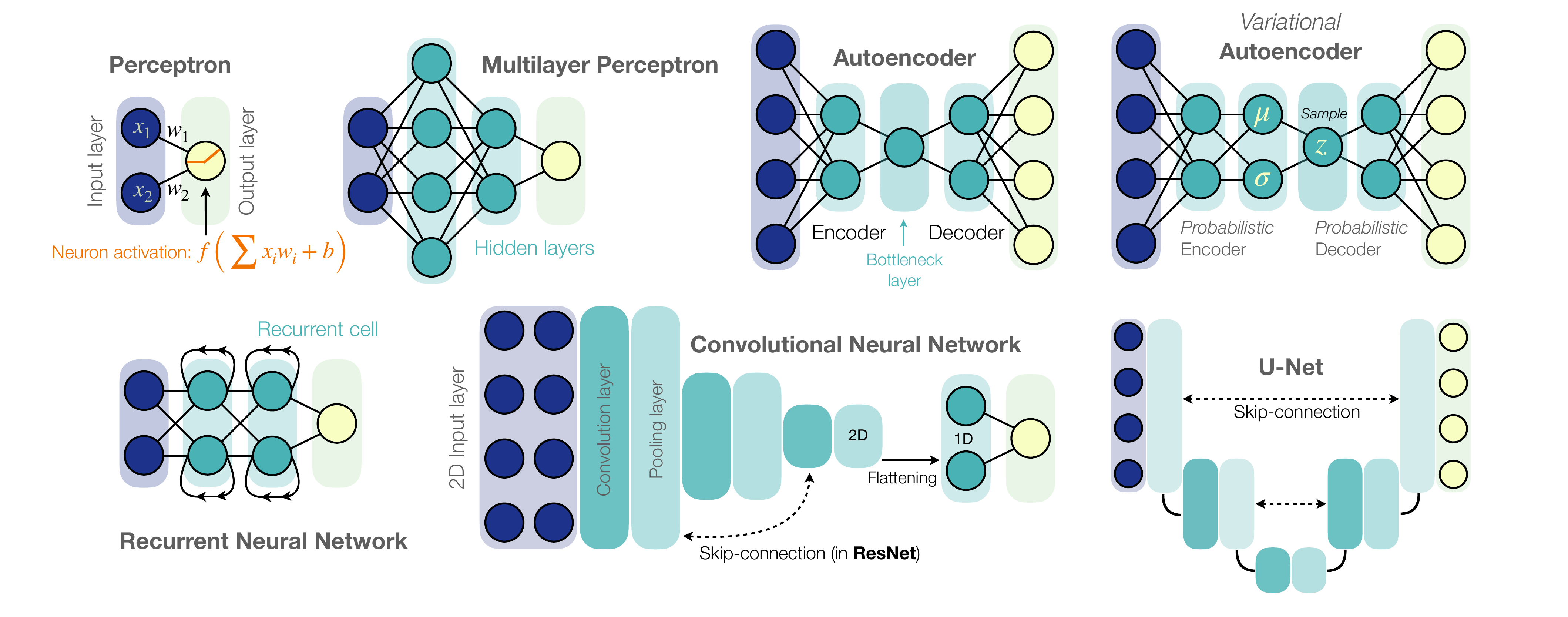}
    \caption{\textbf{Simplified sketch of some popular neural network architectures.} The simplest possible neural network is the perceptron, which consists of an input, which is fed into the neuron that processes the input based on the weights, an individual bias and its activation function. Multiple such layers can be stacked within so-called hidden layers, resulting in the popular multilayer perceptron (or fully-connected network). Besides the direct connection between subsequent layers, there are also special connections common in many modern neural network architectures. Examples are the recurrent connection (which feeds the output of the current layer back into the input of the current layer), the convolutional connection (which replaces the direct connection between two layers by the convolutional operation) or the residual connection (which adds the input to the output of the current layer; note that the above illustration is simplified and the layers should be equal in size).}
    \label{fig:neuralNet}
\end{figure*}

\subsubsection{Neural networks}\label{neural_nets}

Neural networks offer a versatile framework for fitting of arbitrary functions by training of a network of connected nodes or \emph{neurons}, which are loosely inspired by biological neurons.
In the case of a fully connected neural network, historically called a multilayer perceptron, the inputs to one node are the outputs of all of the nodes from the preceding layer.
The very first layer is simply all of the inputs for the function to be modelled, and the very last layer is the function outputs.
One of the very attractive properties of neural networks is the capacity to have many outputs and inputs, each of which can be multi-dimensional.
The weighting of each connection $w_i$ and the bias for each node $b$ are the parameters of the network, which must be trained to provide the best approximation of the function to be modelled. Each node gets activated depending on both weights and biases according to a pre-defined \emph{activation function}. The non-linear properties of activation functions are widely seen as the key property to allow neural networks to model arbitrary functions and achieve general learning properties.

One of the simplest, but also most common activation functions is the rectified linear unit (ReLU), which is defined as
\begin{equation}
  \text{ReLU}(x) = 
  \begin{cases}
    x, & \text{if } x \ge 0\\
    0, & \text{otherwise.}
  \end{cases}
\end{equation}
The function argument $x=\sum x_i w_i +b$ is the sum of the weights of incoming connections, multiplied with their values $x_i$, and the bias of the node. The ReLU function is easy to compute, and it has the main advantage that it is linear for $x > 0$, which greatly simplifies the gradient calculation in training (see next paragraph). Drawbacks are that it is not differentiable at $x = 0$, it is unbounded, and since it yields a constant value below $0$, it has the potential to produce ``dead'' neurons. The last issue is solved in the so-called \emph{leaky} ReLU, which uses a reduced gradient for $x<0$, giving an output of $\alpha x$ where $0<\alpha<1$. 
Other common activation functions include the logistic or sigmoid function, $\mbox{sig}(x)= ({1 + e^{-x}})^{-1}$, and the 
hyperbolic tangent function function, $\tanh(x)=({e^x-e^{-x}})/({e^x+e^{-x}})$.

The training is performed iteratively by passing corresponding input-output pairs to the network and comparing the true outputs to those given by the network to calculate the \emph{loss function}. This is often chosen to be the mean squared error (MSE) between the model output and the reference from the training data, but many other types of loss functions are also used (see \cref{scn:objective_functions}) and the choice of loss function strongly affects the model's training. The loss is then used to modify the weights and biases via an algorithm known as ``back-propagation''\cite{backpropagation}. Here the gradient of the loss function is calculated with respect to the weights and biases via the chain rule. One can then optimize the parameters using for instance stochastic gradient descent (SGD) or Adam (Adaptive moment estimation) \cite{Kingma2014arxiv}. 

A number of hyperparameters are used to control the training process. Typical ones include: the number of epochs - the number of times the model sees each data point; the 
batch size - the number of data points the model sees before updating its gradients; and the learning rate - a factor determining the magnitude of the update to the gradient. Each factor can have a critical effect on the training of the model. 

In the training process, the weights and biases are optimized in order to best approximate the function for converting the the inputs into the corresponding outputs.
The resulting model will yield an approximation for the unknown function $f(x)$. However, learning via this method only optimizes how well the model reproduces the training data and, in the worst case, the network essentially ``memorizes'' the training data, and learns little about the desired function $f(x)$. To quantify this, a subset of the of data is kept separate and used solely for validating the model.
Ideally, the loss should be similar on both training and validation data, and if the model has significant smaller loss on the training data than the validation set it is said to overfit, which signifies that the model has learnt the random noise of the training set. The validation loss thus quantifies how well this model generalizes and with it, how useful it is for predictions on unseen data. 

Common methods to combat overfitting include early stopping, i.e. terminating the training process once the validation loss stagnates; or the use of dropout layers to randomly switch off some fraction of the network connections for each batch of the training process, thereby preventing the network from learning random noise by reducing its capacity. A further approach is to incorporate variational layers into the network.
In these layers, pairs of nodes are used which represent the mean $\mu$ and standard deviation $\sigma$ of a Gaussian distribution. 
During training, output values are sampled from this distribution and then passed on the subsequent layers, thereby requiring the network to react smoothly to these small random variations to achieve a small training loss.
Both dropout layers and variational layers provide \emph{regularization} which smooths the network response and prevents single nodes from dominating, resulting in improved interpolation performance and better validation loss \cite{Kristiadi2020PMLR,Wager2013ANIPS} (see \cref{sec:regularization} for a brief, general explanation of regularization).

Beyond the classical multilayer perceptron network, there exists a plethora of different neural network architectures, as partially illustrated in \cref{{fig:neuralNet}}, which are suited to different tasks.
In particular, convolutions layers, which extract relevant features from 1D and 2D by learning suitable convolution matrices, are commonly used in the analysis of physical signals and images.
Architecture selection and hyperparameter tuning is a central challenge in the implementation of neural networks, and is often performed by an additional machine-learning algorithm, e.g. using Bayesian optimization (see \cref{BO}).

The great strength of neural networks is their flexibility and relatively straightforward implementation, with many openly accessible software platforms available to choose from.
However, trained networks are effectively ``black-box'' functions, and do not, in their basic form, incorporate uncertainty quantification.
As a result, the networks may make over-confident predictions about unseen data while giving no explanation for those predictions, leading to false conclusions.
Various methods exist for incorporating uncertainty quantification into neural networks (see for example \cite{Gawlikowski2021arxiv}), such as by including variational layers (discussed above) and training an ensemble of networks on different training data subsets.
There are several approaches to try and make neural network models explainable and a review of methods for network interpretation is for instance given by Montavon \emph{et al.}\cite{Montavon.2018}.

Another strength of neural networks is that the performance of a model on a new task can be improved by leveraging knowledge from a pre-trained model on a related task. This so-called \emph{transfer learning} is typically used in scenarios where it is difficult or expensive to train a model from scratch on a new task, or when there is a limited amount of training data available. For example, transfer learning has been used to successfully train models for image classification and object detection tasks (see \cref{classification}), using pre-trained models that have been trained on large image datasets such as ImageNet \cite{Huh2016arxiv}. In the context of laser-plasma physics, transfer learning could be used to improve the performance of a neural network by leveraging knowledge from a pre-trained model that has been trained on data from previous experiments or simulations.

\begin{figure*}
    \centering
    \includegraphics[width=.9\linewidth]{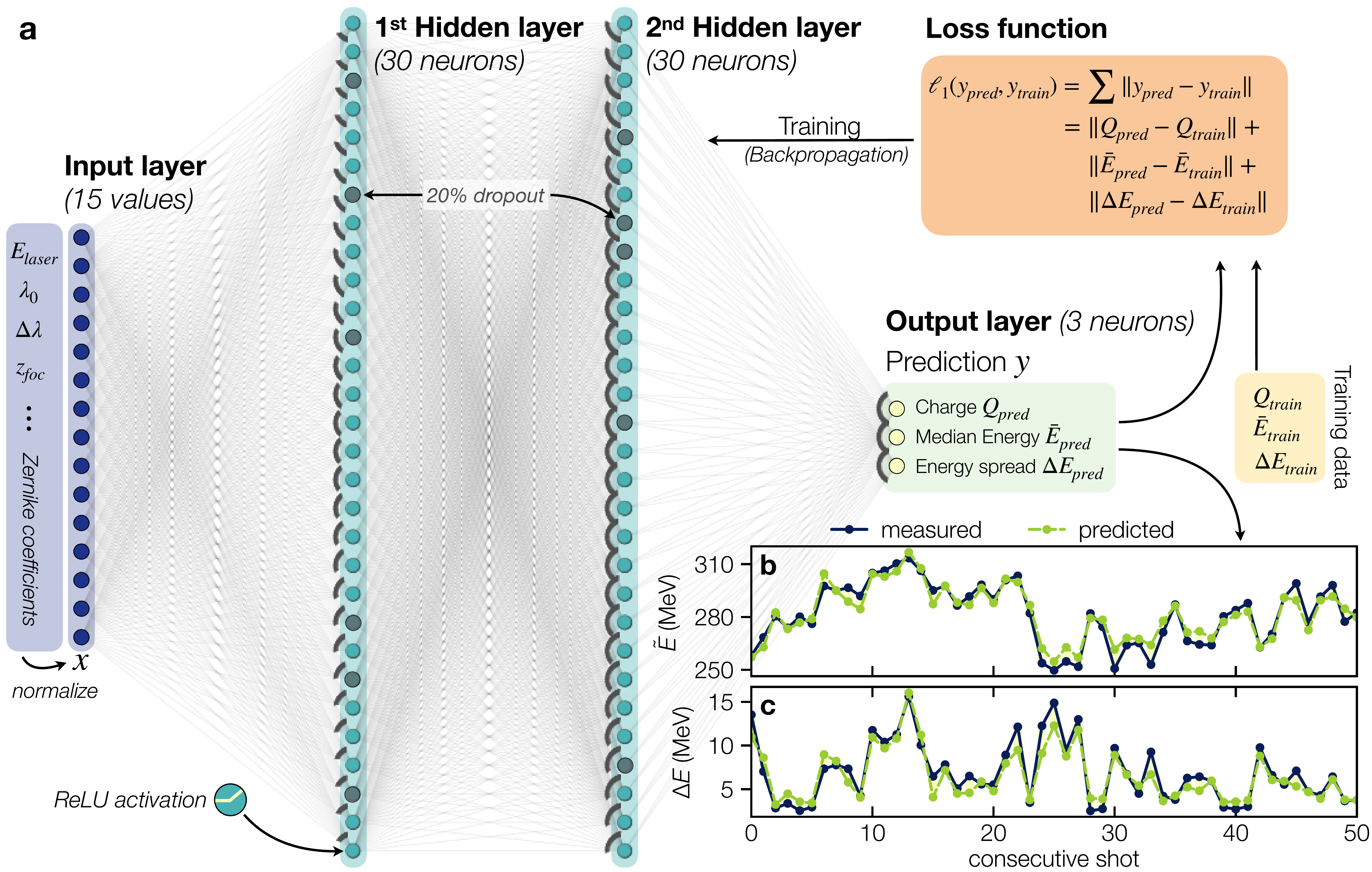}
    \caption{\textbf{Real-world example of a multilayer perceptron for beam parameter prediction.} (a) The network layout\cite{Kirchen2021PRL} consists of 15 input  neurons, two hidden layers with 30 neurons and 3 output neurons (charge, mean energy, and energy spread). The input is derived from parasitic laser diagnostics (laser pulse energy $E_{laser}$, central wavelength $\lambda_0$ and spectral bandwidth $\Delta \lambda$, longitudinal focus position $z_{foc}$ and Zernike coefficients of the wavefront). Neurons use a non-linear ReLU activation and 20\% of neurons drop out for regularization during training. The (normalized) predictions are compared to the training data to evaluate the accuracy of the model, in this case using the mean absolute $\ell_1$ error as loss function. In training the gradient of the loss function is then propagated back through the network to adjust its weights and biases. (b)  Measured and predicted median energy ($\bar E$) and (c) measured and predicted energy spread ($\Delta$E), both for a series of 50 consecutive shots. Subfigures b-c adapted from Kirchen \emph{et al.}\cite{Kirchen2021PRL}}
    \label{fig:Kirchen}
\end{figure*}

\example{Kirchen \emph{et al.} \cite{Kirchen2021PRL} recently demonstrated the utility of even seemingly small neural networks for modeling laser-plasma accelerator performance. Their multilayer perceptron design is shown in \cref{fig:Kirchen} and as can been seen in \cref{fig:Kirchen}b-c, the network accurately predicts electron beam energy and energy spread. Interestingly this performance is achieved without using target parameters such as plasma density as an input, thus highlighting the relative importance of laser stabilization in this context.
Gonoskov \emph{et al.}\cite{gonoskov2019employing} trained a neural network to read features in theoretical and experimental spectra from high-order-harmonic generation (HHG) in high-power laser-plasma interactions. Rodimkov \emph{et al.}\cite{rodimkov2021towards} used a neural network to extract information that was not directly measured in experiments, including the preplasma scale length and the pulse carrier-envelope phase, from the spectrum of XUV radiation generated in laser-plasma interactions. Another recent example of deep learning modeling is the work by Djordjević \emph{et al.} \cite{Djordjevic2021PoP}, where the authors used a multilayer perceptron to model the output of a laser-driven ion accelerator based on training with 1,000 simulations.
In the work of Watt\cite{Watt2021Thesis}, a strong-field QED model incorporating a trained neural network was used to provide an additional computation package to the Geant4 particle physics platform. 
Neural networks are also trained to assist hohlraum design for ICF experiments by predicting the time evolution of the radiation temperature, in the recent work by McClarren \emph{et al.}\cite{mcclarren2021high}.
In the work by Simpson \emph{et al.}\cite{simpson2021development}, a fully-connected neural network with three hidden layers is constructed to assist the analysis of a x-ray spectrometer, which measures the x-rays driven by MeV electrons produced from high-power laser-solid interaction.
Finally, Streeter \emph{et al.} \cite{Streeter2023HPLSE} used convolutional neural networks to predict the electron spectrum produced by a laser wakefield accelerator, taking measurements from secondary laser and plasma diagnostics as the inputs.}

\subsubsection{Physics-informed machine learning models}\label{scn:PINNs}

The ultimate application of machine learning for modeling physics systems would arguably be to create an ``artificial intelligence physicist'', as coined by Wu and Tegmark\cite{Wu.2019uov}. One prominent idea at the backbone of how we build physical models is Occam's razor, which states that given multiple hypotheses, the simplest one which is consistent with the data is to be preferred. In addition to this guiding principle, it is furthermore assumed that a physical model can be described using mathematical equations. A program should therefore be able to automatically create such equations, given experimental data. While a competitive AI physicist is still years away\footnote{Recently there has also been astonishing progress in the area of large language models such as generative pre-trained transformer (GPT) models\cite{radford2018improving}. Very similar to forecasting networks discussed in \cref{sec:forecasting_networks}, these use an attention mechanism to predict the next token (word, etc.) following input provided by the user. It was found that large models ($\sim 10^{8-9}$ parameters) become increasingly capable with sufficient training, e.g. gaining the ability to do basic math and to write code, including to some claims first hints at artificial general intelligence\cite{bubeck2023sparks}.}, first steps have been undertaken in this direction. For instance, in 2009 Schmidt et al.\cite{Schmidt.2009} presented a genetic algorithm approach (\cref{scn:genetic}) that independently searched and identified governing mathematical representations such as the Hamiltonian from real-life measurements of some mechanical systems. More recently, research has concentrated more on the concept of Occam's razor and the underlying idea that the ``simplest'' representation can be seen as the sparsest in some domain. A key contribution was SINDy (Sparse Identification of Nonlinear Dynamics), a framework for discovering sparse nonlinear dynamical systems from data\cite{Brunton.2016}. As in compressed sensing (\cref{compressed}), the sparsity constraint was imposed using an $\ell_1$-norm regularization, which results in the identification of the fewest terms needed to describe the dynamics.

An important step towards combining physics and machine learning was undertaken in physics-informed neural networks (PINNs)\cite{Karniadakis.2021}. A PINN is essentially a neural network that uses physics equations, which are often described in form of ordinary differential equations (ODEs) or partial differential equations (PDEs), as a regularization to select a solution that is in agreement with physics. This is achieved by defining a custom loss function that includes a discrepancy term, the residuals of the underlying ODEs or PDEs, in addition to the usual data-based loss components. As such, solutions that obey the selected physics are enforced. In contrast to SINDy, there is no sparsity constraint imposed on the network weights, meaning that the network could still be quite complex. Early examples of PINNs were published by Raissi \emph{et al.} \cite{Raissi.2017, Raissi.2017428, Raissi.2019, Raissi.2020} and Long \emph{et al.}\cite{long2018pde} in 2017. Since then, the architecture has been applied to a wide range of problems in the natural sciences, with a quasi-exponential growth in publications\cite{Cuomo.2022mo}. Applications include for instance (low-temperature) plasma physics, where PINNs have been successfully used to solve the Boltzmann equation\cite{kawaguchi2022physics}, or quantum physics, where PINNs were used to solve the Schrödinger equation of a quantum harmonic oscillator\cite{stiller2020large}. The work by Stiller \emph{et al.}\cite{stiller2020large} uses a a scalable neural solver that could possibly also be extended to solve e.g. the Vlasov-Maxwell system governing laser-plasma interaction.

\subsection{Time series forecasting}\label{forecasting}

A related problem meriting its own discussion is time series forecasting. While models in the previous section are based on interpolation or regression within given data points, forecasting explicitly deals with the issue of \textit{extrapolating} parameter values to the future based on prior observations. Most notably this includes modeling of long-term \emph{trends} (in a laser context often referred to as \textit{drifts}) or periodic oscillations of parameters and short-term fluctuations referred to as \textit{jitter}.

If available, models may also use \emph{covariates}, auxiliary variables that are correlated to the observable, to improve the forecast. These covariates may even extend into the future (seasonal changes being the traditional example in economic forecasting).

In this section we are going to first discuss two common approaches to time series forecasting, autoregressive models and state-space models, followed by a discussion of modern techniques based on neural networks. Note that the modeling approaches presented in the preceding section may also be used, to some extend, to extrapolate data. We are not aware of any recent examples on time series forecasting in the context of laser-plasma physics. However, we feel that the topics merits inclusion in this overview as implementations are likely in the near future, e.g. for laser stabilization purposes.

\subsubsection{Classical models}\label{ARIMA}

To model a time series one usually starts with a set of assumptions regarding its structure i.e. the interdependence between values at different times. A simple, approximate assumption is that the observed values in a discrete-time time series are linearly related. An important, wide-spread class of such models are so-called autoregressive models which assert that the next value \(y_t\) in a time series is given as a linear function of the \(p\) prior values \(y_{t-1}, y_{t-2}, \dots, y_{t-p}\). Additionally, each value is assumed to be corrupted by additive white noise \(\varepsilon_t\sim\mathcal{N}(0, \sigma^2)\), representing e.g. measurement errors or inherent statistical fluctuations of the underlying process, which endows the models with stochasticity. In its most common form the autoregressive model of order \(p\), denoted by \(\mathrm{AR}(p)\), is defined via the recurrence relation

\begin{equation}
    \label{eq:ar}
    y_t=\sum\limits_{i=1}^p\varphi_i y_{t-i}+\varepsilon_t,
\end{equation}
where \(\varphi_1,\dots, \varphi_p\) are the model's parameters to be estimated. In this form the problem of finding the model's parameters is a linear regression problem that can be solved via the least squares method

\begin{equation}
    \label{eq:ar_ols}
    \{\hat\varphi_1,\dots,\hat\varphi_p\}=\arg\min_{\varphi_1,\dots,\varphi_p}\left(y_t - \sum\limits_{i=1}^p\varphi_i y_{t-i}\right)^2 \ .
\end{equation}

In another approach we might assume that the next value \(y_t\) is instead given by a linear combination of (external) statistical fluctuations, sometimes called shocks, from the past \(q\) points in time, again encoded in white noise terms \(\varepsilon_t\sim\mathcal N(0,\sigma^2)\). The corresponding model, called the \emph{moving average} model and denoted by \(\mathrm{MA}(q)\), is defined as

\begin{equation}
    \label{eq:ma}
    y_t=\mu + \sum\limits_{i=1}^q\vartheta_i\varepsilon_{t-i}+\varepsilon_t \ ,
\end{equation}
where \(\mu\) is the mean of the time series.

The parameters of the moving average process cannot be inferred by linear methods like least squares, but have to be estimated by means of maximum likelihood methods. 

Contrary to the \(\mathrm{AR}(p)\) model, which is only stationary\footnote{A time series \(\{y_t\}\) is said to be strictly stationary, given the joint cumulative distribution \(F(y_0, \dots, y_t)\), if \(F(y_0,\dots, y_t)=F(y_{0+\tau},\dots, y_{t+\tau})\quad\forall\;\tau\in\mathbb N\). Correspondingly all moments of the joint distribution are invariant under time translation. If this invariance only holds for the first two moments (mean, variance) the time series is said to be (weakly) stationary.} for certain parameters \(\varphi_1,\dots, \varphi_p\), the \(\mathrm{MA}(q)\) model is always (strictly) stationary per definition. Put loosely, stationarity can be understood as the stochastic properties (mean, variance, ...) of the time series being constant over time.

Another distinction between the autoregressive and the moving average model is how far into the future the effects of statistical fluctuations (shocks) are propagated. In the moving average model the white noise terms are only propagated \(q\) steps into the future, while in the autoregressive model the effects are propagated infinitely far into the future. This is because the white noise terms are part of the prior values which themselves are part of future values in \cref{eq:ar}.

If the time series in question cannot be explained by the \(\mathrm{AR}(p)\) or the \(\mathrm{MA}(q)\) model alone, both models can be combined into what is called an autoregressive-moving-average model, denoted by \(\mathrm{ARMA}(p,q)\)

\begin{equation}
    \label{eq:arma}
    y_t=\mu+\sum\limits_{i=1}^p\varphi_i y_{t-i}+\sum\limits_{j=1}^q \vartheta_j\varepsilon_{t-j}+\varepsilon_t.
\end{equation}

Note that for the special cases of \(p=0\) and \(q=0\) the \(\mathrm{ARMA}(p,q)\) reduces to the \(\mathrm{MA}(q)\) and \(\mathrm{AR}(p)\) respectively. 

When fitting autoregressive models special care should be taken that the time series to be modelled is stationary before fitting, otherwise spurious correlations are introduced. Spurious correlations are apparent correlations between two or more time series that are \emph{not} causally related thereby potentially leading to fallacious conclusions as warned of in the well-known adage "correlation does not imply causation". 

If the time series \(y\) is non-stationary in general but stationary with respect to its mean, i.e. the variations relative to the mean value are stationary, it might be possible to transform it into a stationary time series. For this we introduce a new series \(z_t=\nabla^d y_t\) by differencing the original series \(d\) times, where we defined the differencing operator \(\nabla y_t\equiv y_t-y_{t-1}\). Applying the differencing operator once (\(d=1\)), for example, removes a linear trend from \(y_t\), applying it twice removes a quadratic trend and so forth.
If \(z\) is stationary after differencing \(y\) \(d\)-times it can be readily modelled by \cref{eq:arma} which leads us to the autoregressive-integrated-moving-average model \(\mathrm{ARIMA}(p,d,q)\)

\begin{equation}
    z_t = \sum\limits_{i=1}^p\varphi_i z_{t-i} + \sum\limits_{j=1}^q\vartheta_j \varepsilon_{t-j} + \varepsilon_t \ .
\end{equation}

Note that in contrast to the \(\mathrm{ARMA}(p,q)\) model in \cref{eq:arma} the time series \(z_t\) appearing here is a \(d\)-times differenced version of the original series \(y_t\). Further extensions that will only be noted here for completeness here are seasonal ARIMA models called SARIMA that allow the modelling of time series that exhibit seasonality (periodicity) and exogeneous ARIMA models called ARIMAX that allow the modelling of time series that are influenced by a separate, external time series. Both extensions can be combined to yield the so-called SARIMAX model which is general enough to cover a large class of time series problems. However we are still in any case limited to problems in which the time series is generated by a \emph{linear} stochastic process. To cover more general, non-linear problems we introduce state-space models.

\subsubsection{State-Space Models}

State-space models (SSMs) offer a very general framework to model time series data\cite{commandeur2007introduction}. In this framework, presuming that the time series is based on some underlying system, it is assumed that there exists a certain true state of the system $x_t$ of which we observe a value $y_t$ subject to measurement noise. The true state $x_t$, usually inaccessible and \textit{hidden} to us, and the observed state $y_t$ are modeled by the \textit{state equation} and the \textit{observation equation} respectively. A prominent example of SSMs in Machine Learning are Hidden Markov models (HMMs)\cite{zucchini2009hidden}, in which the hidden state $x_t$ is modeled as a Markov process. In general there are no restrictions on the functional form of the state and observation equations, however, the most common type of SSMs are linear Gaussian SSMs, also referred to as dynamic linear models, in which the state and observation equations are modeled as linear equations and the noise is assumed Gaussian. The prototypical example of such linear Gaussian SSMs is the \emph{Kalman filter}\cite{kalman1960new}, ubiquitous in control theory, robotics and many other fields\cite{auger2013industrial}, including for instance adaptive optics\cite{poyneer2010kalman} and particle accelerator control\cite{ushakov2017detuning}. The term ''filter'' originates from the fact that it filters out noise to estimate the underlying state of a system. The state equations can be written as

\begin{equation}
    \begin{split}
        x_t &= A_t x_{t-1} + B_t u_t + C_t \varepsilon_t\\
        y_t &= D_t x_t + E_t \eta_t \ ,
    \end{split} 
\end{equation}
where the system noise \(\varepsilon_t\) and observation noise \(\eta_t\) are sampled from a normal distribution \(\varepsilon_t,\eta_t\sim\mathcal N(0,1)\). Note that the separation of these two noise terms differs from the so-far discussed models. This permits the modelling of systems that are driven by noise, while also accounting for observation noise of possibly differing amplitude. Another difference is the addition of a second process \(u_t\), commonly called the \emph{control input} in signal processing, which is a deterministic external process driving the system state. This permits the modelling of systems with known external inputs (e.g. a controller-driven motor) or with known deterministic drivers (e.g. system temperature). 
Note that we can recover the autoregressive models discussed earlier as a special case of the Kalman filter. For example, the \(\mathrm{AR}(1)\) process is obtained by setting \(A_t=\text{const.}=\varphi_1\), \(B_t=0\), \(C_t=\text{const.}=\sigma^2\), \(D_t=1\) and \(E_t=0\). For a more detailed discussion on Kalman filters we refer the interested reader to the literature\cite{welch1995introduction}. Other (non-linear) ways to perform inference in state-space models exist, for instance using sequential Monte-Carlo estimation\cite{ristic2003beyond}. 

\subsubsection{Forecasting networks}\label{sec:forecasting_networks}

While predictions based on autoregression or state-space models can suffice for many applications, the non-linear nature of neural networks can be harnessed to model time series with complex, non-linear dependencies\cite{ismail2019deep,lim2021time}. A particularly relevant architecture are \emph{recurrent neural networks} (RNNs). These are similar to the ``traditional'' neural networks we discussed in \cref{neural_nets}, but they have a ``memory'' that allows them to retain information from previous inputs. This is implemented in a somewhat similar way to the linear recurrence relations discussed in the previous sections, with a key difference being that the state $x_t$ of RNNs is updated according to an arbitrary nonlinear function. The current state $x_t$ is calculated from the previous states $x_{t-1}, x_{t-2}, \dots$ through the recurrence equation
\begin{equation}
    x_t = f(w_\text{rec}x_{t-1}+b_\text{rec}),
\end{equation}
where $f$ is a nonlinear activation function, $w_\text{rec}$ a matrix of recurrent weights, and $b_\text{rec}$ is a bias vector. This equation allows the update to each element of the current state vector, $x_t$, to be be dependent on the whole of the previous state vector, $x_{t-1}$.  The output $y_t$ of the network is calculated from the current state $x_t$ through the output equation
\begin{equation}
    y_t = f(w_\text{out}x_t+b_\text{out}),
    \label{eqn:RNN_out}
\end{equation}
where $w_\text{out}$ contains the output weights and $b_\text{out}$ is another bias vector.  Thus, the entire network state is updated from the previous state through a recurrent equation, and the current output is calculated from the current state through an output equation. The network state therefore contains all previous information about the time series. However, as the network state is updated by a simple multiplication of the weights $w_\text{rec}$ with the previous state $x_{t-1}$, the so-called vanishing gradient problem may occur\cite{hochreiter1991untersuchungen,Hochreiter.1998}. This problem is solved in a seminal work by Hochreiter \cite{Hochreiter.1997}, who introduced a special type of RNNs, the \emph{long short-term memory} (LSTM) network. In contrast to the simple update equation of RNNs, LSTMs use a special type of memory cell that can learn long-term dependencies. This includes a so-called \emph{forget gate} \(f_t\) to update the previous state $x_{t-1}$ to the current state $x_t$:
\begin{equation}
    x_t = f_t \odot x_{t-1} + i_t \odot h_t
\end{equation}
where $\odot$ denotes the element-wise Hadamard product, $[A\odot B]_{ij}=A_{ij}B_{ij}$. Additionally, the LSTM uses two more gates, the input gate $i_t$ and output gate $o_t$, where the latter is used to calculate the hidden state $h_t$ from the previous hidden state via $h_t = o_t \odot h_{t-1}$. The three gates $f_t$, $i_t$ and $o_t$ are calculated from the current input $x_t$, the previous hidden state $h_{t-1}$, and the previous gate states $f_{t-1}$, $i_{t-1}$, and $o_{t-1}$ using individually set weights and biases. The gates determine how much of the previous state $x_{t-1}$ and the current hidden state $h_t$ is used to calculate the current state $x_t$. The output of the LSTM network is then calculated from the current state $x_t$ through the same output equation, \cref{eqn:RNN_out}, as used in the RNN. Despite it being introduced in the early 1990s, the LSTM architecture remains one of the most popular network architectures for predictive tasks to date.

A more recently introduced type of neural network that can learn to interpret and generate sequences of data are \emph{transformers}. Transformers are similar to RNNs, but instead of processing the time series in a sequential manner, they use a so-called attention mechanism\cite{vaswani2017attention} to capture dependencies between all elements of the time series simultaneously and thus, focus on specific parts of an input sequence. Assuming again a time series $x_t, x_{t-1}\dots$, a transformer maps each point $x_t$ to a representation $h_t$ using a linear layer, e.g., $h_t = w x_t + b$. The transformer network then calculates a new representation for each point $x_t$ using the attention mechanism
\begin{equation}
    \tilde{h_t} = \sum \alpha_{ti}h_i,
\end{equation}
where the attention weights $\alpha_{ti}$ are calculated using the point representations $h_i$ and the previous representation $\tilde{h_t}$. The attention weights $\alpha_{ti}$ are used to calculate the new representation $\tilde{h_t}$ of each point $x_t$ from all previous representations $\tilde{h_i}$, $i < t$, of the previous point $x_i$, $i < t$. The new representations are then used to calculate the output $y_t$ of the network as in \cref{eqn:RNN_out}. Thus, the attention mechanism of a transformer network can be interpreted as a non-linear function that updates the representation of each point $x_t$ from all previous representations of the previous points $x_i$, $i < t$. It can be applied multiple times and enables transformers to learn complex patterns in data, outperforming RNNs on a variety of tasks such as machine translation and language modeling. It has also been shown that Transformers are more efficient than RNNs, meaning they can be trained on larger datasets in less time. One recent example of a transformer developed for time series prediction is the Temporal Fusion Transformer\cite{lim2021temporal}. Beyond RNNs, LSTMs and Transformers there exist many other network architectures that can be used for forecasting, e.g. Fully Convolutional Networks as discussed by Wang \emph{et al.}\cite{wang2017time}. 

For longer-term forecasting, predictive networks are usually employed iteratively, meaning that a single-step forecast uses only real data, while a two-step forecast will use the historical data and the most recent prediction value, and so forth. In the context of laser-plasma physics and experiments, predictive neural networks could be used to model the time-series data from diagnostic measurements, in order to make predictions about the future performance, e.g. for predictive steering of laser or particle beams.

\subsection{Prediction and Feedback}\label{sec_feedback}

Both (surrogate) models and forecasts can be used to make predictions about the state of a system at a so-far unknown position, in parameter space or time, respectively. This type of operation is sometimes referred to as open-loop prediction. Closed-loop prediction and feedback, on the other hand, use predictions and compare them to the actual system state, continuously updating and improving the surrogate model of the system. This is particularly relevant for dynamic systems, where parameters change over time. A complete discussion how to implement a closed-loop system using machine learning goes beyond the scope of this review, as it would also require an extensive discussion of control systems and so forth. The following discussion will thus be restricted to a brief outline of a few relevant concepts.

A feedback loop is most generally a system where a part of the output serves as input to the system itself and we have already discussed some models with feedback in the context of forecasting (\cref{forecasting}). Another well-known engineering implementation of closed-loop operation is the proportional–integral–derivative (PID) controller. A PID controller adjusts the process variable (PV) with a proportional, integral and derivative response to the difference between a measured PV and a desired set point. Here the proportional term serves to increase the response time of the system to changes in the set point. The integral term is used to eliminate the residual steady-state error that occurs if the PV is not equal to the set point. The derivative term improves the stability of the control, reducing the tendency of the PID output to overshoot the set point.

In the context of laser-plasma physics an the implementation of the feedback loop would most likely look slightly different. One possible implementation would be that a model is used to predict the output of the system at a given set of parameters, which is then compared to the actual output and then used to update the model again. In order to keep improving the model, it is important that new data be acquired in regions of parameter space that deviate from the previously acquired data. In other words, it is important to \textit{explore} parameter space in addition to \textit{exploiting} the knowledge from existing data points. This can be done through random sampling or through so-called \textit{Bayesian optimization}, as we will discuss in \cref{optimization}.

Such a model that is continuously updated with new data is sometimes referred to as a dynamic model. There are two main approaches to updating such models: 

\begin{itemize}
    \item First, completely re-training the model with all available data, including the newly acquired data points. This results in an increasingly large training data set and training time.
    \item The second method is to update the model by adding new points to the training set and then re-training the existing model on these points in a process known as \textit{incremental learning}. This method is thus much faster and uses less memory than a full re-training.
\end{itemize}

However, not all techniques we discussed to construct models are compatible with both training methods. For instance, Gaussian process regression can historically only be trained on full data sets, hindering its applicability in settings requiring (near) real-time updates. However, recent work on regression in a streaming setting might alleviate this problem\cite{bui2017streaming}. Learning dynamic models is further complicated by the fact that systems may change over the course of which training data is acquired, a problem referred to as \textit{concept drift}\cite{vzliobaite2016overview}.

\example{Many laboratories are currently stepping up their efforts to integrate prediction and feedback systems into their lasers and experiments\cite{cassou:2022,Weisse2022HPLSE}. One of the first groups to extensively make use of these techniques was the team around A. Maier at DESY's LUX facility\cite{Maier.2020}. Among the most comprehensive proposals are plans recently presented by T. Ma and colleagues from NIF, proposing an extensive integration of feedback systems at high-energy-density (HED) experiments\cite{ma2021accelerating}. The system, referred to as "full integrated high-repetition rate HED laser-plasma experiment" consists of multiple linked feedback loops. These are hierarchically organized, starting from a "laser loop", over an "experiment loop"  to a "modeling \& simulation loop" at its highest level.}

\section{Inverse problems}\label{scn:inverse_problems}

In the previous section we have looked at the \emph{forward} problem of modeling the black box function $f(x)=y$. In many scientific and engineering applications, it is necessary to solve the inverse problem, namely to determine $x$ given $y$ and $f$ (or an approximation $f^{*}\approx f$). Inverse problems are extremely common in experimental physics, as they essentially describe the measurement process and subsequent retrieval of underlying properties in a physics experiment.

In many cases the problem takes the form of a discrete linear system 
\begin{equation}
    A x =  y
    \label{inverse_problem}
\end{equation}

where $y$ is the known observation and $A$ describes the measurement process acting on the unknown quantity $x$ to be estimated. In most cases we can assume that the system behaves linear with respect to the input parameters $x$ and hence, $A$ can be written as a single \emph{sensing} matrix. However, more generally, $A$ can be thought of as an operator, which maps the input vector $x$ to the observation vector $y$.\footnote{The sensing matrix/operator is known by different names, e.g. the instrument response, the system matrix, the response matrix, the transfer function, or most generally, the forward operator.}

A classical example of an inverse problem is computed tomography (CT), whose goal is to reconstruct an object from a limited set of projections at different angles. Other examples are wavefront sensing in optics, the "FROG" algorithm for ultrafast pulse measurements, the "unfolding" of X-ray spectra or, in the context of particle accelerators, the estimation of particle distributions from different measurements of a beam.

\subsection{Least squares solution}\label{sec:leastsquares}

A common approach to the problem described by \cref{inverse_problem} is to use the least squares approach. Here, the problem is reformulated as minimizing the quadratic, positive error between observation and estimate
\begin{equation}
    \hat x_{LS} =  \underset{x}{\mbox{argmin}} \{||A x -  y||^2\}.
    \label{eq:leastsquares}
\end{equation}

If $A$ is square and non-singular, the solution to this equation is obtained via the inverse, $A^{-1}y$. For non-square matrices, the pseudo-inverse can be used, which can for instance be computed via singular value decomposition\cite{barata2012moore}. Alternatively, a multitude of iterative optimization methods can be employed, e.g. iterative shrinkage-thresholding algorithms \cite{Bioucas-Dias.2007,Teboulle2009} or gradient descent, to name a few. For more details on the solution to this optimization problem, the reader is referred to Ref.\cite{Nocedal.2006}.

The approach given by \cref{eq:leastsquares} is widely used in overdetermined problems, where the regression between data points with redundant information can help to reduce the influence of measurement noise. Prominent application areas are for instance wavefront measurements and adaptive optics\cite{tyson2022principles}.

\subsection{Statistical inference}\label{scn:inverse_probabilistic}

As for predictive models (\cref{probabilistic}) one can also approach inverse problems via probabilistic methods, e.g. if the underlying model $f(x)$ is stochastic in nature or measurements are corrupted by noise. One popular approach is to use the maximum likelihood estimator (MLE) \cite{tarantola2005inverse}. As we have seen before, MLE consists in finding the value that maximizes the likelihood, see \cref{eq:log-likelihood-factorization}. To this end, one often uses Markov chain Monte Carlo methods\cite{geyer1991markov} and/or gradient descent algorithms. Alternatively, Expectation-Maximization (EM) is a popular method to compute the maximum likelihood estimate (MLE) \cite{dempster1977maximum} and is, for instance, often used in statistical iterative tomography (SIR)\cite{zhang2018regularization}. EM alternates between an estimate step, where one computes the expectation of the likelihood function, and a maximization step, where one obtains a new estimate that maximizes the posterior distribution. By sequentially repeating the two steps, the estimate converges to the maximum likelihood estimate. 

Both the least squares and MLE approaches suffer from the issue that the result is a point estimate. Thus, if the underlying problem is ill-posed or underdetermined, this estimate is often not unique or representative, or the least squares solution is prone to artifacts resulting from small fluctuations in the observation. Consequently, it is often desirable to obtain an estimate of the entire solution space, i.e. a probability distribution of the unknown parameter $x$. The latter allows to not only compute the estimate ${\hat x}$, but to also obtain a measure of the estimation uncertainty ${\sigma}_{\hat x}$. 

To this end, one can reformulate the inverse problem as a Bayesian inference problem and get both an expectation value and an uncertainty from the posterior probability $p(y|x)$. As we have seen in \cref{probabilistic}, calculating the posterior requires knowledge of the likelihood $p(x|y)$ \cite{tarantola2005inverse}. However, in some settings the forward problem $f(x)$ can be very expensive to evaluate, which is for instance the case for accurate simulations of laser-plasma physics. In this case the likelihood function $p(x|y)$ becomes intractable because it requires the computation of $f(x)$ for many values of $x$. Historically, solutions to this problem have been referred to as methods of likelihood-free inference, but more recently the term \emph{simulation-based inference} is being used increasingly\cite{cranmer2020frontier}. One of the most popular methods in this regard is \emph{approximate Bayesian computation} (ABC)\cite{sisson2018handbook}\footnote{While both rely on Bayesian statistics and thus have some conceptual overlap, approximate Bayesian computation should not be confused with Bayesian \emph{optimization} discussed in \cref{BO}.}, which addresses the issue by employing a reject-accept scheme to estimate the posterior distribution $p(y| x)$ without needing to compute the likelihood $p(x|y)$ for any value of $x$. Instead, ABC first randomly selects samples $x^\prime$ from a pre-defined prior distribution, which is usually done using a MCMC\cite{gilks1995markov} algorithm or, in more recent work, via Bayesian optimization\cite{gutmann2016bayesian}. Defining good priors might require some expert knowledge about the system. Subsequently, one simulates the observation $y^{\prime}=f(x^\prime)$ corresponding to each sample. Finally, the ABC algorithm accepts only those samples which match the observation within a set tolerance $\epsilon$, thereby essentially approximating the likelihood function with a rejection probability $\rho(y,y^{\prime})$. This yields a reduced set of samples, whose distribution approximates the posterior distribution. The tolerance $\epsilon$ determines the accuracy of the estimate; the smaller $\epsilon$ the more accurate the prediction, but due to the higher rate of rejection one also requires more samples of the forward process. This can make ABC prohibitively expensive, especially for high-dimensional problems. Approximate Frequentist computation\cite{brehmer2018guide} is a conceptually similar approach that approximates the likelihood function instead of the posterior. Information on this method and other simulation-based inference techniques, including their recent development in the context of deep learning e.g. for density estimation in high dimensions, can for instance be found in the recent overview paper by Cranmer \emph{et al.}\cite{cranmer2020frontier}.

\subsection{Regularization}\label{sec:regularization}

One way to improve the estimate of the inverse in the case of ill-posed or underdetermined problems is to use \emph{regularization} methods. Regularization works by further conditioning the solution space, thus replacing the ill-conditioned problem with an approximate, well-conditioned one. In \emph{variational regularization methods} this is done by simultaneously solving a second minimization problem that incorporates a desirable property of the solution, e.g. minimization of the total variation to remove noise\footnote{Another class of regularization methods are based on smoothing in some sense. For instance, filtered back projection can be understood as a smoothing of the projection line integrals by means of a convolution with a filter function.}. Such regularization problems hence take the form
\begin{equation}
    {\hat x_{REG}} =  \underset{x}{\mbox{argmin}} \{||{A}x - y||^2 + \lambda \mathcal R( x) \}.
\label{reg}
\end{equation}
where $\lambda$ is a hyperparameter controlling the impact of the regularization and $\mathcal R$ is the regularization criterion, e.g. a matrix description of the first derivative.
The effect of the regularization can be further adjusted by the norm that is used to calculate the residual term, e.g. using the absolute difference, the squared difference or a mix of both such as the Huber norm.\footnote{It should be noted that a very similar problem formulation can also be found in the context of training neural networks, where regularization terms are often added to the cost function.} The more complex optimization problem \cref{reg} itself is typically solved using iterative optimization algorithms, as already mentioned in the previous section. As will be discussed in \cref{deepunrolling}, there has been recent developments in data driven approaches to learning the correct form of  $\mathcal R (x)$, by representing it with a neural network.

\example{A recent demonstration in the context of laser-plasma acceleration is the work by A. Döpp et al.\cite{Dopp.2018,Gotzfried.2018}, who presented results on tomographic reconstruction using betatron X-ray images of a human bone sample. The 180 projection images were acquired within three minutes and an iterative reconstruction of the object's attenuation coefficients was performed. For regularization, a variation of \cref{reg} was used that used the Huber norm and included a weighting factor in the data term, whose objective it was to adjust the weight of poorly-illuminated detector pixels.}

 \begin{figure}
    \centering
    \includegraphics[width=\linewidth]{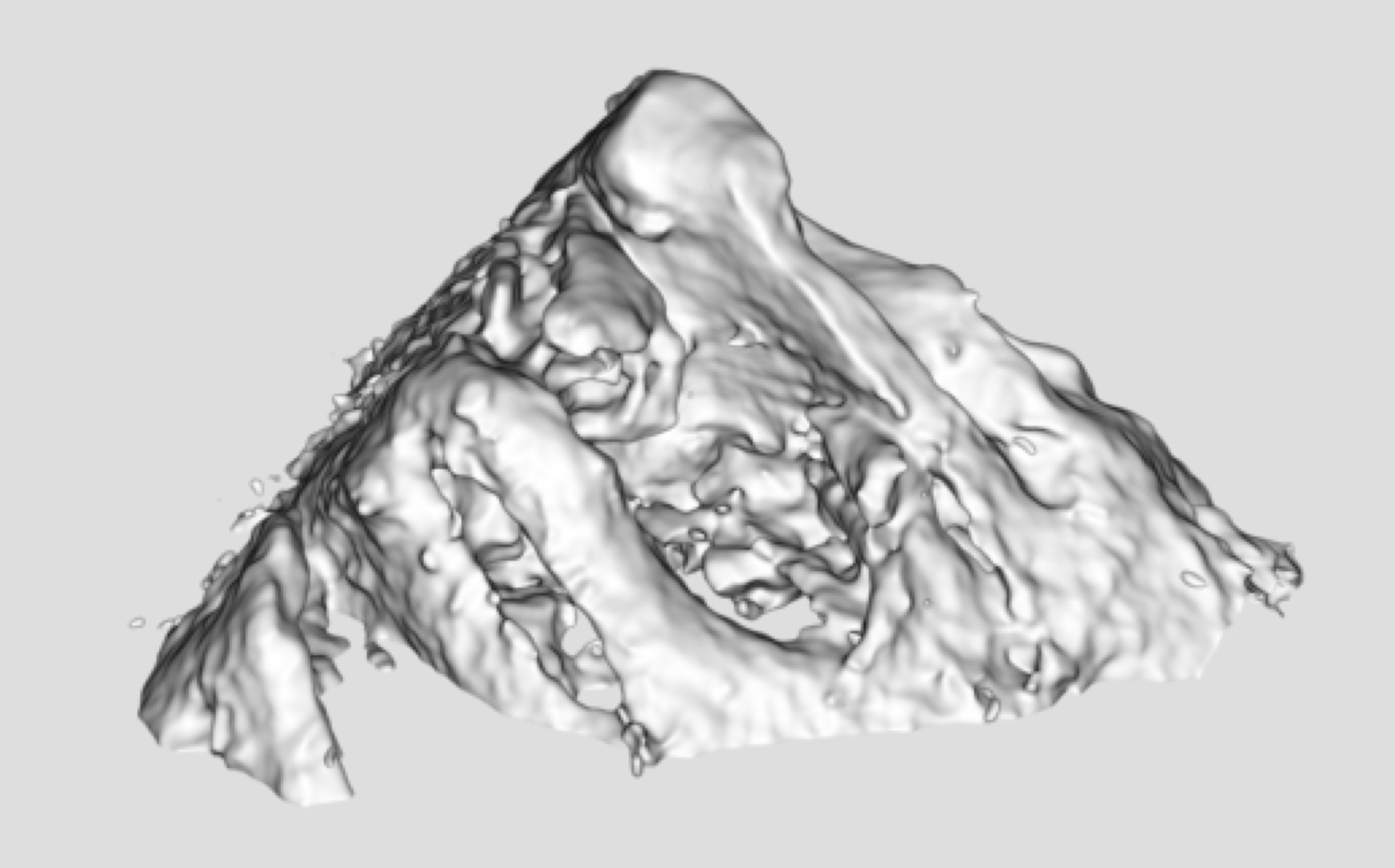}
    \caption{\textbf{Tomography of a human bone sample using a laser-driven betatron X-ray source.} Reconstructed from 180 projections using statistical iterative reconstruction. Based on the data presented in Döpp \textit{et al.}\cite{Dopp.2018}. }
    \label{fig:opticalTomo}
\end{figure}

\subsection{Compressed sensing}\label{compressed}
Compressed sensing (CS)\cite{Candes.2008, Oliveri.2017, Yuan.2021} is a relatively new research field that has attracted significant interest in recent years, since it efficiently deals with the problem of reconstructing a signal from a small number of measurements. The mathematical theory of CS has proven that it is possible to reliably reconstruct a complex signal from a small number of measurements, even below the Shannon-Nyquist sampling rate, as long as two conditions are satisfied. First, the signal must be ``sparse'' in some other representation (i.e. it must contain few non-zero elements). In this case we can replace the dense unknown variable $x$ with its sparse counterpart $\tilde{x}$ by using the transformation matrix $\Psi$ corresponding to a different representation, i.e.  $ x = \Psi \tilde{ x}$. Here $\Psi$ could for instance be the wavelet transformation. The second condition concerns the way the measurements are taken (hence compressed \textit{sensing}): the sensing matrix $ A$ must be incoherent with respect to the sparse basis $ \Psi$, which ensures that the sparse representation of the signal is fully sampled.  

At its core CS is closely related to the concepts of regularization discussed in the previous section. In order to reconstruct the signal from the measurements, the ideal regularization, $\mathcal R(\tilde{x})$, is that which sums the number of non-zero components of $\tilde{x}$ and thus promotes sparsity when minimized\cite{Donoho.2006}. However, it has been shown that using the vastly more computationally efficient $\ell_1$ norm leads to the same results in many occasions \cite{Candes.2008a}, and thus the CS reconstruction problem can be written as:
\begin{equation}
    {\hat x_{CS}} =  \underset{\tilde{ x}}{\mbox{argmin}} \{||{A} \Psi\tilde{{x}} -  y||^2 + \lambda ||\tilde{ x}||_1 \}.
\label{eqn:CS}
\end{equation}
It should be noted that compressed sensing is not the first method to achieve sub-Nyquist sampling in a certain domain, see e.g. Band-limited sampling\cite{kohlenberg1953exact}. The strength of the formalism is rather that it is very flexible, because it only requires the coefficients of a signal to be sparse, without exactly knowing beforehand \textit{which ones} are non-zero.  

It should be noted that in the most general case, the basis in which the signal is sparse in, $\Psi$, is unknown. Nonetheless, the incoherence with the sampling basis can be satisfied by sampling randomly. To reconstruct the signal from such measurements, one returns to solving \cref{reg}. This can still be considered as compressed sensing, due to the fact that the nature of the sensing process is designed to exploit the sparsity of the signal. Therefore, CS can be used alongside deep learning based approaches to solve \cref{reg}, that will be discussed in the subsequent sections. 

\example{CS has been used in a number of fields related to laser-plasma physics, for example ultrafast polarimetric imaging \cite{Liang:21}, and ICF radiation symmetry analysis\cite{HUANG2014459}. There exist also several examples from the context of tomographic reconstruction\cite{compressedtomography,ma2020region}.
Such enhanced reconstruction algorithms can reduce the number of projections beyond what is usually possible with regression techniques. For instance, in the work by Ma \emph{et al.}\cite{ma2020region} a test object was illuminated using a Compton X-ray source and a compressive tomographic reconstruction algorithm\cite{yu2009compressed} was used to reconstruct the sample's well-defined inner structure from only 31 projections.}

\begin{figure}
    \centering
    \includegraphics[width=.9\linewidth]{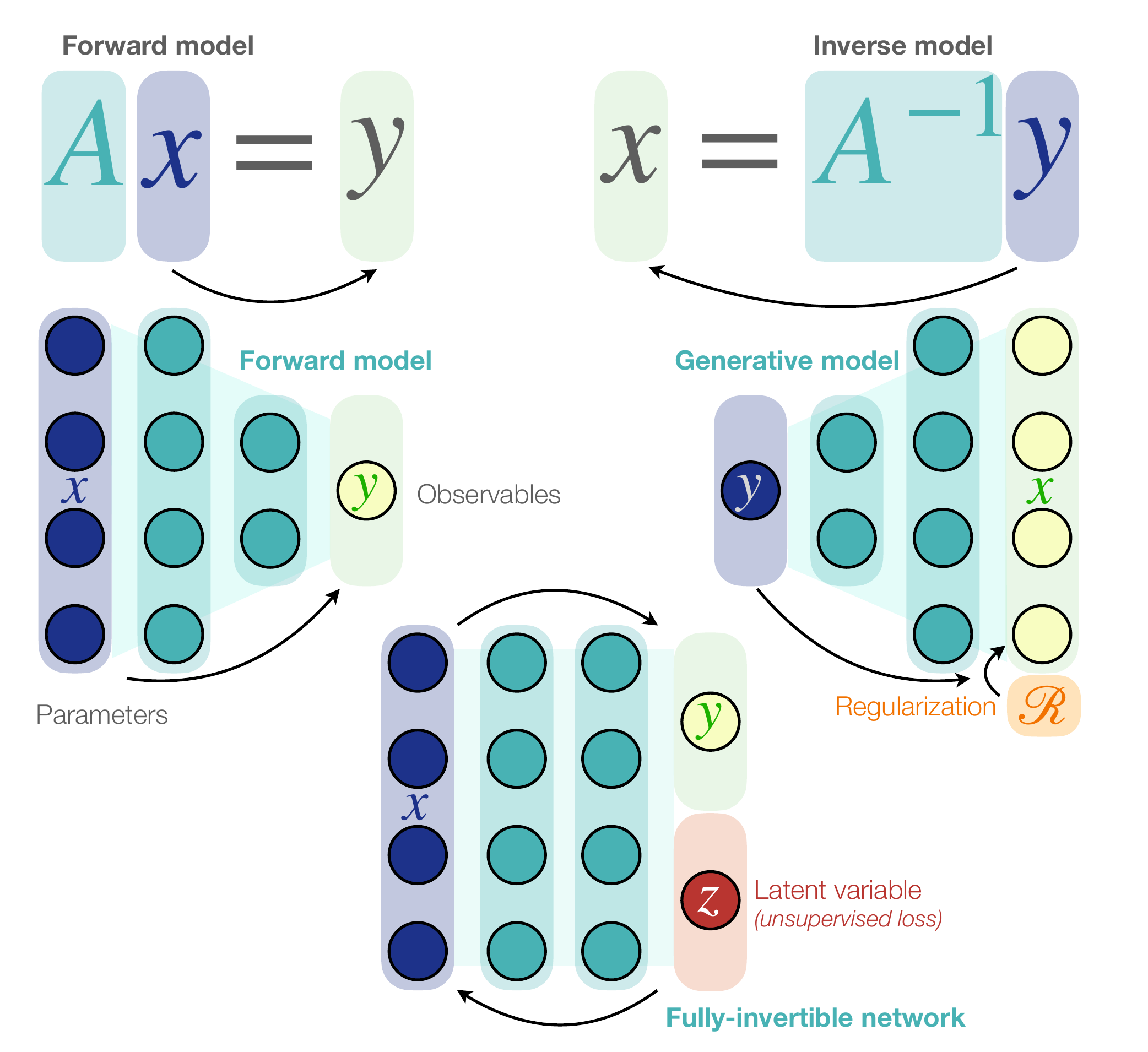}
    \caption{\textbf{Deep-learning for inverse problems.} Sketch explaining the relation between predictive models, inverse models and fully-invertible models.}
    \label{fig:inverseModek}
\end{figure}

\subsection{End-to-end deep learning methods}\label{deep_inverse_problems}

In recent years, there has been growing interest in applying deep learning to inverse problems\cite{Arridge.2019}. In general, these can be categorized into two types of approaches, namely those that aim to entirely solve the inverse problem end-to-end using a neural network and hybrid approaches that replace a subpart of the solution with a network (see next section). Denoting the artificial neural network as $\mathcal{A}$, the estimate ${\hat x_{\mathcal{A}}}$ from the end-to-end network could be simply written as
\begin{equation}
    {\hat x_{\mathcal{A}}} =  \mathcal{A}{y}.
\end{equation}

Artificial neural networks can thus be interpreted as a non-linear extension of linear algebra methods like singular value decomposition\cite{Bermeitinger.2019}. As such, for well-posed problems, the ANN is acting similarly to the least-squares method and, if provided with non-biased training data, directly approximates the (pseudo-)inverse. 

However, using neural networks for ill-posed problems is more difficult, as training tends to get unstable when the networks need to \textit{generate} data, i.e. the layer containing the desired output $ x$ is larger than the input $ y$. Fortunately, several network architectures have been developed that perform very well at these tasks, such as the generative adversarial networks (GANs) \cite{Goodfellow.2014}. GANs are two-part neural networks, one of which is trying to generate data that resembles the input (the \emph{generator}), while the other is trying to distinguish between the generated and real data  (the \emph{discriminator}). As the two networks compete against each other they improve their respective skills, and the generator will eventually be able to create high-quality data. The generator is then used to estimate the solution of the inverse problem. Training GANs can still be challenging, with common issues such as mode collapse\cite{thanh2020catastrophic}. Autoencoders are a simpler architecture that can perform similar tasks with relevance to inverse problems \cite{Peng.2019}. U-Nets\cite{Ronneberger.2015} are a popular architecture that draws inspiration from autoencoders and has proven to be very powerful for inverse problems, especially in imaging. They essentially combine features of autoencoders with fully convolutional networks, in particular ResNets (see also \cref{CNN}): Similar to an autoencoder, U-Nets consist of an encoder part, followed by a decoder part that usually mirrors the encoder. At the same time, the layers in the network are skip-connected, meaning that the output from the previous layer are concatenated with the output of the corresponding layer in the encoder part of the network (see \cref{fig:neuralNet}). This allows the network to retain information about the data about the input data even when it is transformed to a lower dimensional space.

One sub-class of ANNs that gained considerable recent interest in the context of inverse problems are invertible neural networks (INNs) \cite{ardizzone2018analyzing}. Mathematically, an INN is a bijective function between the input and output of the network, meaning that it can be exactly inverted. Because of this, an INN trained to approximate the forward function $A$ will implicitly also learn an estimate for its inverse. In order to be applied to inverse problems, both forward and inverse mapping should be efficiently computable. In ill-posed problems there is an inherent information loss present in the forward process, which is either counteracted by the introduction of additional latent output variables $z$, which capture the information about $x$ that is missing in $y$, or by adding a zero-padding of corresponding size. The architecture and training of INNs is inspired by recent advances in \textit{normalizing flows} \cite{rezende2015variational}. The name stems from the fact that the mapping learned by the network is composed of a series of invertible transformations, called \textit{coupling blocks}\cite{dinh2014nice}, which "normalize" the data, meaning that they move the data closer to a standard normal distribution $\mathcal N(0,1)$. The choice of coupling block basically restricts the Jacobian of the transformation between the standard normal latent variable and the output. INNs can be trained bi-directionally meaning that the loss function is optimized using both the loss of the forward pass and the inverse pass. Additionally an unsupervised loss such as Maximum Mean Discrepancy\cite{gretton2012kernel} or Negative Log Likelihood\cite{Ardizzone2019} can be used to encourage the latent variable to be as close to a standard normal variable as possible. 

The loss function of end-to-end networks can also be modified such that a classical forward model is used in the loss function. Such physics-informed neural networks (PINNs) (see \cref{scn:PINNs}) respect the physical constraints of the problem, which can lead to more accurate and physically plausible solutions for the inverse problem.

\begin{figure}[!tb]
    \centering
    \includegraphics[width=\linewidth]{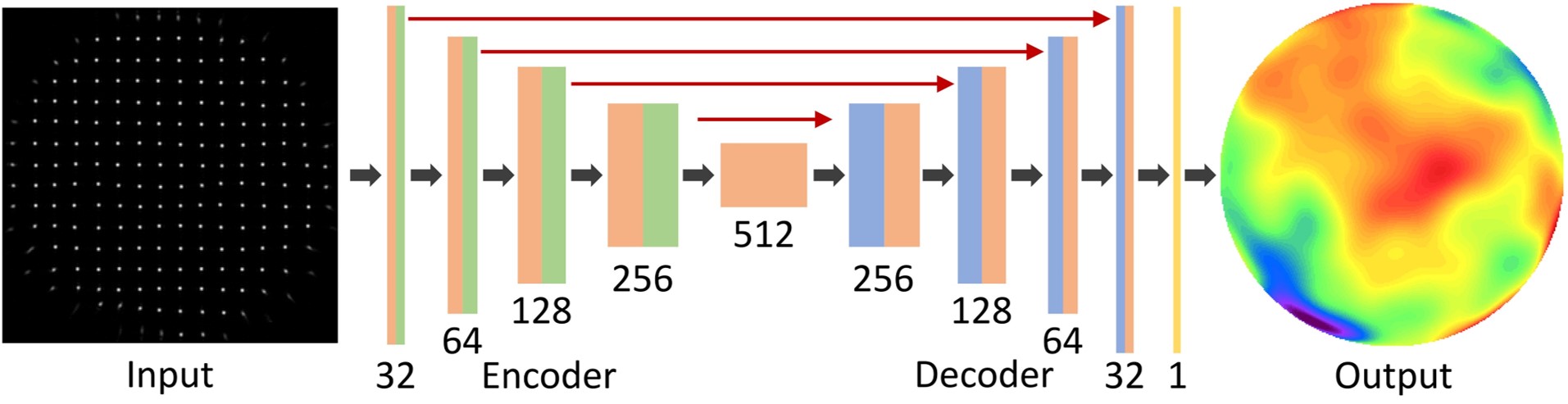}
    \caption{\textbf{Application of end-to-end reconstruction} of a wavefront using a convolutional U-NET architecture\cite{Ronneberger.2015}. The spot patterns from a Shack-Hartmann sensor are fed into the network, yielding a high-fidelity prediction.  Adapted from Hu \emph{et al.} \cite{shackhartmann}.}
    \label{fig:U-net-wavefront}
\end{figure}

\example{
An early example from the context of ultrafast laser diagnostics was published by Krumbügel \emph{et al.} in 1996, where the authors tested a neural network for retrieval of laser pulse profiles from FROG traces\cite{Krumbugel.1996}. At the time numerical capabilities were much more limited than today and the FROG traces containing some $64\times 64$ data points was at the time considered as ``far too much for a neural net''. Because of this, the data had to be compressed and only a polynomial dependence was trained, limiting reconstruction performance. The vast increase in computing power over the past decades has largely alleviated this issue and the problem was revisited by Zahavy \emph{et al.} in 2018\cite{zahavy2018deep}. Their "DeepFROG" network is based on a convolutional neural network architecture and, now using the entire two-dimensional FROG trace as input, achieves a similar performance as iterative methods. 

Another example for an end-to-end learning of a well-conditioned problem was published by R. A. Simpson et al.\cite{simpson2021development}, who trained a fully-connected three-layer network on a large set (47,845 samples) of synthetic spectrometer data to retrieve key experimental metrics, such as particle temperature. 

U-Nets have been used for various inverse problems, including e.g. the reconstruction of wavefronts from Shack-Hartmann sensor images as presented by Hu \emph{et al.} \cite{shackhartmann}, see \cref{fig:U-net-wavefront}. 

An example for the use of invertible neural networks in the context of laser-plasma acceleration was recently pubished by F. Bethke\emph{ et al.}\cite{bethke2021invertible}. In their work they used an INN called iLWFA to learn a forward surrogate model (from simulation parameters to beam energy spectrum) and then used the bijective property of the INN to calculate the inverse, i.e. deduce the simulation parameters from a spectrum.}

\subsection{Hybrid methods}\label{deepunrolling}
In contrast to end-to-end approaches, there exist a variety of hybrid schemes that employ neural networks to solve part of the inverse problem. A collection of such methods focuses on splitting \cref{reg} into two subproblems, separating the quadratic loss from the regularization term. The former is easily minimized as a least squares problem (discussed in \cref{sec:leastsquares}) and the optimum form of regularization can then be learned by a neural network. Crucially, this network can be smaller and more parameter-efficient than in end-to-end approaches, as it has a simpler task and less abstraction to perform. Furthermore, the direct relation that this network has to the regularization $\mathcal R (x)$ allows for one to pick a network structure to exploit prior knowledge about the data.

There are multiple methods available to perform such a separation, for example half quadratic splitting (HQS) or the alternating direction method of multipliers (ADMM). HQS is the simplest method and involves substituting an auxiliary variable, ${p}$, for ${x}$ in the regularization term and then separating \cref{reg} into two subproblems, which are solved for iteratively. The approximate equality of ${p}$ and ${x}$ is ensured by the introduction of a further quadratic loss term into each of the subproblems: $\beta ||{x}-{p}||^{2}$. 
If the same neural network (with the same set of weights) is used to represent the regularization subproblem in each iteration, the method is referred to as plug-and-play, but if each iteration uses a separate network, the method is deep unrolling \cite{monga.2019}. Such methods have achieved unprecedented accuracy whilst reducing computational cost. 

It is worth noting that there exist other similar methods - for instance, neural networks can be used to learn an appropriate regression function $\mathcal R( x)$ in \cref{reg}, as for instance done in Network Tikhonov (NETT)\cite{li2020nett}. 

\begin{figure}
    \centering
     \includegraphics[width=\linewidth]{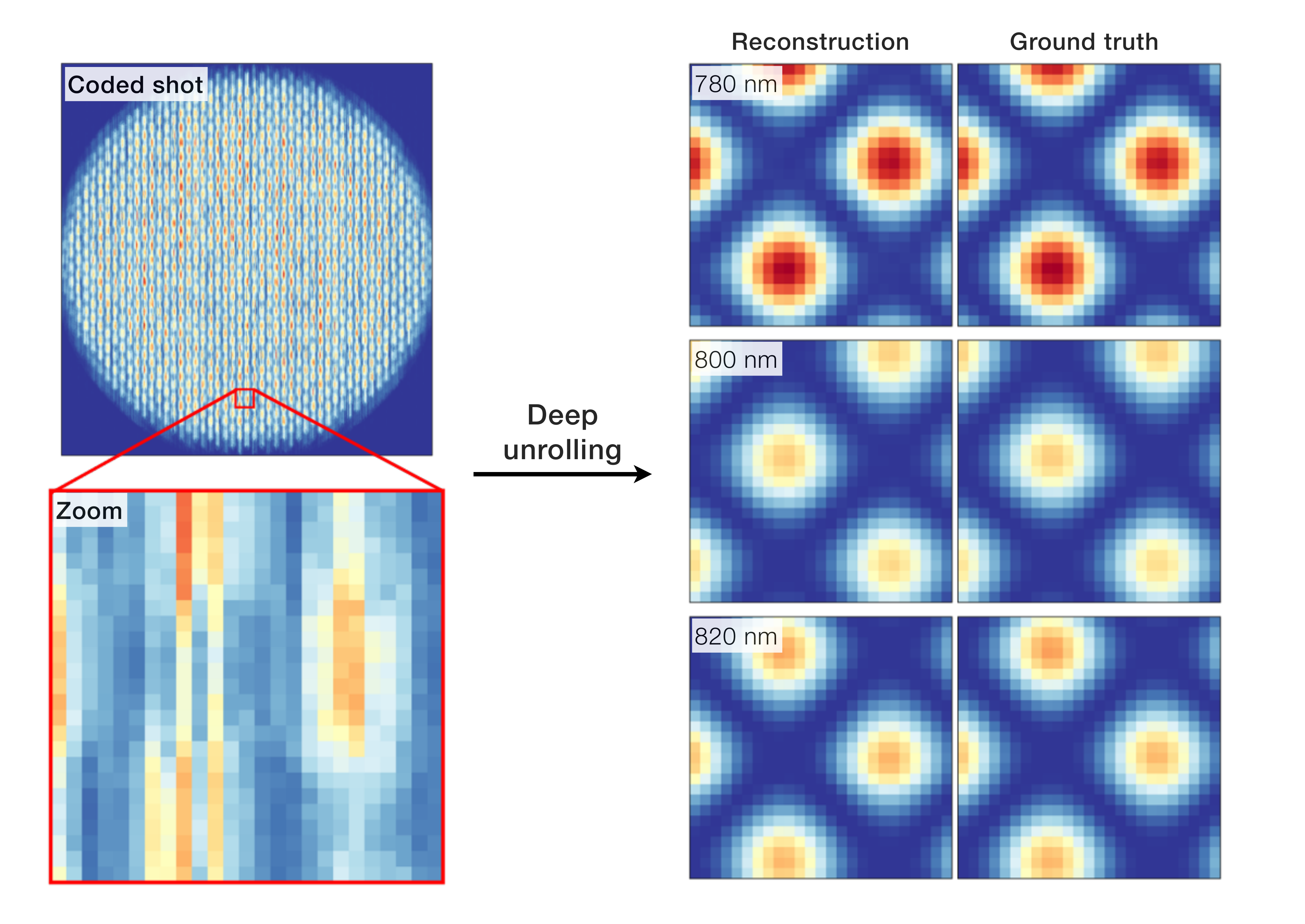}
    \caption{\textbf{Deep unrolling for hyperspectral imaging.} The left displays an example of the coded shot, i.e. a spatial-spectral interferogram hypercube randomly sampled onto a 2D sensor. The bottom left shows a magnification of a selected section. On the right is the corresponding reconstructed, spectrally resolved hypercube. Adapted from Howard \emph{et al.}\cite{Howard2022HPLSE}.}\label{im:howardfig}
    
\end{figure}

\example{Howard \emph{et al.} recently presented an implementation of compressed sensing using deep unrolling for single-shot spatio-temporal characterization of laser pulses\cite{Howard2022HPLSE}. Such a characterisation is a typical example of an underdetermined problem, where one wishes to capture three-dimensional information (across the pulse's spatial and temporal domains) on a two-dimensional sensor. Whilst previous methods mostly resorted to scanning, resulting in long characterisation times and blindness to shot-to-shot fluctuations, this work presented a single-shot approach, which has numerable benefits. The authors' implementation is based on a lateral shearing interferometer to encode the spatial wavefront in an interferogram for each spectral channel of the pulse, creating an interferogram cube. An optical system called CASSI\cite{cassi} was then used to randomly sample this 3D data onto a 2D sensor resulting in a `coded shot', thereby fulfilling the conditions of compressed sensing. For the reconstruction of the interferogram cube, the half quadratic splitting method was utilised, and the regularization term was represented by a network with three-dimensional convolutional layers that can capture correlations between the spectral and spatial domains. An example for a successful reconstruction is shown in \cref{im:howardfig}.}

\section{Optimization}\label{optimization}

One of the most common problems in applied laser-plasma physics experiments, in particular laser-plasma acceleration, is the optimization of the performance through manipulation of the machine controls. Here the goal is to minimize or maximize an objective function, a metric of the system performance according to some pre-defined criteria, see \cref{scn:objective_functions}. A simple case of this would be to optimize the total charge of the accelerated electron beam, but in principle, any measurable beam quantity or combination of quantities can be used to create the objective function.

There are many different general techniques for tackling optimization problems, and their suitability depends on the type of problem.
Single shots in a laser-plasma acceleration experiment (or a single run of numerical simulation) can be considered the evaluation of an unknown function that one wishes to optimize.
The form of this function is not typically known, due to the lack of a full theoretical description of the experiment, and therefore the Jacobian of this function is also unknown.
The input to this function is typically highly dimensional, due to the large number of machine control parameters that affect the output, and these input parameters are coupled in a complex and often non-linear manner.
Evaluation of this unknown function is also relatively costly, meaning that optimization should be as efficient as possible to minimize the required beam time or computational resources.
In addition, the unknown function has some stochasticity, due to the statistical noise in the measurement and also due to the natural variation of unmeasured input parameters, which nonetheless may contribute significantly to variations in the output.
Finally, there are usually constraints placed upon the input parameters due to the operation range of physical devices and machine safety requirements.
These constraints might also be non-trivial due to coupling between different input parameters, and may also depend on system outputs (e.g. to avoid beam loss in an accelerator).

Due to all these considerations, not all optimization algorithms are suitable and only a few different types have been explored in published work, see \cref{tab:opt_summary} for a selection of representative papers. The following sections will focus on these methods. The reader is referred to dedicated reviews, e.g. the comprehensive work by Nocedal and Wright on numerical optimization algorithms\cite{Nocedal.2006}, for further reading.

\begin{table*}
    \centering
    \small
\begin{tabular}{p{2.4cm}|p{2cm}|p{2.0cm}|p{2.9cm}|p{3.9cm}}
\toprule
 \textbf{Author, Year}  & \textbf{Laser type}   &  \textbf{Optimization Method(s)}  &  \textbf{Free Parameters}  & \textbf{Optimization goals}\\
\midrule
He et al., 2015\cite{he2015coherent} & 800 nm Ti:Sa, 15 mJ, 35 fs, 0.5 kHz   &  Genetic algorithm &  deformable mirror (37 actuator voltages)  & Electron angular profile, energy distribution \& transverse emittance, optical pulse compression  \\
Dann et al., 2019\cite{Dann2019PRAB} & 800 nm Ti:Sa, 450 mJ, 40 fs, 5 Hz & Genetic \& Nelder-Mead algorithms & deformable mirror or acousto-optic programmable dispersive filter & Electron beam charge, total charge within energy range, electron beam divergence \\
Shalloo et al., 2020\cite{shalloo2020automation} & 800 nm Ti:Sa, 0.245 J, 45 fs (bandwidth limit), 1 Hz & Bayesian optimization & Gas cell flow rate \& length, laser dispersion ($\partial_\omega^2\phi$, $\partial_\omega^3\phi$, $\partial_\omega^4\phi$), focus position  & Total electron beam energy, Electron charge within acceptance angle, Betatron X-ray counts \\
Jalas et al., 2021\cite{jalas2021bayesian} & 800 nm Ti:Sa, 2.6 J, 39 fs, 1 Hz & Bayesian optimization & Gas cell flow rates (H$_2$ front and back, N$_2$); focus position and laser energy  & Spectral charge density \\
\bottomrule
\end{tabular}
    \caption{Summary of a few representative papers on machine-learning-aided optimization in the context of laser-plasma acceleration and high-power laser experiments.}
    \label{tab:opt_summary}
\end{table*}

\subsection{General concepts}
\subsubsection{Objective functions}\label{scn:objective_functions}
Most optimization algorithms are based on maximizing or minimizing the value of the objective function, which has to be constructed in a way that it represents the actual, user-defined objective of the optimization problem. Typically, the objective function produces a single value, where higher (or lower) values represent a more optimized state\footnote{Depending on the context of optimization problems this function is referred to using many different names, e.g. \emph{merit function} (using a figure of merit), \emph{fitness
function} (in the context of evolutionary algorithms), \emph{cost} or \emph{loss function} (in deep learning) or \emph{reward function} (reinforcement learning)). Some of these are associated with either maximization (reward, fitness, ...) and others with minimization (cost, loss). Here, we will use the general term \emph{objective function} as used in optimization theory.}.
The optimization problem is then a case of finding the parameters which maximize (or minimise) the objective function.

When the objective is to construct a model, the objective function encodes some measure of similarity between the model and what it is supposed to represent, e.g. a measure of how well the model fits some training data. For example, deep learning algorithms minimize an objective function that encodes the difference between desired output values for a given input and actual results produced by the algorithm. A common, basic metric for this is the \emph{Mean Squared Error (MSE)} cost function, in which the difference between predicted and actual value is squared. We already mentioned this metric at several points of this review, e.g. \cref{sec:leastsquares}. The MSE objective function belongs to a class of distance metrics, the $\ell$-norms, which are defined by
\begin{equation}
    \ell_p (x,y)= \left(\sum \|x - y\|^p\right)^{1/p},
    \label{eq:l2norms}
\end{equation} 
where $x$ and $y$ are vectors, and we use the notation $\ell_2 = MSE$. One can also use other $\ell$-norms, e.g., $\ell_1$, which is more robust to outliers than $\ell_2$ since it penalizes large deviations less severely. Note that the number of non-zero values is sometimes referred to as "$\ell_0$" norm (see \cref{compressed}), even though it does not follow from \cref{eq:l2norms}. 

Another popular similarity measure is the \emph{Kullbach-Leibler} (KL) Divergence, sometimes called \emph{relative entropy}, which is used to find the closest probability distribution for a given model. It is defined as
\begin{equation}
    \mathrm{KL}(p|q) = \sum p(x) \log\frac{p(x)}{q(x)},
    \label{eq:kl}
\end{equation}
where $p(x)$ is the probability of observing the value $x$ according to the model, and $q(x)$ is the actual probability of observing $x$. The KL Divergence is minimized when the model prediction $p(x)$ is as close to the actual distribution $q(x)$ as possible. It is a relative measure, i.e., it is only defined for pairs of probability distributions. The KL divergence is closely related to the \emph{Cross Entropy} cost function,

\begin{equation}
    \mathrm{CE}(p, q) = -\sum p(x) \log q(x) = \mathrm{KL}(p|q) - {H}(p),
    \label{eq:ce}
\end{equation}
where $p(x)$ and $q(x)$ are probability distributions, and ${H}(p) = -\sum p(x) \log p(x)$ is the Shannon entropy of the distribution $p(x)$.

Other objective functions may rely on the maximization or minimization of certain parameters, e.g. the beam energy or energy-bandwidth of particle beams produced by a laser-plasma accelerator. Both of these are examples of what is sometimes referred to as \emph{summary statistics}, as they condense information from more complex distributions, in this case the electron energy distribution. While simple at the first glance, these objectives need to be properly defined and there are often different ways to do so\cite{Irshad2022MO}. In the example above, energy and bandwidth are examples for the central tendency and the statistical dispersion of the energy distribution, respectively. These can be measured using different metrics such as weighted arithmetic or truncated mean, the median, mode, percentiles and so forth for the former; and full width at half maximum, median absolute deviation, standard deviation, maximum deviation, etc. for the latter. Each of these seemingly similar measures emphasises different features of the distribution they are calculated from, which can affect the outcome of optimization tasks. Sometimes one might also want to include higher order momenta as objectives, such as the skewness, or use integrals, e.g. the total beam charge.

\subsubsection{Pareto optimization}\label{Pareto Opt}

\begin{figure}
    \centering
    \includegraphics[width=\linewidth]{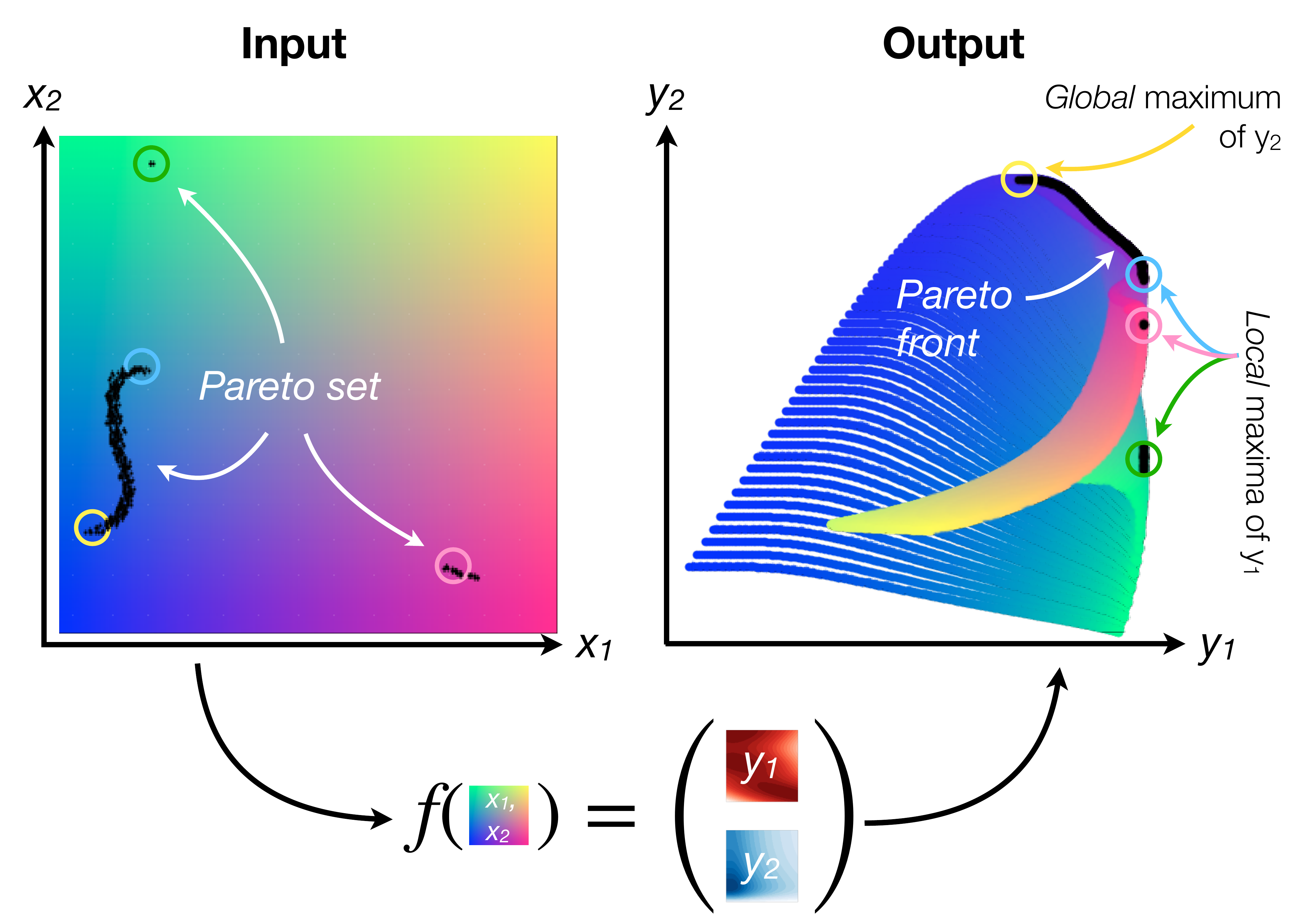}
    \caption{\textbf{Pareto front.} Illustration how a multi-objective function $f( x)=y$ acts on a two-dimensional input space $x = (x_1,x_2)$ and transforms it to the objective space $ y = (y_1,y_2)$ on the right. The entirety of possible input positions is uniquely color-coded on the left and the resulting position in the objective space is shown in the same color on the right. The Pareto-optimal solutions form the Pareto front, indicated on the right, whereas the corresponding set of coordinates in the input space is called the Pareto set. Note that both Pareto front and Pareto set may be continuously defined locally, but can also contain discontinuities when local maxima get involved. Adapted from Irshad \emph{et al.} \cite{Irshad.2021}.
    }
    \label{fig:pareto}
\end{figure}

In practice, optimization problems often constitute multiple, sometimes competing objectives $g_i$. As the objective function should only yield a single scalar value, one has to condense these objectives in a process known as \emph{scalarization}. Scalarization can for instance take the form of a weighted product $g = \prod g_i^{\alpha_i}$ or sum $g= \sum \alpha_i g_i$ of the individual objectives $g_i$ with the hyperparameters $\alpha_i$ describing the weight. Another common scalarization technique is \emph{$\epsilon$-constraint} scalarization, where one seeks to reformulate the problem of optimizing multiple objectives into a problem of single-objective optimization conditioned on constraints. In this method the goal is to optimize one of the $g_i$ given some bounds on the other objectives. All of these techniques introduce some explicit bias in the optimization which may not necessarily represent the desired outcome. Because of this, the hyperparameters of the scalarization may have to be optimized themselves by running optimizations several times.

A more general approach is \emph{Pareto optimization}, where the entire vector of individual objectives $g = (g_1,\dots, g_N)$ is optimized. To do so, instead of optimizing individual objectives, it is based on the concept of \emph{dominance}. A solution is said to dominate other solutions if it is both at least as good on all objectives and strictly better than the other solutions on at least one objective. Pareto optimization uses implicit scalarization by building a set of non-dominated solutions, called the \emph{Pareto front}, and maximizing the diversity of solutions within this set. The latter can be for instance quantified by the hypervolume of the set, or the spread of solutions along each individual objective. As it works on the solution for all objectives at once, Pareto optimization is commonly referred to as \emph{multi-objective} optimization. An illustration of both the Pareto front and set is shown in \cref{fig:pareto}, where a multi-objective function $f$ ''morphs'' the input space into the objective space. In this example $f$ is a modified version of the Branin-Currin function \cite{dixon1978,currin1991}, exhibiting a single, global maximum in $y_2$ but multiple local maxima in $y_1$. The individual two-dimensional outputs $y_1 = f_1(x_1,x_2)$, with $f_1$ being the Branin function, and $y_2=f_2(x_1,x_2)$, $f_2$ being Currin function, are depicted with red and blue colormaps at the the bottom. 

\begin{figure*}[p]
    \centering
    \includegraphics[width=.9\linewidth]{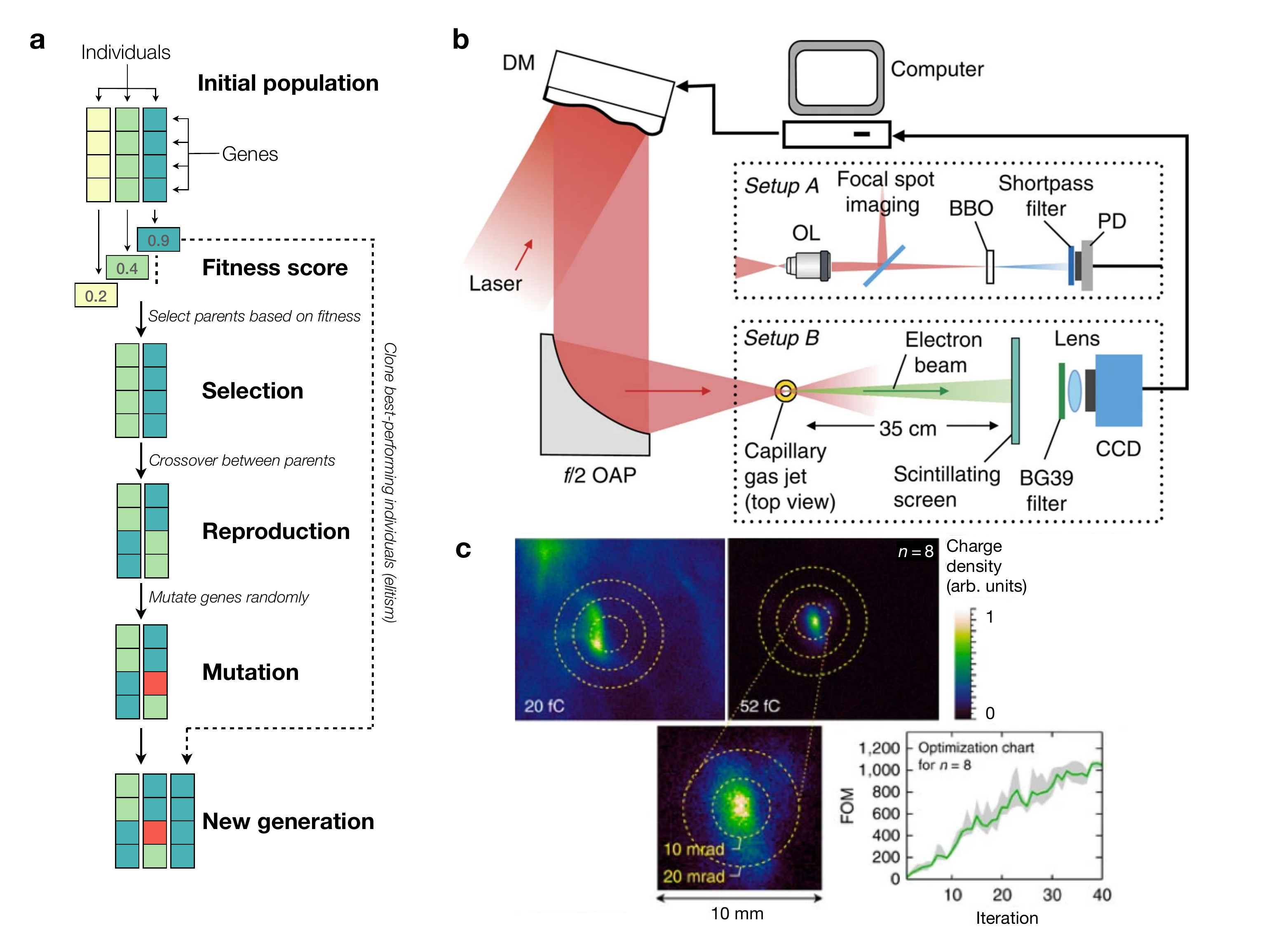}
    \caption{\textbf{Genetic algorithm optimization.} (a) Basic working principle of a genetic algorithm. (b) Sketch of a feedback-optimized LWFA via genetic algorithm. (c) Optimized electron beam spatial profiles using different FoM. Figure (b) and (c) adapted from He et al.\cite{he2015coherent}.
    }
    \label{fig:genetic}
\end{figure*}

\subsection{Grid search and random search}

Once an objective is defined, we can try to optimize it. One of the simplest methods to do so is a grid search, where the input parameter space is explored by taking measurements in regularly-spaced intervals. This technique is particularly simple to implement, especially in experiments, and therefore remains very popular in the setting of experimental laser physics. However, the method severely suffers from the "curse of dimensionality", meaning that the number of measurements increases exponentially with the number of dimensions considered. In practice, the parameter space therefore has to be low-dimensional (one-, two- or three-dimensional at most) and it is applied to the optimization of selected parameters that appear to influence the outcome the most.
One issue with grid scans is that they can lead to aliasing, i.e. high-frequency information can be missed due to the discrete grid with a fixed sampling frequency.

A popular variation, in particular in laser-plasma experiments, is the use of sequential 1-D line searches. Here, one identifies the optimum in one dimension, then performs a scan of another parameter and moves to its optimum, and so forth. This method can converge much faster to an optimum, but is only applicable in settings with a single, global optimum. 

Random search is a related method where the sampling of the input parameter space is random instead of regular. This method can be more efficient than grid search, especially if the system involves coupled parameters and has a lower effective dimensionality \cite{Bergstra.2012}. It is therefore often used in optimization problems with a large number of free parameters. However, purely random sampling also has drawbacks. For instance, it has a tendency to cluster, i.e. to sample points very close to another while leaving some areas unexplored. This behavior is undesirable for signals without high-frequency components and instead one would rather opt for a sampling that explores more of the parameter space. For this case, a variety of schemes exist that combine features of grid search and random search. Two popular examples are jittered sampling and Latin hypercube sampling \cite{Kollig.2002}. For the former samples are randomly placed within regularly-spaced intervals, while the latter does so while maintaining an even distribution in the parameter space.

Grid and random search are often used for initial exploration of a parameter space to seed subsequent optimization with more advanced algorithms. An example for this is shown in \cref{fig:Opt_Comparison}a, where grid search is combined with the downhill simplex method discussed in the next section.

\subsection{Downhill simplex method and gradient-based algorithms}\label{gradient}

In the downhill simplex method, also known as Nelder-Mead algorithm\cite{Nelder.1965}, an array of $(n+1)$ input parameter sets from an $n$-dimensional space is evaluated to get the corresponding function values.
The worst-performing evaluation point is modified at each iteration by translating it towards or through the center of the simplex. This continues until the simplex converges to a point at which the function is optimized. The method is simple and efficient, which is why it is popular in various optimization settings. The convergence speed crucially depends on the initial choice of the simplex, with a large distance between input parameters leading to a more global optimization, while small simplex settings result in local optimization.

In the limit of small simplex size, the Nelder-Mead algorithm is conceptually related to gradient-based methods for optimization. The latter are based on the concept of using the gradient of the objective function to find the direction of the steepest slope. The objective function is then minimized along this direction using a suitable algorithm such as gradient descent. Momentum descent is a popular variation of gradient descent where the gradient of the function is multiplied by a value, in analogy to physics called \emph{momentum}, before being subtracted from the current position. This can help the algorithm converge to the local minimum faster. These methods typically require more and smaller iterations than the downhill simplex method, but can be more accurate.

In both cases, measurement noise can easily result in a wrong estimation of the gradient. In the setting of laser-plasma experiments, it is therefore important to reduce such noise, e.g. by taking several measurements at the same position. While this may be possible when working with high-repetition-rate systems, as was demonstrated by Dann \emph{et al.} \cite{Dann2019PRAB},  gradient-based methods are in general less suitable to be used in high-power-laser settings. Other popular derivative-based algorithms are the \emph{conjugate gradient} (CG) methods, \emph{quasi-Newton} (QN) methods,
and the \emph{limited-memory Broyden-Fletcher-Goldfarb-Shanno} (L-BFGS) method, all of which are discussed by Nocedal and Whright\cite{Nocedal.2006}.

\subsection{Genetic algorithms}\label{scn:genetic}

\begin{figure}
    \centering
    \includegraphics[width=.85\linewidth]{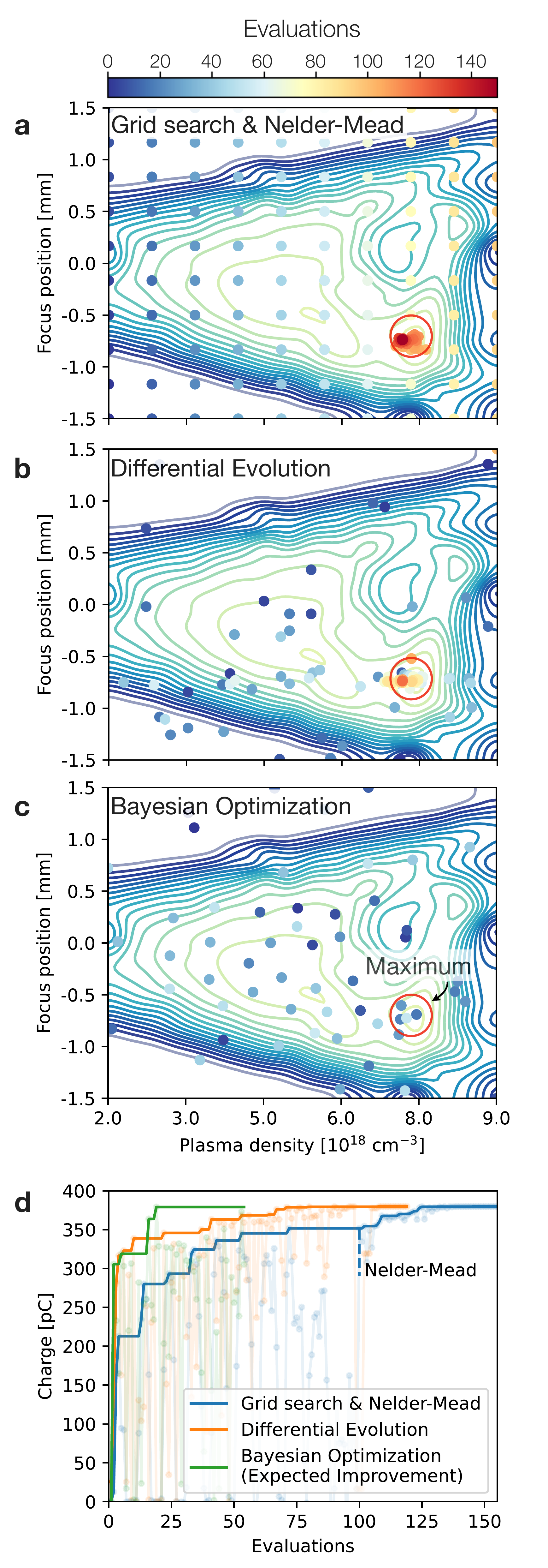}
    \caption{\textbf{Illustration of different optimization strategies} for a non-trivial two-dimensional system, here based on a simulated laser wakefield accelerator with laser focus and plasma density as free parameters. \textit{[Continued on next column ...]}}
    \label{fig:Opt_Comparison}
\end{figure}

One of the first family of algorithms to find applications in the field laser-plasma acceleration was genetic algorithm. As a subclass of evolutionary methods, these nature-inspired algorithms start with a pre-defined or random set, called \textit{population}, of measurements for different input settings. Each free parameter is called a \textit{gene} and, similar to natural evolution, these genes can either \textit{cross over} between most successful settings or randomly change (\textit{mutate}). This process is guided by a so-called fitness function, which is designed to yield a single-valued figure of merit that is aligned with the optimization goal. Depending on the objective, the individual measurements are ranked from most to least fit. The fittest ``individuals'' are then used to spawn a new generation of ``children'', i.e. a new set of measurements based on crossover and mutation of the parent genes. A popular variation to genetic algorithms is differential evolution\cite{storn1997differential}, which employs a different type of crossover.  Instead of crossing over two parents to create a child, differential evolution uses three parent measurements. The child is then created by adding a weighted difference between the parents to a random parent. This process is repeated until a new generation is created, see \cref{fig:Opt_Comparison}b for an example.

One particular strength of genetic algorithms compared to many other optimization methods is their ability to perform multi-objective optimization, i.e. when multiple, potentially conflicting, objectives are to be optimized. A popular example would be the nondominated sorting genetic algorithms (NSGAs)\cite{Deb.2000}. Here the name-giving sorting technique ranks the individuals by their population dominance, i.e. how many other individuals in the population the individual dominates. Other multi-objective approaches are for instance based on optimizing \emph{niches}, similar-valued regions of the Pareto front\cite{horn1994niched}.

It should be noted that genetic algorithms intrinsically operate on population batches and not on a single individual. While this can be beneficial for parallel processing in simulations, it can make it more difficult to employ in an online optimization setup. 

\example{Genetic algorithms have been used since the early 2000s to control high-harmonic generation\cite{Bartels.2000,Yoshitomi.2004}, cluster dynamics\cite{Moore.2005,Zamith.2004} or ion-acceleration\cite{nayuki2005production}. An influential example in the context of high-power lasers was published by He \emph{et al.} in 2015\cite{he2015coherent}, and a sketch of their feedback-looped experimental setup is presented in Fig. \ref{fig:genetic}. In the paper, a genetic algorithm is used to optimize various parameters of a laser-plasma accelerator, namely the electron beam angular profile, energy distribution, transverse emittance and optical pulse compression. This was done by controlling the voltage on 37 actuators of a deformable mirror (DM), which was used to shape a laser pulse before it entered a gas jet target. The genetic algorithm was initialized with a population of 100 individuals and each subsequent generation was generated based on the 10 fittest individuals\cite{He.2015y53}. A similar experiment was performed in a self-modulated LWFA\cite{lin2019adaptive} driven by a mid-infrared laser pulse. The electron beam charge, energy spectra, and beam pointing have been optimized. The combination of genetic algorithm and deformable mirror has been applied to other high-power laser experiments as well\cite{hah2017enhancement, lin2018focus, noaman2018controlling, finney2021filament, englesbe2021optimization}.

Streeter \emph{et al.} used a similar technique to optimize ultrafast heating of atomic clusters at Joule-level energies \cite{Streeter.2018}, but instead of controlling the spatial wavefront, they controlled the spectral phase up to its fourth order. The genetic algorithm was based on a population of 15 individuals and $4-8$ children generations were evaluated. This method can also be used for optimizing other laser parameters such as focal spot size, focus position, and chirp \cite{Dann2019PRAB}.

Another example is the use of differential evolution for optimization of the laser pulse duration in a SPIDER and DAZZLER feedback loop, as presented by D. Peceli et al.\cite{peceli}.
The genetic algorithm method has also been employed in ion acceleration in a laser-plasma accelerator, where Smith \emph{et al.} \cite{smith2020optimizing} optimized the conversion efficiency from laser energy to ion energy by exploring thousands of target density profiles in 1-D PIC simulations.}

\subsection{Bayesian optimization}\label{BO}

Bayesian optimization (BO)\cite{Shahriari.2015,frazier2018tutorial} is a model-based global optimization method that uses probabilistic modeling, which was discussed in \cref{probabilistic}. The strength of BO lies in its efficiency with regard to the number of evaluations. This is particularly useful for problems with comparably high evaluation costs or long evaluation times. To achieve this, BO uses the probabilistic surrogate model to make predictions about the behavior of the system at new input parameter settings, providing both a value estimate and an uncertainty. 

At the core of BO lies what is called the acquisition function; a pre-defined function that is suggest the next points to probe on a probabilistic model. The latter is usually\footnote{While most work on BO is done using GP regression, the method is in principle model agnostic. This means that other types of (probabilistic) surrogate models of the system can be employed, such as decision trees (see \cref{trees}) or deep neural networks (see \cref{neural_nets}).} a Gaussian process fitted from training points, see \cref{GPP}, thus providing a cheap-to-evaluate surrogate function. A simple and intuitive example of an acquisition function is the upper confidence bound, \begin{equation}
    UCB(x) = \mu(x) + \kappa \sigma(x),
\end{equation}
which weighs the mean prediction $\mu$ versus the variance prediction $\sigma$, with a user-chosen hyperparameter $\kappa$. For $\kappa=0$ the optimization will act entirely exploitatively, i.e. it will move to the position of the highest expected return, whereas a large $\kappa$ incentivises to reduce the uncertainty and explore the parameter space. Other common acquisition functions are the expected improvement \cite{jones1998efficient}, knowledge gradient\cite{frazier2009knowledge} and entropy search\cite{hennig2012entropy}. As the surrogate model can be probed at near-negligible evaluation cost, this optimization can be performed using numerical optimization methods such as the gradient-based methods discussed in \cref{gradient}. The position of the acquisition function's optimum is then used as input parameter setting to evaluate the actual system. This process is repeated until some convergence criteria is achieved, a predefined number of iterations are reached, or the allocated resources have been otherwise exhausted.

Bayesian optimization provides a very flexible framework that can be further adapted to various different optimization settings. For instance, it has proven to be a valid alternative to evolutionary methods when it comes to solving multi-objective optimization problems. The importance of this method for laser-plasma acceleration stems from the fact that many optimization goals, such as beam energy and beam charge, are conflicting in nature and require a definition of a trade-off. The goal of the Pareto-optimization is to find the Pareto-front, which is a surface in the output objective space that is comprised of all the non-dominated solutions (see \cref{Pareto Opt}). A common metric that is used to measure the closeness of a set of points to the Pareto-optimal points is the hypervolume. The BO algorithm works by using the expected hypervolume improvement\cite{Yang.2019} to increase the extent of the current non-dominated solutions, thus optimizing all possible combinations of individual objectives. Note that the Pareto-front, like to the global optimum of single-objective optimization, is usually not known \textit{a priori}. 

Another possible way to extend BO is to incorporate different information sources\cite{marco2017virtual}. This is often done by adding an additional information input dimension to the data model. This method is often referred to as \textit{multi-task} (MT) or \textit{multi-fidelity} (MF) BO. Both terms being used somewhat interchangeably in the literature, although MT often refers to optimization with entirely different systems (codes, etc.), whereas MF focuses on different fidelity (resolution, etc.) of the same information source. The core concept behind these methods is that the acquisition function not only encodes the objective, but also minimizes the measurement cost. These multi-information-source methods have the potential to speed up optimization significantly. They can also be combined with multi-objective optimization, as shown by Irshad et al.\cite{Irshad.2021}.

\clearpage

\begin{figure}
    \centering
    \includegraphics[width=1\linewidth]{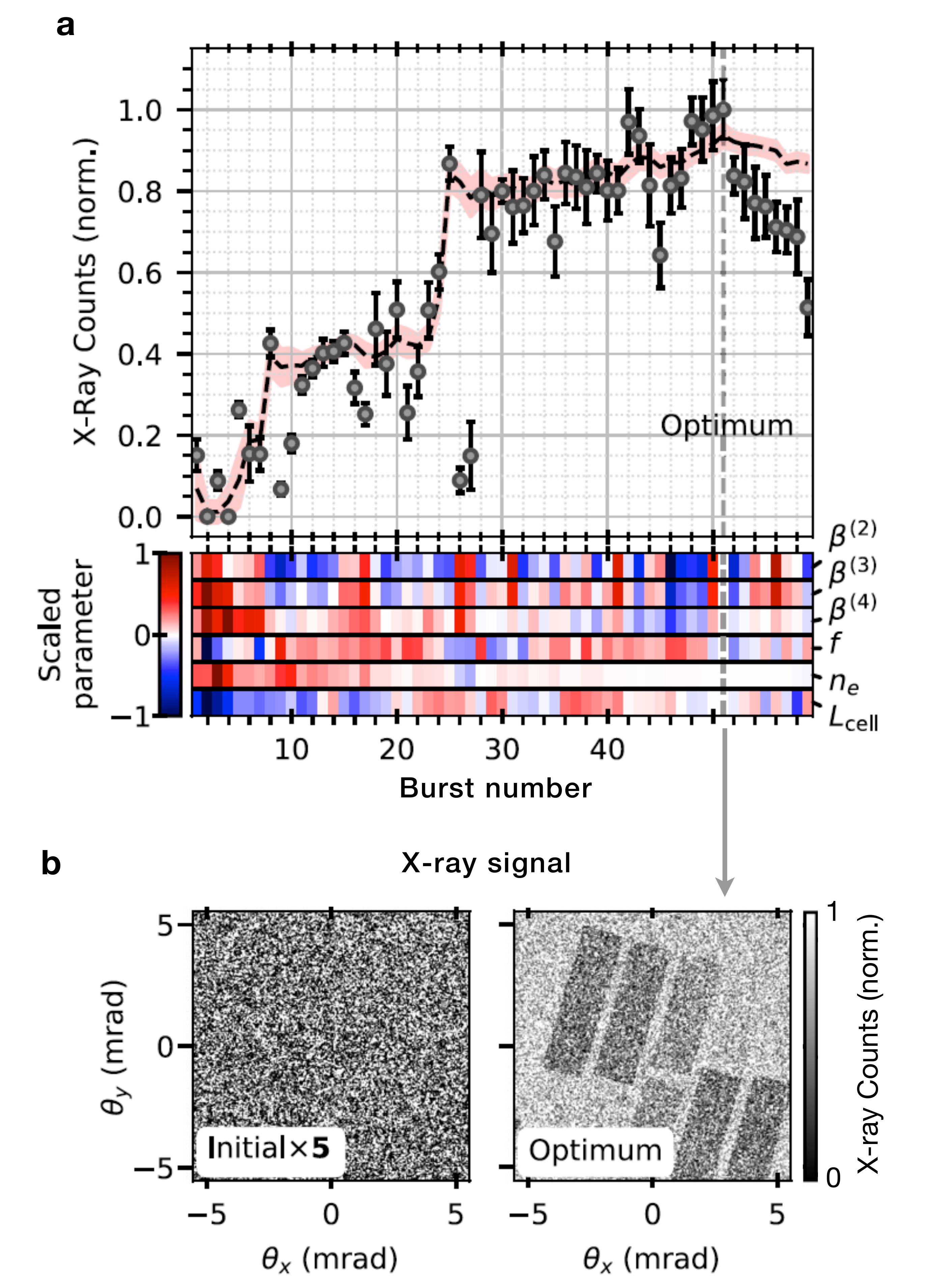}
    \caption{\textbf{Bayesian optimization of a laser-plasma X-ray source.} a) The objective function (x-ray counts) as a function of iteration number (top) and the variation of the control parameters (bottom) during optimization.
    b) X-ray images obtained for the initial (bottom) and optimal (top) settings. 
    Adapted from Shalloo \emph{et al.} \cite{shalloo2020automation}.
    }
    \label{fig:ShallooBayesian}
\end{figure}

\example{
A first demonstration of BO in the context of laser-plasma acceleration was presented by Lehe, who used the method to determine the injection threshold in a set of particle-in-cell simulations\cite{Lehe.2017}. 
The use of BO in experiments was pioneered by Shalloo \emph{et al.}\cite{shalloo2020automation}, who demonstrated optimization of electron and x-ray beam properties from an LWFA by automated control of laser and plasma control parameters.
Another work by Jalas \emph{et al.}\cite{jalas2021bayesian} focused on improving the spectral charge density using the objective function $\sqrt{Q}\tilde E/E_{MAD}$, thus incorporating the beam charge $Q$, the median energy $\tilde E$ and the energy spread defined here by the median absolute deviation $E_{MAD}$.
In contrast to Shalloo \emph{et al.}, they used shot-to-shot measurements of the control parameters to train the model rather than relying on the accuracy of their controllers, reducing the level of output variation attributed purely to stochasticity. 
BO has also been applied to the optimization of laser-driven ion acceleration in numerical simulations \cite{Dolier2022NJP}, and experiments \cite{Loughran2023arxiv}.
A first implementation of multi-objective optimization in numerical simulations of plasma acceleration was published by Irshad \textit{et al.}\cite{Irshad2022MO}, who showed that multi-objective optimization can lead to superior performance than manually trying different trade-off definitions or settings for the individual objectives, in this case beam charge, energy spread and distance to a target energy.
An example of multi-task BO in laser-plasma simulations was recently published by Pousa \textit{et al.}, who combined the Wake-T and FBPIC codes\cite{ferranpousa.2022}, while Irshad \textit{et al.} \cite{Irshad2022MO} used the FBPIC code at different resolutions for multi-fidelity optimization. }

\setcounter{figure}{14}
\begin{figure}
    \caption{\textit{[...]} The total beam charge, shown as contour lines in plots (a)-(c) serves as optimization goal. The position of the optimum is marked by a red circle, located at a focus position of $\SI{-0.74}{\mm}$ and a plasma density of $\SI{8e18}{\per\cubic\cm}$. In panel (a), a grid search strategy with subsequent local optimization using the downhill simplex (Nelder-Mead) algorithm is shown. Panel (b) illustrates differential evolution and (c) is based on Bayesian optimization using the common Expected Improvement acquisition function. The performance for all three examples is compared in panel (d). It shows the typical behavior that Bayesian optimization needs the least and grid search requires the most iterations. The local search via Nelder mead converges within some 20 iterations, but requires a good initial guess (here provided by the grid search). Individual evaluations are shown as shaded dots. Note how the Bayesian optimization starts exploring once it has found the maximum, whereas the evolutionary algorithm tends more towards exploitation around the so-far best value. This behavior is extreme for the local Nelder-Mead optimizer, which only aims to exploit and maximize to local optimum.}
\end{figure}

\begin{figure}
    \centering
    \includegraphics[width=.98\linewidth]{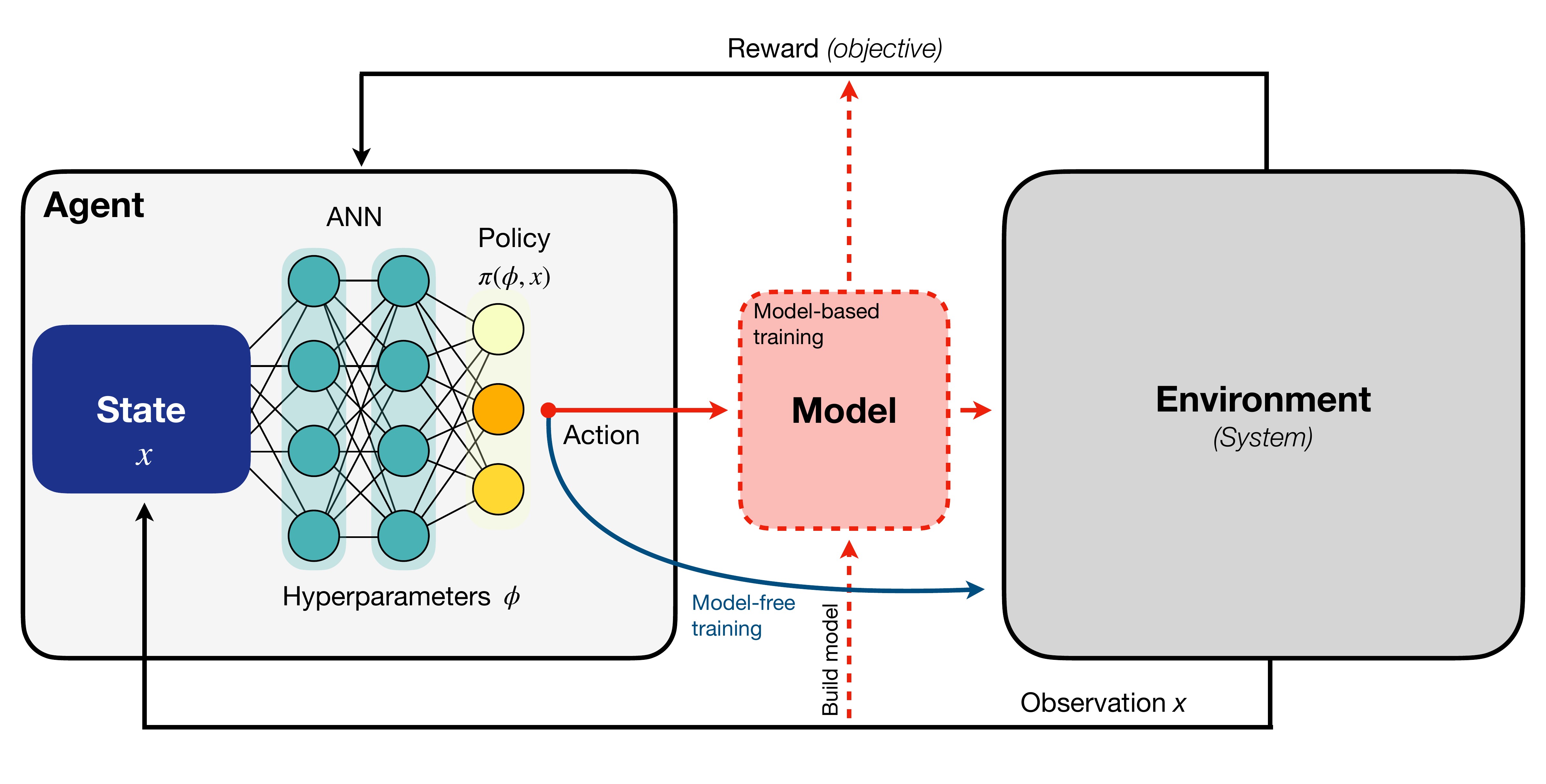}
    \caption{\textbf{Sketch of deep reinforcement learning.} The agent, which consists of a policy and a learning algorithm that updates the policy, send an action to the environment. In the case of model-based reinforcement learning, the action is sent to the model, which is then applied onto the environment. Upon the action to the environment, an observation is made and sent back to the agent as a reward. The reward is used to update the policy via the learning algorithm in the agent, which leads to an action in the next iteration.}
    \label{fig:reinforcement}
\end{figure}

\subsection{Reinforcement learning}

Reinforcement learning (RL)\cite{sutton2018reinforcement} differs fundamentally from the optimization methods discussed so far. RL is a method of learning the optimal behavior (the \emph{policy} $\pi$) to control a system. The learning occurs via repeated trial and error, called \textit{episodes}, where each episode is started in an initial state of the system, and then the agent (i.e., the optimizer) interacts with the system according to its policy. The agent then receives a \textit{reward} signal, which is a scalar value that indicates the success of the current episode and its goal is to maximize the expected reward, analogous to the objective function in other optimization methods. The agent is said to \textit{learn} when it is able to improve its policy to achieve a higher expected reward.

The policy itself has traditionally been represented using a Markov decision process (MDP), but in recent years deep reinforcement learning (DRL) has become widely used, in which the policy is represented using deep neural networks\cite{arulkumaran2017deep,Li.2018.DRL}. However, while we commonly update weights and biases via backpropagation in supervised deep learning, the learning in DRL is done in an \textit{unsupervised} way. Indeed, while the agent is trying to learn the optimal policy to maximize the reward signal, the reward signal itself is unknown to the agent. The agent only knows the reward signal at the end of the episode, so it is not possible to perform backpropagation. Instead, the policy network can for instance be updated using evolutionary strategies (see \cref{scn:genetic}), where the agents achieving the highest reward are selected to create a new generation of agents. Another common approach is to use a so-called \textit{actor-critic} strategy\cite{grondman2012survey}, where a second network is introduced, called the critic. At the end of each episode, the critic is trained to estimate the expected long-term reward from the current state, called the value. This expected reward signal is then used to train the actor network to adjust the policy. The \textit{policy gradient}\cite{sutton1999policy} is a widely used algorithm for training the policy network using the critic network to calculate the expected reward signal.

Reinforcement learning algorithms can be further divided into two main classes: model-free and model-based learning. Model-free methods learn as discussed above directly by trial and error, only implicitly learning about the environment. Model-based methods, on the other hand, explicitly build a model of the environment, which can be used for both planning and learning (somewhat similar to optimization using surrogate models discussed earlier). The arguably most popular method to build models in reinforcement learning is again the use of neural networks, as they can learn complex, non-linear relationships and are also capable of learning from streaming data, which is essential in reinforcement learning. A crucial advantage of the model-based approach is that it can drastically speed up training, although performance is always limited by the quality of the model. In the case of real-life systems this is sometimes referred to as the `reality gap'.

One crucial advantage of reinforcement learning is that once the training process is completed, the computational requirements of running an optimization are heavily reduced. A simplified representation of the reinforcement learning workflow is sketched in Fig. \ref{fig:reinforcement}. 

\example{
An example of a reinforcement learning application is the work of Kain et al.\cite{kain2020sample} for trajectory optimization in the Advanced Wakefield (AWAKE) experiment in PWFA, which found the optimum in just a few iterations based on 300 iterations of training. There are many other examples for the use of reinforcement learning at accelerator facilities, e.g. Gao et al.\cite{Gao.2019}, Bruchon et al.\cite{Bruchon.2020}, O'Shea et al.\cite{OShea.2020}, and John et al.\cite{John.2021}.}

\section{Unsupervised Learning}\label{scn:correlation}

In this section we are going to discuss unsupervised learning techniques for exploratory data analysis. The term `unsupervised' refers to the case where one does not have access to training labels, and therefore the aim is not to find a mapping between training data and labels, as it is often the case for deep learning. Rather, the aim is to detect relationships between features of the data.

For high-power laser experiments, many parameters will be coupled in some way such that there are correlations between different measurable input quantities. For example, increasing the laser energy in the amplifier chain of a high-power laser can affect the laser spectrum or beam profile. To understand the effect a change to any one of these parameters will have, it is important to consider their correlation. However, an experimental setup can easily involve tens of parameters and interpreting correlations becomes increasingly difficult. In this context, it can be useful to distill the most important (combinations of) parameters in a process called dimensionality reduction. The same method also plays a crucial role in efficient data compression, which is becoming increasingly important due to the large amount of data produced in both experiments and simulation. These methods are also closely related to \emph{Pattern Recognition}, which addresses the issue of automatically discovering patterns in data. 

\subsection{Clustering}
Data clustering is a machine learning task of finding similar data points and dividing them into groups, even if the data points are not labeled. This can for instance be useful to separate signal from background in physics experiments.

A popular \emph{centroid-based clustering} algorithm is the $K$-means algorithm, which consists of two steps: First, the algorithm randomly assigns a cluster label to each point. Then, in a second step, it calculates the center point of each cluster and re-assigns the cluster label to each point based on the proximity to the cluster center. This process is repeated until the cluster assignment does not change anymore. The $K$ in the algorithm's name represents the number of clusters, which can be guessed or - more quantitatively - be estimated using methods such as the "silhouette" value or the "elbow method" \cite{rousseeuw1987silhouettes}. As the method is quite sensitive to the initial random choice of the cluster assignment, it is often run several times with different initial choices to find the optimal classification. 

In contrast to centroid-based clustering, in which each cluster is defined by a center point, in \emph{distribution-based clustering}, each cluster is defined by a (multivariate) probability distribution. In the simplest case this can be a Gaussian distribution with a certain mean and variance for each cluster. More advanced methods use a Gaussian mixture model (GMM), in which each cluster is represented as a combination of Gaussian distributions. A popular distribution-based clustering method is the expectation maximization (EM) algorithm \cite{dempster1977maximum}.

\begin{figure}
    \centering
    \includegraphics[width=\linewidth]{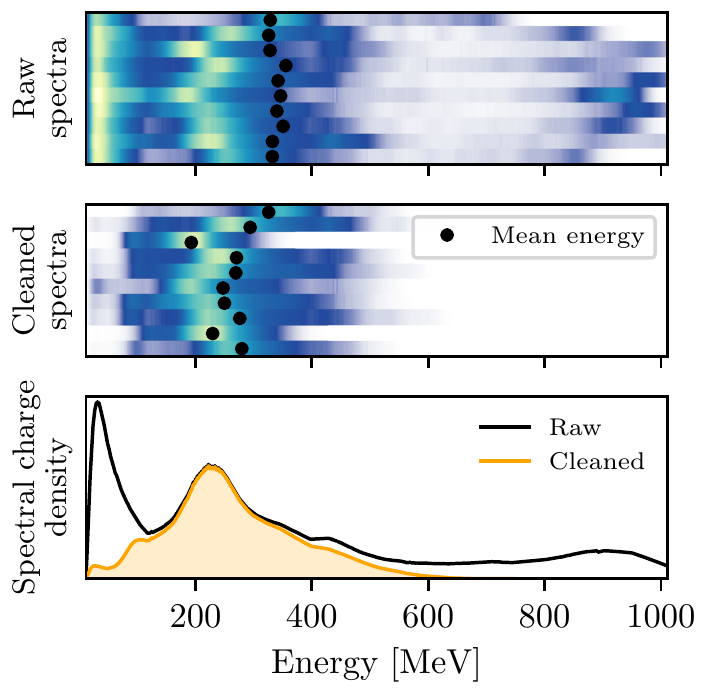}
    \caption{\textbf{Data treatment using a Gaussian Mixture Model (GMM)}. \textit{Top:} 10 consecutive shots from a laser wakefield accelerator. \textit{Middle:} The same shots using a GMM to isolate the spectral peak at around 250 MeV. \textit{Bottom:} Average spectra with and without the GMM cleaning. From Irshad et al.\cite{irshad2023pareto}.}
    \label{fig:GMM}
\end{figure}

\example{
An example for the application of a GMM in data processing is shown in \cref{fig:GMM}. There a number of energy spectra from a laser wakefield accelerator are displayed. As the spectra exhibit multiple peaks, standard metrics such as the mean and standard deviation are not characteristic of neither the peaks' positions nor their widths. To avoid this problem a mixture model is used that isolate the spectral peaks. To this end a combination of Gaussian distributions is fitted to the data and then spectral bin is assigned with a certain probability to each distribution.}

\subsection{Correlation analysis}

A simple method for exploring correlations is the correlation matrix, which is a type of matrix that is used to measure the relationship between two or more variables. We can calculate its coefficients, also known as Pearson correlation coefficients, on a set of $n$ measurements of each pair of parameters $x_i$ and $y_i$ as
\begin{align}
    r_{xy} =\frac{\sum ^n _{i=1}(x_i - \bar{x})(y_i - \bar{y})}{\sqrt{\sum ^n _{i=1}(x_i - \bar{x})^2} \sqrt{\sum ^n _{i=1}(y_i - \bar{y})^2}} \;.
\end{align}
where $\bar{x}$ and $\bar{y}$ are the mean values of the variables $x$ and $y$, respectively. The correlation coefficient $r$ is a number between -1 and 1. A value of 1 means that two variables are perfectly correlated, while a value of -1 means that two variables are perfectly anti-correlated. A value of 0 means that there is no correlation between two variables. 

\begin{figure}[tb]
    \centering
    \includegraphics[width=\linewidth]{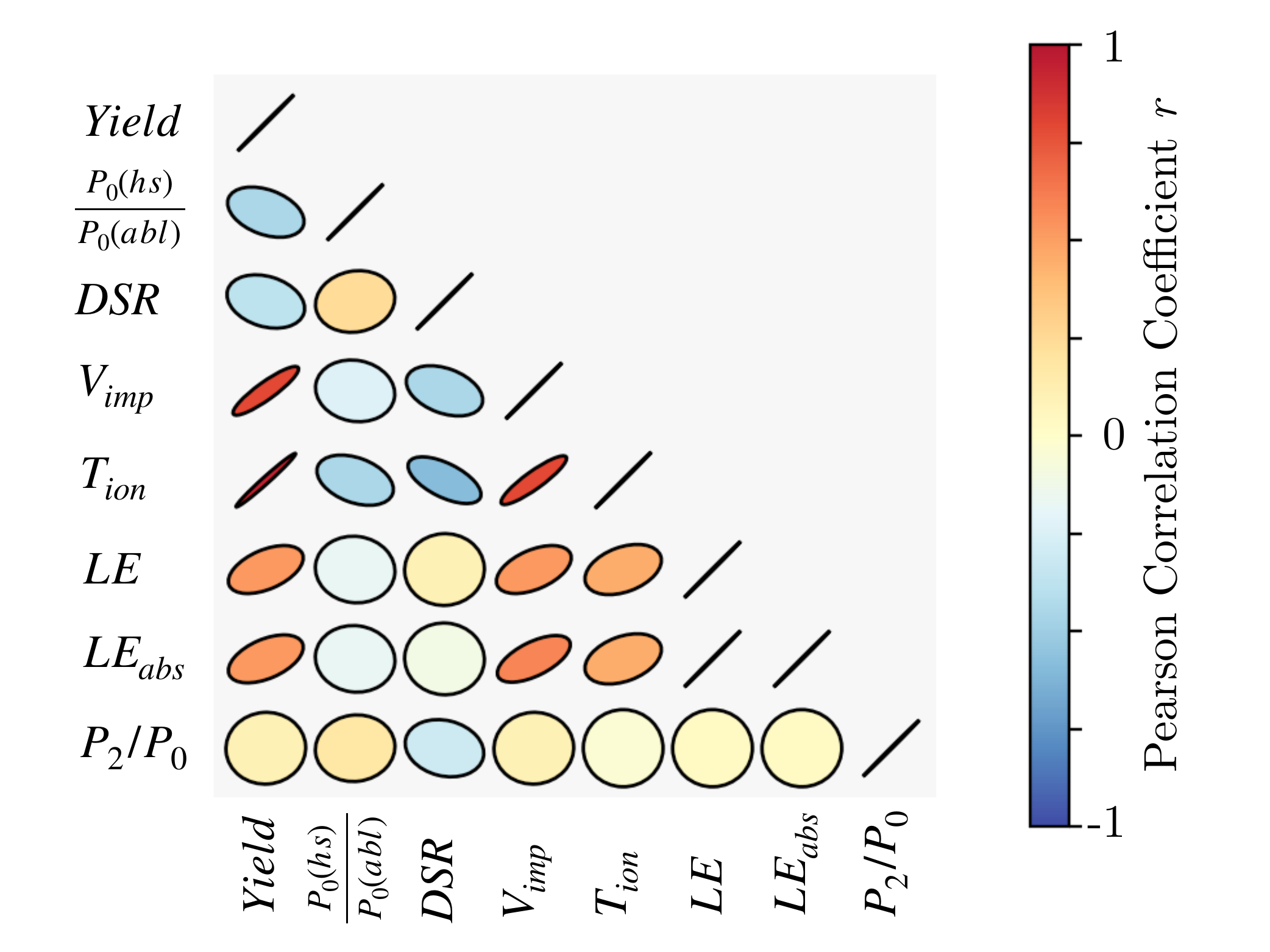}
    \caption{\textbf{Correlogram} - a visualization of the correlation matrix - of different variables versus yield at NIF. Color indicates the value of the correlation coefficient. In this particular representation the correlation is also encoded in the shape and angle of the ellipses, helping intuitive understanding. The strongest correlation to the fusion yield is observed with the implosion velocity $V_{imp}$ and the ion temperature $T_{ion}$. There is also a clear anti-correlation observable between down-scattered ratio (DSR) and $T_{ion}$ and, in accordance with the previously stated correlation of $T_{ion}$ and yield, a weak anti-correlation of DSR and yield. Note that all variables perfectly correlate with themselves by definition. Plot generated based on data presented in Hsu \emph{et al.}\cite{hsu2020analysis}. Further explanation (labels etc.) can be found therein.}
    \label{fig:Correlation_Hsu}
\end{figure}

The correlation matrix and its visualization, sometimes called \emph{correlogram}, allow for a quick way to look for interesting and unexpected correlations. An example for this is shown in \cref{fig:Correlation_Hsu}. Note that by reducing correlations to a single linear term one can miss more subtle or complex relationships between variables. For such cases more general measures of correlation exist, such as the Spearman correlation coefficient that measures how well two variables can be described by any monotonic function.

\subsection{Dimensionality reduction}\label{scn:dimensionality_redcution}

Many data sets are high-dimensional data but are governed by few important underlying parameters. Signal decomposition and \emph{dimensionality reduction} are processes that reduce the dimensionality of the data by separating a signal into its components or projecting it onto a lower-dimensional subspace so that the essential structure of the data is preserved. There are many ways to perform dimensionality reduction, two of the most common ones being principal component analysis and autoencoders.

\textit{Principal component analysis (PCA)} is a very popular linear transformation technique that is used to convert a set of observations of possibly correlated variables into a (smaller) set of values of linearly uncorrelated variables, called the principal components. This transformation is defined in such a way that the first principal component has the largest possible variance, and each succeeding component in turn has the highest variance possible under the constraint that it is orthogonal to the preceding components. One method to perform PCA is to use singular value decomposition  (SVD), which is used to decompose the matrix of data into a set of eigenvectors and eigenvalues. PCA shares a close relationship with correlation analysis, as the eigenvectors of the correlation matrix match those of the covariance matrix, which is utilized in defining PCA. Additionally, the eigenvalues of the correlation matrix equate to the squared eigenvalues of the covariance matrix, provided that the data has been normalized. Kernel PCA\cite{scholkopf1997kernel} is an extension of PCA that uses a nonlinear transformation of the data to obtain the principal components. A relatively new variation, with some relation to the priorly discussed compressed sensing (\cref{compressed}), is robust principal component analysis (RPCA)\cite{candes2011robust}. RPCA is a modification of the original algorithm, which is better suitable to handle the presence of outliers in data sets. PCA should not be confused with the similarly named independent component analysis (ICA)\cite{hyvarinen2000independent}, which is a popular technique to decompose a multivariate signal into a sum of statistically independent non-Gaussian signals.

There are also many neural network approaches to dimensionality reduction, one of the most popular ones being \textit{autoencoders (AEs)}. The purpose of an AE is to learn an approximation to the identity function, i.e. a function that reproduces the input. In a standard AE, a neutral network is created with an intermediate bottleneck layer with a reduced number of nodes, known as the \emph{latent space}.
During training, the neural network hyperparameters are optimized so that the output matches the input as closely as possible, typically by minimizing the mean-squared error. 
Due to the bottleneck, the autoencoder automatically discovers an efficient representation for the data in the latent space. 
The hidden layers up to the latent space are known as the \emph{decoder} and the the hidden layers from the latent space to the output layer are the \emph{encoder}.
The encoder can then be used separately to perform dimension reduction, equivalent to lossy data compression.
With the corresponding decoder, an approximation of the original data can be extracted from its latent space.

There exist many different types of autoencoder architecture, a particularly popular one being variational autoencoders (VAEs). VAEs replace deterministic encoder-decoder layers with a stochastic architecture (c.f. \cref{fig:neuralNet}) to allow the model to provide a probability distribution over the latent space. As a result, a VAEs latent space is smooth and continuous in contrast to a standard AEs latent space, which is discrete. This allows VAEs to also generate new data by sampling from the latent space. 

Autoencoders have also shown a strong potential as advanced compression techniques that can be highly adapted to many kinds of inputs. In this case, one trains an AE model to find an approximation of the identity function for some raw data. After training, the raw data is sent through the encoding layer and only the dimensionality-reduced, highly-compressed latent space representation in the bottleneck layer is saved. Decompression is achieved by sending the data through the decoding layer. This method is not only relevant to reduce disc space occupied by data. Autoencoders are nowadays frequently used as an integral part of complex machine learning tasks, where the latent space is used a lower-dimensional input for e.g. a diffusion network (as part of what is called a latent diffusion model\cite{rombach2022high}) or a Bayesian optimizer\cite{griffiths2020constrained}.

\example{An example of the modeling using the latent space of an autoencoder was recently publihsed by Willmann \emph{et al.}\cite{willmann2021data}. Working on the problem of simulating shadowgrams from plasma probing, they used an autoencoder to reduce the three-dimensional input data and then trained a small four-layer perceptron network to learn how to approximate the shadowgram formation. An example of pure compression was recently presented by Stiller \emph{et al.} who applied an autoencoder to compress data from particle-in-cell simulations, showing promising first results\cite{stiller2022continual}. }

\section{Image analysis}\label{classification}

\begin{figure}
    \centering
    \includegraphics[width=\linewidth]{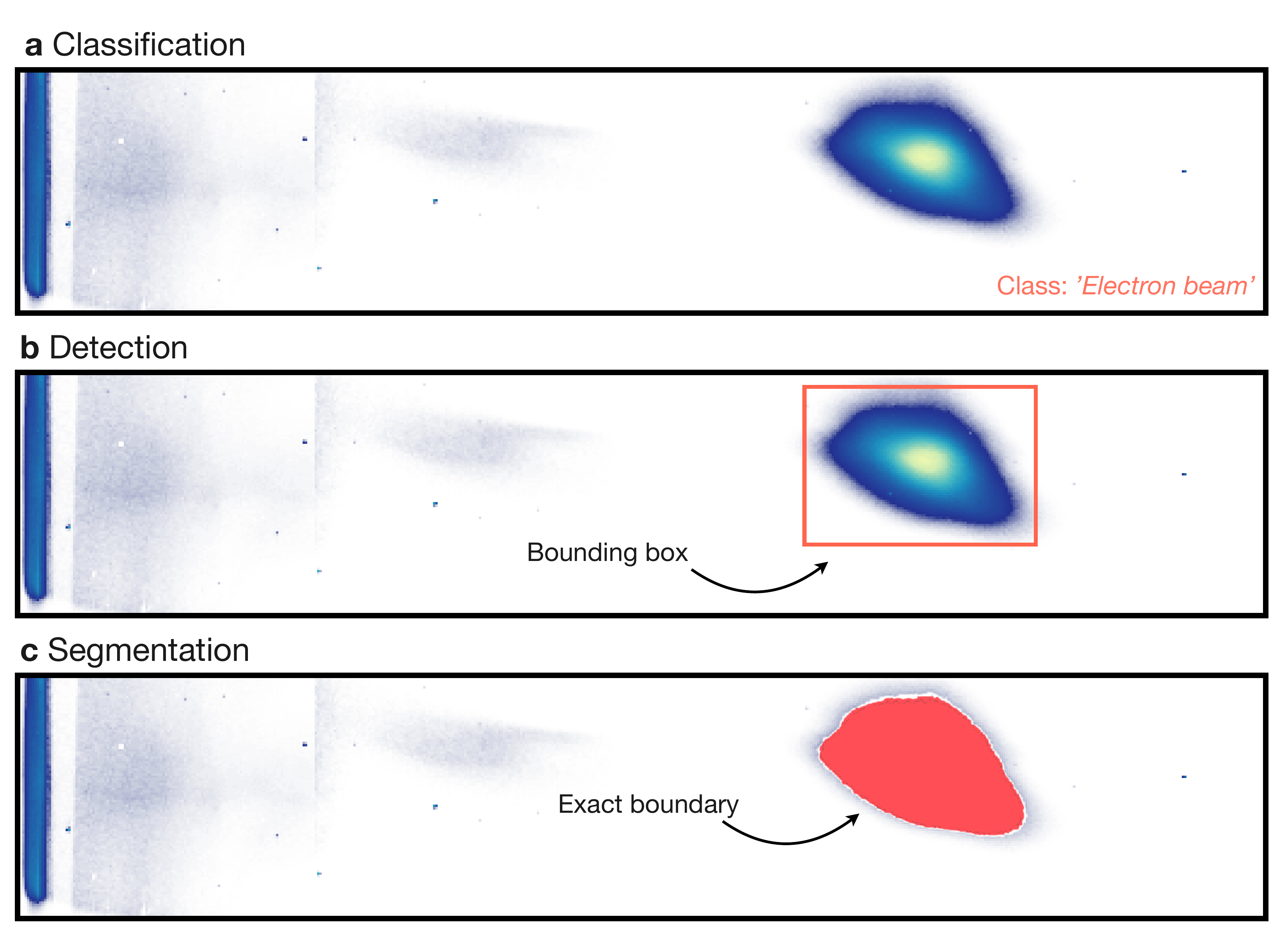}
    \caption{\textbf{Illustration of common computer vision tasks}. (a) Classification is used to assign (multiple) labels to data. (b) Detection goes a step further and adds bounding boxes. (c) Segmentation provides a pixel maps with exact boundaries of the object or feature. }
    \label{fig:classificationDetection}
\end{figure}

In the previous section we have discussed how to analyze datasets by looking for correlations or compressed representations. A closely related group of tasks occurs when dealing with image data, i.e. image recognition or classification, object detection and segmentation. While the methods in the previous section dealt with features learnt from the data itself, the techniques discussed in this section aim to identify or locate \emph{specific} features in our data (in particular images). As such, these are all considered \emph{supervised} learning methods.

\subsection{Classification}

The classification problem in machine learning is the problem of correctly labeling a data point from a given set of data points with the correct label. The data points are labeled with a categorical class label, such as ``cat'', ``dog'', ``electron'' or ``ion'', see \cref{fig:classificationDetection}a.

In the following we are going to briefly discuss some of the most important machine-learning techniques used for classification. It should be noted that classification is closely related to regression, with the main difference being essentially that the model's output is a class value instead of a value prediction. As such, methods working in regression can in general also be applied to classification tasks. One example is the decision tree method, which we already discussed in \cref{trees}. 

\subsubsection{Support vector machines}

Support vector machines (SVMs) are a popular set of machine learning methods used primarily in classification and recognition. For a simple binary classification task, an SVM draws a hyperplane that divides all data points of one category from all data points of the other category. While there could be many hyperplanes, an SVM looks for the optimal hyperplane that best separates the dataset by maximizing the margin between the hyperplane and the data points in both categories. The points that locate right on the margin are called ``supporting vectors''. For a dataset with more than two categories, classification is performed by dividing the multi-classification problem into several binary classification problems and then finding a hyperplane for each. In practice, the data points in two categories can mix together so that they can not be clearly divided by a linear hyperplane. For such nonlinear classification tasks, the kernel trick is used to compute the nonlinear features of the data points and map them to a high dimensional space, so that a hyperplane can be found to separate the high dimensional space. The hyperplane will then be mapped back to the original space.

The ideal application scenario for an SVM is for datasets with small samples but high dimensions. The choice of various kernel functions also adds to the flexibility of this method. However, an SVM would be very computationally expensive for large datasets. Besides, its accuracy can significantly decrease when analyzing datasets with large noise level as the hyperplanes can not be defined precisely. Therefore, especially in the context of high-power laser experiments, one has to be cautious to apply SVMs if there are considerable stability issues.

\subsubsection{Convolutional neural networks}\label{CNN}
Convolutional neural networks (CNNs) are a type of neural network (cf. \cref{neural_nets}) that is particularly well-suited for image classification\cite{Rawat.2017}, but are also used in various other problems. 

Such a network is composed of sequential convolutional layers, in each of which, an $N\times N$ kernel (or ``filter'' matrix) is convolved with the output of the previous layer. This operation is done by sliding the kernel over the input image, and each pixel in the output layer is calculated by the dot product of the kernel with a sub-section of the input image centered around the corresponding pixel. Resultingly, convolutional layers are capable of detecting local patterns in the $N\times N$ region of the kernel. The kernel is parameterised with weights, which are learned via backpropagation as in a regular neural network. Within a layer, there can also be multiple channels of the output; practically thought of as multiple kernels being passed over the image, allowing for different features to be detected. The early layers of a CNN detect simple structures such as edges, but by adding multiple layers with varying kernel sizes, the network can perform high level abstraction in order to detect complicated patterns. 

The convolutional layer makes the CNN very efficient for image classification. It allows the network to learn translation-invariant features - a feature learned at a certain position of an image can be recognized at any position on the same image. 

In order to detect patterns that are non-local in the image, \emph{pooling} is often applied. There exist many schemes of pooling, but the general concept is to take a set of pixels in the input and to apply some operation that turns them into 1 pixel. Examples include max-pooling (taking the maximum of the set of pixels) and average pooling. This operation decreases the dimensions of the image, and therefore allows a subsequent convolutional layer with the same $N\times N$ kernel to detect features that were much further apart in the original image. Typical CNNs will use multiple pooling layers to decrease the dimensions until a kernel can nearly span the whole image to detect any non-local pattern. The output is then flattened and several fully connected layers can be used to manipulate the data for the relevant (ie. regression or classification) task.

While the use of deeper CNNs with an increasing number of layers tends to improve performance\cite{Szegedy.2015}, architectures can suffer from unstable gradients in training via backpropagation\cite{Balduzzi.2017}, showing that some deeper architectures are not as easy to optimize. One particularly influential solution to this problem was the introduction of the residual shortcut, where the input to a block is added to the output of the block. In backpropagation, this enables the `skipping' of layers, effectively simplifying the network and leading to better convergence. This was first proposed and implemented in ResNet~\cite{He.2016}, which has since become a standard for deep CNN architectures\cite{Szegedy.2017}, with any number of variations. It should be mentioned though that a number of competitive networks without residual shortcuts exist, e.g. AlexNet \cite{Krizhevsky.2017}.

\begin{figure}[t]
    \centering
    \includegraphics[width=\linewidth]{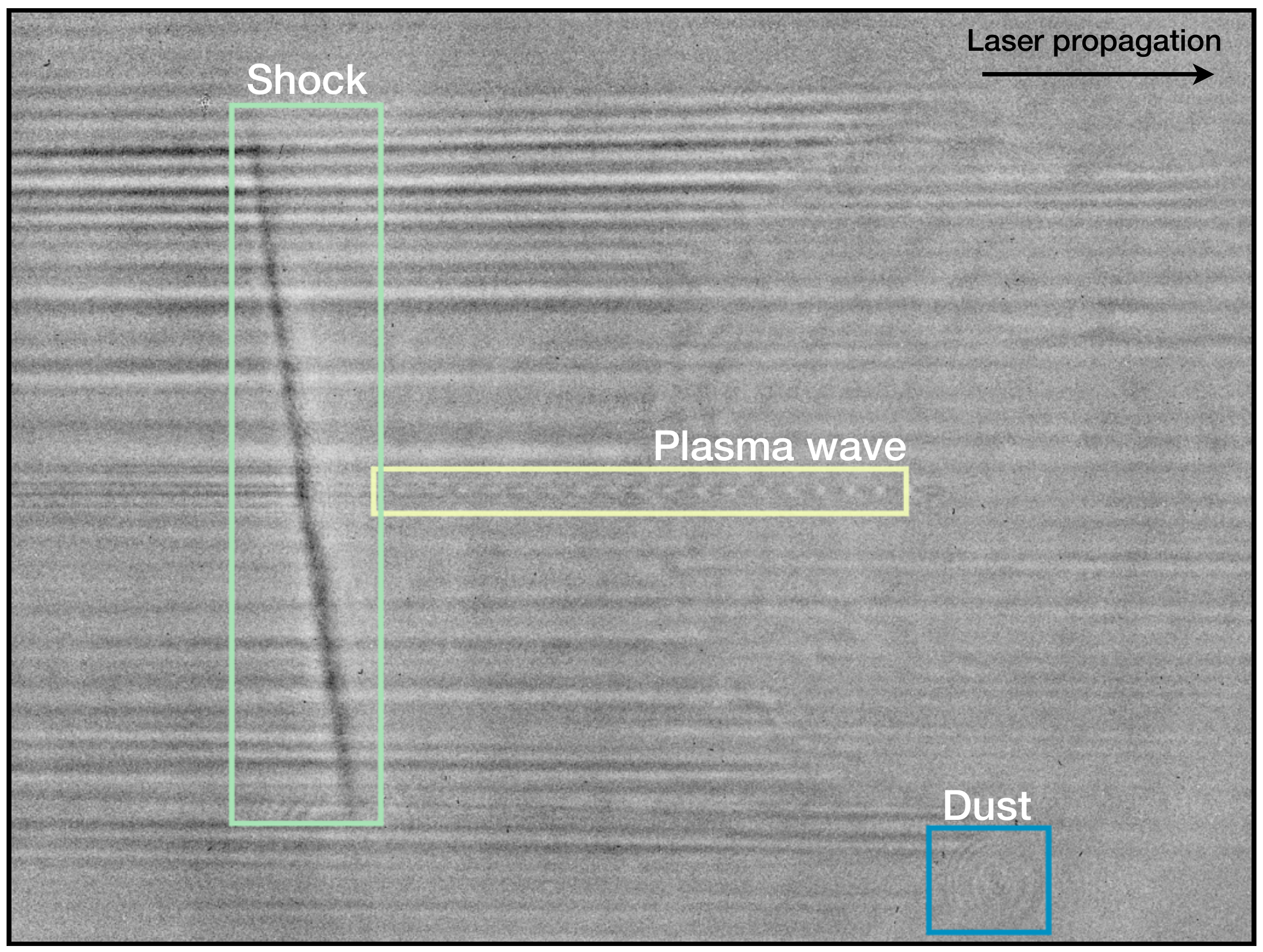}
    \caption{\textbf{Application of object detection} to a few-cycle shadowgram of a plasma wave: The plasma wave, the shadowgram of a hydrodynamic shock, and the diffraction pattern caused by dust are correctly identified by the object detector and located with bounding boxes. Adapted from Lin \emph{et al.} \cite{Lin2022HPLSE}.
    }
    \label{fig:LinObjectDetect}
\end{figure}

\subsection{Object detection}

In the context of image data, object detection can be seen as an extension of classification, yielding both a label as in classification tasks and the position of that object, illustrated in \cref{fig:classificationDetection}b. The main challenge is that complex images can contain many objects with different features, while the number of objects can also differ from one image to another. Therefore, object detection techniques require a certain flexibility regarding their structure. 

The Viola-Jones algorithm\cite{Viola.2001} is one of the most popular object detection algorithms from the 2000s, pre-dating the recent network-based object detectors. Generally, it involves detecting objects by looking at the image as a set of small patches, and identifying so-called \emph{Haar-like} features. The latter are patterns that occur frequently in images, and can be used to distinguish between different objects. Viola-Jones detects objects by first analyzing  an image at different scales. For each scale, it looks for features by scanning the image with a sliding window, and for each window, it computes a list of features used to identify the object. If the object is detected, the algorithm returns the bounding box of the object. The Viola-Jones framework does not allow for simultaneous classification and instead requires a subsequent classifier such as an SVM for that task. Compared to its network-based alternatives, Viola-Jones is worse in terms of precision but better in terms of computational cost, allowing real-time detection at high frame rates\cite{granger2017comparison}.

Object detection networks are more complicated than a regular CNN, because the length of the output layer of the neural network can not be pre-defined due to the unknown number of objects on an image. A possible solution is to divide the image into many regions and to construct a CNN for each region, but that leads to significant computational cost. Two family of networks are developed to detect objects at reasonable computational cost.
The region proposal network (RPN) takes an image as input and outputs a set of proposals for possible objects in the image. A CNN is used to find features in each possible region and to classify the feature into known category. The RPN is trained to maximize the overlap between the proposals and the ground truth objects. The state-of-the-art algorithm in this family is the Faster-RCNN\cite{ren2015NIPS}. 

An alternative to region-based CNNs is the YOLO\cite{Redmon2016CVPR} family. YOLO stands for ``You Only Look Once'' as it only looks at the image once with a single neural network to make predictions. This is different from other object detection algorithms, which often employ many neural networks for the image. As a result, YOLO is typically much faster, allowing for real-time object detection. The disadvantage is that it is not as accurate as some of the other object detection algorithms. Nevertheless, the fast inference speed of YOLO is particularly appealing to high-power laser facilities with high-repetition-rate capability.

\subsection{Segmentation}

Semantic segmentation\cite{Chen.2017} is a related task in computer vision that seeks to create a pixel-by-pixel mapping of the input image to a class label, not only a bounding box as in object detection, see \cref{fig:classificationDetection}c. By doing so, segmentation defines the exact boundary of the objects.

Many semantic segmentation architectures are based on a fully convolutional networks (FCNs) \cite{Long.2016} and have evolved considerably in recent years. Since FCNs suffered from the issue of semantic gaps, where the output had a much lower resolution than the input, skip connections were introduced to allow the gradient to backpropagate through the layers to improve the performance. An example of such advanced network architectures is the U-Net \cite{Ronneberger.2015} and the DeepLab network \cite{Chen.2017}, which is based on ResNet-101. Both of these architectures use residual skip connections to maintain gradient flow. The advantage of semantic segmentation compared to standard object detection is that the network can easily localize multiple objects of the same class in an image. The disadvantage is that one needs to train a separate network for each class.

Related is instance segmentation\cite{hafiz2020survey}, which goes a step further and distinguishes each individual instance of an object, not just the class. Instance segmentation is a significant challenge, as it requires the network to be able to distinguish between two instances of the same object, e.g. two cats. Instance segmentation architectures are typically based on Mask R-CNN\cite{He.2018}, which combines a CNN with a region-based convolutional neural network (R-CNN)\cite{Girshick.2015} and a fully convolutional network (FCN)\cite{Long.2016}. Note that mask R-CNN can be used for both semantic and instance segmentation.

\example{One of the prime examples for machine learning assisted image analysis is the automated detection and classification of laser damage. Researchers at the National Ignition Facility (NIF) have pioneered this approach with several works on neural networks for damage classification. For instance, Amorin \emph{et al.} \cite{amorin2019hybrid} trained CNNs based on the AlexNet and Inception architectures to distinguish between different kinds of laser damages. Another example for the use of both SVM and CNN-based classification in high-power laser systems was recently presented by T. Pascu \cite{Pascu.2022}, who used both techniques for (supervised) anomaly detection in a laser beam profile at the ELI-NP facility. Chu \emph{et al.} \cite{chu2019detection} presented a first application of image segmentation to locate laser-induced defects on the optics in real time using a U-Net. Soltane \emph{et al.}\cite{soltane2022estimating} recently presented a deep learning pipeline to estimate the size of damages in glass windows at the Laser Mégajoule (LMJ) facility, using a similar U-Net architecture for segmentation. Li \textit{et al.}\cite{Li:20} combined damage detection via a deep neural network with postprocessing to position laser damage in three-dimensional space. The axial distance between the damage site and the imaging system is obtained numerically by the principle of holographic focusing. More examples for applications of object detection in a high-power laser laboratory have been reported in the work of Lin \emph{et al.} \cite{Lin2022HPLSE}. In addition to the aforementioned case of optical damages in a high-power laser beamline, the authors fine-tuned the YOLO network for object detection in few-cycle shadowgraphy of plasma waves and electron beam detection in an electron spectrometer. An example for the detected features in a shadowgram is presented in Fig. \ref{fig:LinObjectDetect}. The position and size of the detected objects are used to determine information oh physical quantities, such as the plasma wavelength and plasma density distribution.}

\begin{table*}[hp!]
    \centering
    \small
\begin{tabular}{p{4cm}|p{3.1cm}|p{3.5cm}|p{0.7cm} |p{0.7cm}| p{5cm}}
\toprule
 \textbf{Author, Year}  &  \textbf{Problem Type} & \textbf{ML Technique}   &  \textbf{Sim.} & \textbf{Exp.}   &  \textbf{Research field} \\
\midrule
Humbird \emph{et al.}, 2018 \cite{humbird2018deep} & Forward model & Neural net $\&$ decision tree & \cmark & \xmark & Inertial confinement fusion \\
Humbird \emph{et al.}, 2018 \cite{humbird2019transfer} & Forward model & Transfer learning & \cmark & \cmark & Inertial confinement fusion \\
Gonoskov \emph{et al.}, 2019\cite{gonoskov2019employing} & Forward model & Neural Network & \cmark & \cmark & High-harmonic generation \\
Maier \emph{et al.}, 2020 \cite{Maier.2020} & Forward model & Linear Regression & \cmark & \xmark & Laser wakefield acceleration \\
Kluth \emph{et al.}, 2020 \cite{kluth2020deep} & Forward model & Autoencoder $\&$ DJINN & \cmark & \cmark & Inertial confinement fusion \\
Kirchen \emph{et al.}, 2021 \cite{Kirchen2021PRL} & Forward model & Neural Network & \xmark & \cmark & Laser wakefield acceleration \\
Rodimkov \emph{et al.}, 2021\cite{rodimkov2021towards} & Forward model & Neural Network & \cmark & \xmark & Noise robustness in PIC codes\\
Djordjevic \emph{et al.}, 2021 \cite{Djordjevic2021PoP} & Forward model & Neural Network & \cmark & \xmark & Laser-ion acceleration \\
Watt \emph{et al.}, 2021 \cite{Watt2021Thesis} & Forward model & Neural Network & \cmark & \xmark & Strong-field QED \\
McClarren \emph{et al.}, 2021 \cite{mcclarren2021high} & Forward model & Neural Network & \cmark & \xmark & Inertial confinement fusion \\
Simpson \emph{et al.}, 2021 \cite{simpson2021development} & Forward model & Neural Network & \xmark & \cmark & Laser-solid interaction \\
Streeter \emph{et al.}, 2023 \cite{Streeter2023HPLSE} & Forward model & Neural Network & \xmark & \cmark & Laser wakefield acceleration \\
\midrule
Krumbügel \emph{et al.}, 1996 & Inverse problem & Neural network & \xmark & \cmark & Spectral Phase Retrieval \\
Sidky et al., 2005\cite{sidky2005robust} & Inverse problem & EM algorithm & \xmark & \cmark & X-ray spectrum reconstruction \\
Döpp et al., 2018\cite{Dopp.2018} & Inverse problem & Statistical iterative reconstruction & \xmark & \cmark & X-ray tomography with betatron radiation \\
Huang \emph{et al.}, 2014\cite{HUANG2014459} & Inverse problem & Compressed sensing &  \cmark & \xmark& ICF radiation analysis \\
Zahavy \emph{et al.}, 2018\cite{zahavy2018deep} & Inverse problem & Neural network & \xmark & \cmark & Spectral Phase Retrieval \\
Hu \emph{et al.}, 2020\cite{shackhartmann} & Inverse problem & Neural network & \xmark & \cmark & Wavefront measurement \\
Ma et al., 2020\cite{ma2020region} & Inverse problem & Compressed sensing & \xmark & \cmark &  Compton X-ray tomography \\
Li \emph{et al.}, 2021\cite{li2021efficient} & Inverse problem & Compressed sensing &  \cmark & \xmark & ICF radiation analysis \\
Howard \emph{et al.}, 2023 & Inverse problem & Compressed sensing / Deep unrolling&  \cmark & \xmark & Hyperspectral phase imaging \\
\midrule
Bartels et al., 2000\cite{Bartels.2000} & Optimization   &  Genetic algorithm &  \xmark  & \cmark & High-harmonic generation \\
Yoshitomi et al., 2004\cite{Yoshitomi.2004} & Optimization   &  Genetic algorithm &  \xmark  & \cmark & High-harmonic generation \\
Zamith et al., 2004\cite{Zamith.2004} & Optimization   &  Genetic algorithm &  \xmark  & \cmark & Cluster dynamics \\
Yoshitomi et al., 2004\cite{Moore.2005} & Optimization   &  Genetic algorithm &  \xmark  & \cmark & Cluster dynamics \\
Nayuki et al., 2005\cite{nayuki2005production} & Optimization   &  Genetic algorithm &  \xmark  & \cmark & Ion acceleration \\
He et al., 2015\cite{he2015coherent,He.2015y53} & Optimization   &  Genetic algorithm &  \xmark  & \cmark & Laser wakefield acceleration  \\
Streeter \emph{et al.}, 2018\cite{Streeter.2018} & Optimization &  Genetic algorithm &  \xmark  & \cmark & Cluster dynamics \\
Lin et al., 2019\cite{lin2019adaptive}  & Optimization & Genetic algorithm & \xmark  & \cmark & Laser wakefield acceleration \\
Dann et al., 2019\cite{Dann2019PRAB} & Optimization & Genetic \& Nelder-Mead algorithms & \xmark & \cmark & Laser wakefield acceleration \\
Shalloo et al., 2020\cite{shalloo2020automation} & Optimization & Bayesian optimization & \xmark & \cmark  & LWFA, betatron radiation \\
Smith et al., 2020\cite{smith2020optimizing} & Optimization & Genetic algorithm & \cmark & \xmark & Laser-ion acceleration \\
Kain et al., 2020\cite{kain2020sample} & Optimization & Reinforcement learning & \cmark & \cmark & Plasma wakefield acceleration \\
Jalas et al., 2021\cite{jalas2021bayesian} & Optimization & Bayesian optimization & \cmark & \cmark & Laser wakefield acceleration \\
Pousa et al., 2022\cite{ferranpousa.2022} & Optimization & Bayesian optimization & \cmark & \xmark & Laser wakefield acceleration \\
Dolier et al., 2022\cite{Dolier2022NJP}  & Optimization & Bayesian optimization & \cmark & \xmark & Laser-ion acceleration \\
Irshad et al., 2023\cite{Irshad2022MO,irshad2023pareto} & Optimization & Bayesian optimization & \cmark & \cmark & Laser wakefield acceleration \\
Loughran et al., 2023\cite{Loughran2023arxiv}  & Optimization & Bayesian optimization & \xmark & \cmark & Laser-ion acceleration \\
\midrule
Chu \emph{et al.}, 2019 \cite{chu2019detection} &  Image Analysis & Neural Network & \xmark & \cmark & Laser damage segmentation \\
Amorin \emph{et al.}, 2019 \cite{amorin2019hybrid} & Image Analysis & Neural Network & \xmark & \cmark & Laser damage analysis \\
Li \textit{et al.}, 2020 \cite{Li:20} & Image Analysis & Neural Network & \xmark & \cmark & Laser damage detection in 3D \\
Hsu \emph{et al.}, 2020\cite{hsu2020analysis} & Feature analysis & Six supervised learning methods & \xmark & \cmark & Inertial confinement fusion \\
Lin \emph{et al.}, 2021\cite{Lin2021PoP} & Feature analysis & Four supervised learning methods & \xmark & \cmark & Laser wakefield acceleration \\
Willmann \emph{et al.}, 2021\cite{willmann2021data} & Dimensionality reduction & Autoencoder & \cmark & \xmark & Laser wakefield acceleration \\
Stiller \emph{et al.}, 2022\cite{stiller2022continual} & Data compression & Autoencoder & \cmark & \xmark & Laser wakefield acceleration \\
Pascu, 2022 \cite{Pascu.2022} & Image Analysis & SVM / Neural Network & \xmark & \cmark & Laser anomaly detection \\
Soltane \emph{et al.}, 2022\cite{soltane2022estimating} &  Image Analysis & Neural Network & \xmark & \cmark & Laser damage segmentation \\
Lin \emph{et al.}, 2023 \cite{Lin2022HPLSE} & Image Analysis & Neural Network & \xmark & \cmark & Laser wakefield acceleration and damage detection \\
\bottomrule
\end{tabular}
    \caption{Summary of papers used as application examples in this review, sorted by year for each section.}
    \label{tab:summary}
\end{table*}

\section{Discussion and Conclusions}

In this paper, we have presented an overview of techniques and recent developments in machine learning for laser-plasma physics. As we have seen, early proof-of-concept papers appeared in the late 1990s and early 2000s, but the computing power available at the time was typically not sufficient to make the approaches competitive with established methods or to reach a suitable level of accuracy. In the mid-2010s, a resurgence of interest in the field began, with an ever-increasing number of publications. A significant fraction of the papers that have been reviewed here are experimental in nature, especially regarding optimization, see \cref{tab:summary}. On one hand, this can be attributed to the increasing digitization of the laboratory environment, with control systems, data acquisition, and other developments providing access to large amounts of data. On the other hand, the complexity of modern experiments acts as a catalyst for the development of automated data analysis and optimization techniques. Deployment of machine-learning techniques in a real-world environment can however be challenging, e.g. due to noise, jitter and drifts. This is certainly one of the reasons why the most advanced machine learning techniques, such as multi-objective optimization or deep compressed sensing, tend to be first tested with simulations.

Among the methods being employed we also observed some general trends. For instance, while genetic algorithms have been very popular in the past for global optimization, there has been an increasing amount of papers focusing on Bayesian optimization. This is likely due to the fact that both simulations and experiments in the context of laser-plasma physics are very costly, making the use of Bayesian approaches more appealing. In most experimental settings, local optimization algorithms such as gradient descent or the Nelder-Mead algorithm are less suitable because of the large number of iterations needed and their sensitivity to noise. Reinforcement learning, especially in its model-free incarnation, suffers from the same issue, which explains why only very few examples of its use exist in adjacent research fields. While rather popular among data scientists due to their simplicity and interpretability, decision tree methods have not seen wide application in laser-plasma physics. In part, this is likely due to the fact that these methods are often considered to have more limited capabilities in comparison to neural networks, making it more attractive to directly use deep neural networks. In the context of ill-posed inverse problems it is to be expected that end-to-end neural networks or hybrid approaches will gradually replace traditional methods, such as regularization via total variation. That said, the simplicity and bias-free nature of least-squares methods are likely to ensure their continued popularity, at least in the context of easier to solve well-posed problems.

Much of the success of machine-learning techniques stems from the fact that they are able to leverage prior knowledge, be it in the form of physical laws (e.g. via physics-informed neural networks) or in the form of training data (e.g. via deep learning). Regarding the latter, the importance of preparing input data cannot be overstated. A popular saying in supervised learning is "garbage in, garbage out", meaning that the quality of a model's output heavily depends on the quality of the training data. Important steps are for instance pre-processing\cite{garcia2016big} (noise removal, normalization, etc.) and data augmentation\cite{shorten2019survey} (rotation, shifts, etc.). The latter is of particular importance when dealing with experimental data, for which data acquisition is usually costly, making it challenging to acquire enough data to train a well-performing model. Furthermore, even when using regularization techniques, diversity of training data is very important to ensure good generalization capabilities and to avoid bias.

Two outstanding issues for a wide adoption of machine learning models in the laser-plasma community are interpretability and trustworthiness, both regarding the model itself and on the user side. While some machine-learning models such as decision trees can be interpreted comparably easily, the inner workings of advanced models like deep neural networks are often difficult to understand. This issue is amplified by the fact that integrated machine learning environments allow users to quickly build complex models without a thorough understanding of the underlying principles. We hope that this review will help to alleviate this issue, by providing a better understanding of the origin, capabilities and limitations of different machine-learning techniques. Regarding trustworthiness, quantification of \textit{aleatoric} uncertainty in training data and \textit{epistemic} uncertainty of the model remains an important research area\cite{hullermeier2021aleatoric}. For instance, well-tested models may break down when exposed to unexpected input data, e.g. due to drifts in experimental conditions or changes in the experimental setup. Such issues can for instance be addressed by incorporating uncertainty quantification into models to highlight unreliable predictions.

As our discussion has shown, there is an ever increasing interest in data-driven and machine learning techniques within the community and we hope that our paper provides useful guidance for those starting to work in this rapidly evolving field. To facilitate some hands-on experimentation, we conclude with a short guide how to get started to implement the techniques we discussed in this paper. Most of these are readily implemented in several extensive libraries:  Scikit-learn\cite{pedregosa2011scikit}, Tensorflow\cite{tensorflow2015-whitepaper}, and PyTorch\cite{NEURIPS2019_9015} being among the most popular ones. Each of these libraries has its own strengths and weaknesses. In particular, deep learning libraries such as Tensorflow and PyTorch are tailored for fast computations on graphics processing units (GPUs), whereas libraries such as Scikit-learn are designed for more general machine learning tasks. Higher level frameworks exist to facilitate the training of neural networks, e.g. MLflow or PyTorch lightning.

The Darts library\cite{Darts} contains  implementations of various time series forecasting models, and also acts as a wrapper for numerous other libraries related to forecasting. Many numerical optimization algorithms such as derivative-based methods and differential evolution can be easily explored using the optimization library of SciPy\cite{2020SciPy-NMeth}. While this includes for instance differential evolution, more sophisticated evolutionary algorithms such as multi-objective evolutionary methods require dedicated libraries like pymoo\cite{pymoo} or PyGMO\cite{Biscani2020}. Bayesian optimization can for instance be implemented within Scikit-learn or using the experimentation platform Ax. The highly-optimized BoTorch library\cite{balandat2020botorch} can be used for more advanced applications, including for instance multi-objective, multi-fidelity optimization. Some libraries are specifically tailored to hyperparameter optimization, such as the popular optuna library\cite{akiba2019optuna}.

While all of the above examples are focused on python as an underlying programming language, machine learning tasks can also be performed using many other programming platforms or languages such as Matlab or Julia. Jupyter notebooks provide a good starting point and some online implementations, such as Google Colab, even give limited access to GPUs for training. The reader is encouraged to explore the various frameworks and libraries to find the one that best suits their needs.

\section*{Acknowledgements}

We thank all participants of the \textit{LPA Online Workshop on Control Systems and Machine Learning} for their contributions to the workshop, many of which are referenced throughout this manuscript. We particularly thank N. Hoffmann, S. Jalas, M. Kirchen, R. Shalloo and A.G.R. Thomas for helpful feedback and comments on the manuscript. The authors acknowledge the use of GPT-3\cite{brown2020language} (text-davinci-003) in the copy-editing process of this manuscript.

\end{document}